\documentclass[11pt]{article}

\usepackage{booktabs}
\usepackage{multirow}
\usepackage{array}
\usepackage{makecell}
\usepackage{subcaption}
\usepackage{enumitem}

\usepackage{pifont} 
\newcommand{\cmark}{\ding{51}} 
\newcommand{\xmark}{\ding{55}} 

\usepackage{booktabs}
\usepackage[normalem]{ulem}

\usepackage[final]{acl}

\usepackage{times}
\usepackage{latexsym}

\usepackage[T1]{fontenc}

\usepackage[utf8]{inputenc}

\usepackage{microtype}

\usepackage{inconsolata}

\usepackage{graphicx}

\title{Decomposed Prompting Does Not Fix Knowledge Gaps, But Helps Models Say \emph{“I Don’t Know”}}

\author{
Dhruv Madhwal$^{1,*}$ \quad
Lyuxin David Zhang$^{2,*}$ \quad
Dan Roth$^{2,3}$ \quad \\
\textbf{Tomer Wolfson$^{2,\dagger}$} \quad
\textbf{Vivek Gupta$^{1,\dagger}$} \\
\vspace{0.3em}
$^{1}$Arizona State University \qquad
$^{2}$University of Pennsylvania \qquad
$^{3}$Oracle AI \\
{\normalsize
\texttt{\{dmadhwal, vgupt140\}@asu.edu, \{davidzlx, danroth, wolfsont\}@upenn.edu}
}
}

\usepackage{amsmath}  

\begin{document}
\maketitle
\renewcommand{\thefootnote}{\fnsymbol{footnote}}
\footnotetext[1]{Equal contribution.}
\footnotetext[2]{Equal advising.}
\renewcommand{\thefootnote}{\arabic{footnote}}
\setcounter{footnote}{0}
\begin{abstract}

Large language models often struggle to recognize their knowledge limits in closed-book question answering, leading to confident hallucinations. While decomposed prompting is typically used to improve accuracy, we investigate its impact on reliability. We evaluate three task-equivalent prompting regimes: Direct, Assistive, and Incremental, across different model scales and multi-hop QA benchmarks. We find that although accuracy gains from decomposition diminish in frontier models, disagreements between prompting regimes remain highly indicative of potential errors. Because factual knowledge is typically stable while hallucinations are stochastic, cross-regime agreement provides a precise signal of internal uncertainty. We leverage this signal to implement a training-free abstention policy that requires no retrieval or fine-tuning. Our results show that disagreement-based abstention outperforms standard uncertainty baselines as an error detector, improving both F1 and AUROC across settings. This demonstrates that decomposition-based prompting can serve as a practical diagnostic probe for model reliability in closed-book QA. All code, data and prompts are available at: \url{https://github.com/dhruvmadhwal/disagreement-based-abstention}.


\end{abstract}

\section{Introduction}
\label{sec:intro}

Large language models (LLMs) are increasingly deployed in settings where they must answer factual questions without access to external retrieval or verification, such as privacy-restricted systems, on-device applications, and time-critical decision-making pipelines \cite{WANG2025100755, urlana2025llmsindustriallensdeciphering, qu2025mobileedgeLLM}. In these closed-book environments, models must rely entirely on their internal knowledge and reasoning capabilities. This makes closed-book factual question answering particularly challenging: when a model lacks sufficient knowledge it cannot defer to evidence and instead risks producing confident but incorrect answers \cite{simhi-etal-2025-trust}. As a result, a reliable closed-book QA model should not merely have high factual accuracy but should also be able to recognize uncertainty and appropriately refrain from answering when the requisite knowledge is unknown.

A widely adopted way to improve closed-book QA is to change how models are prompted to reason. \emph{Direct} reasoning asks the model to generate an answer in one step without an explicit intermediate structure, relying on the model's internal reasoning. Question decomposition \cite{min-etal-2019-multi, wolfson-etal-2020-break, khot2023decomposed} explicitly structures the solution process before producing an answer. Two decomposition variants are \emph{Assistive}, where a full set of intermediate steps is generated and then used to answer, and \emph{Incremental}, where the model is guided through a sequence of subquestions and answers them one step at a time. Crucially, these decompositions change only the reasoning path and do not introduce new facts or supervision, which makes them a controlled lens for studying accuracy and reliability in closed-book QA.

Our study spans across six multi-hop QA benchmarks and nine LLMs at varying scales. Its results reveal a striking, scale-dependent shift in how question decomposition functions. For smaller and mid-scale models, decomposition acts primarily as a scaffold that improves accuracy by providing structure. However, for frontier LLMs, these accuracy gains diminish, while the diagnostic value of the prompt increases. We find that for large-scale models, a disagreement between direct and decomposed outputs is a highly precise signal of an underlying error. In effect, as models scale, decomposition evolves from a tool for improving performance into a powerful training-free auditor for identifying fragile beliefs.

Based on these insights, we introduce Disagreement-Based Abstention (\textsc{DBA}), a simple and effective reliability method. \textsc{DBA} compares a model’s Direct answer to an answer obtained through a task-equivalent decomposition and triggers an \textit{``I don’t know''} response whenever the two final answers are conflicting. Unlike other LLM uncertainty methods, \textsc{DBA} requires no additional training or external retrieval and relies on the model’s exhibited behavior rather than self-reported confidence scores. Across nine LLMs, this approach outperforms standard uncertainty baselines, improving both F1 and AUROC and flagging confident-but-incorrect answers that other calibration methods often miss.

In summary, our contributions are: (a) a comprehensive empirical analysis of how factual accuracy and cross-prompt consistency jointly vary across LLM scales, evaluation tasks, and prompting methods in closed-book multi-hop QA, comparing direct prompting with Assistive and Incremental decomposed prompting; and (b) \emph{Disagreement-Based Abstention} (\textsc{DBA}), a simple training-free abstention method that treats cross-prompt disagreement between direct and decomposed prompting as a reliability signal for deciding when closed-book answers should be trusted.

\section{Decomposition-based Prompting}
\label{sec:setup}

\label{sec:problem_setup}

We evaluate closed-book multi-hop QA across three prompting regimes that share a fixed gold decomposition for each question: \textbf{(a) Direct:} where the model answers the query directly without intermediate structure or explicit decomposition; \textbf{(b) Assistive:} the model receives the full sequence of sub-questions as its fixed context and generates all intermediate answers in a single call; and \textbf{(c) Incremental:} where the decomposition is executed through separate model calls, with sub-questions being answered one at a time and follow-up questions are derived using the answers from previous steps. In the incremental approach, each LLM call is focused solely on the current sub-question, without being distracted by the previous computation history. 

For each question, we measure: (1) correctness relative to the ground truth
and (2) consistency, defined as semantic agreement between the Direct answer
and the final answer produced by each decomposed prompting regime.


\paragraph{Why these prompting regimes?} Direct provides the baseline for closed-book multi-hop QA, and is a conventional single-step answer prompt in which question decomposition is done implicitly by the LLM. The two decomposed prompting regimes, Assistive and Incremental, capture the most common decomposition strategies used in the literature. Assistive mirrors single-prompt, full-decomposition approaches that present an explicit plan or chain of intermediate steps up front \cite{press-etal-2023-measuring, radhakrishnan2023questiondecompositionimprovesfaithfulness, wu2024_gendec}, while Incremental mirrors step-wise or least-to-most execution that decomposes the problem into a sequence of sub-questions answered one at a time \cite{zhou2023leasttomost, khot2023decomposed}. By having a fixed gold-standard question decomposition, we ensure that the semantics of the Assistive and Incremental approaches are identical to that of the Direct question. 
This controlled comparison enables us to evaluate how the execution style, whether single-call (Direct), single-call with explicit plan (Assistive), or iterative QA (Incremental), affects both accuracy and answer consistency.

\paragraph{Verified Reference Decompositions}

\begin{figure}[h]
\vspace{-1.0em}
  \includegraphics[width=\columnwidth]{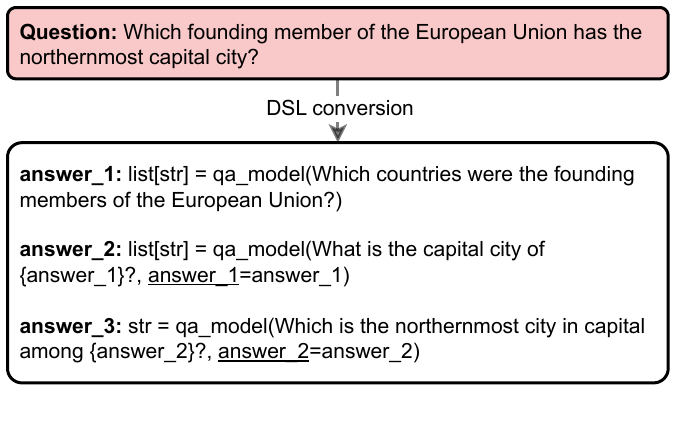}
  \vspace{-2.0em}
  \caption{DSL decomposition for a multi-hop question.}
  \label{fig:dsl}
  \vspace{-1.0em}
\end{figure}

A key requirement of our study is a fixed, gold-standard decomposition for each question. Using the syntax outlined in \citet{wolfson-etal-2026-monaco}, we represent the question decomposition using a domain-specific language (DSL) that specifies variables, answer types, and sub-questions (Figure~\ref{fig:dsl}). This representation gives us a precise, model-agnostic plan that can be reused across our prompting regimes.

We generate DSL decompositions using a two-stage pipeline with a strong LLM for decomposition. For benchmarks that provide step-by-step annotations, we prompt the LLM to translate the original steps into DSL. For benchmarks that lack labeled decompositions, we prompt \texttt{Gemini-2.5-Flash} to synthesize a DSL program from the original question. We then manually verify every DSL decomposition to ensure it is semantically and syntactically accurate, thereby eliminating any planning errors as a confounding factor.

The structured decomposition has three roles in our experimental design. First, decomposition steps can refer to the answers of previous steps using placeholders, thereby outlining the solution plan without the need to explicitly generate  all intermediate answers in advance. Second, it promotes deterministic execution by defining expected answer types (for example, integers or lists), which keeps intermediate outputs in a consistent format for subsequent steps. Third, it guarantees a semantic equivalence between the decomposition and the original query, such that if a model answers all sub-questions correctly, its final answer must be identical to the answer of the original query. As a result, Direct, Assistive and Incremental differ only in how they execute the same plan. Cross-regime answer disagreements can be interpreted as errors in factual knowledge or execution rather than an erroneous plan \cite{sinha2025illusiondiminishingreturnsmeasuring}.

\section{Experiments}
\label{sec:models_experiments}

\label{sec:experimental_setup}

\paragraph{Models.} 

To study how reasoning consistency behaves across different models at varying scales, we evaluate nine instruction-tuned LLMs spanning three distinct parameter sizes alongside a suite of state-of-the-art frontier systems. At the smaller scale ($\approx$8B parameters), we employ \texttt{Mistral-7B-Instruct-v0.3} \citep{jiang2023mistral7b}, \texttt{Llama-3.1-8B-Instruct} \citep{grattafiori2024llama3herdmodels}, and \texttt{Qwen3-8B} \citep{yang2025qwen3}. Our medium-scale representative is \texttt{Qwen3-32B} \citep{yang2025qwen3}, while for the large-scale tier ($\approx$70B parameters), we use \texttt{Llama-3.3-70B-Instruct} \citep{grattafiori2024llama3herdmodels} and \texttt{Qwen2.5-72B-Instruct} \citep{qwen2024qwen25}. Furthermore, we include three closed-source frontier models: \texttt{GPT-5.1} \citep{openai2025gpt51systemcard}, \texttt{Gemini-2.5-Pro}, and \texttt{Gemini-2.5-Flash} \citep{geminiteam2025gemini25}. All models are evaluated in their released instruction-tuned form using fixed decoding settings to reduce output variance. For all models except \texttt{GPT-5.1}, we use greedy decoding ($T=0$). For \texttt{GPT-5.1}, temperature is not exposed, so we use the API default with the reasoning effort set to `medium'.

\paragraph{Datasets.} We evaluate on six multi-hop QA datasets: \textit{Bamboogle} \citep{press-etal-2023-measuring}, \textit{FRAMES} \citep{el-asri-etal-2017-frames}, \textit{MuSiQue} \citep{trivedi-etal-2022-musique}, \textit{CRAG} \citep{yang2024crag}, \textit{HotpotQA} \citep{yang-etal-2018-hotpotqa}, and \textit{Mintaka} \citep{sen-etal-2022-mintaka}. We filter questions to avoid ambiguity and to ensure their multi-hop complexity, temporal-independence, and semantic clarity---resulting in 1,433 verified instances. Our full filtering statistics and benchmark splits are
reported in Appendix~\ref{sec:appendix_filtering}.




\paragraph{Prompting Regimes.}
Direct, Assistive, and Incremental all share the same closed-book multi-hop QA instruction and decoding settings. Direct is prompted using only instructions, while Assistive adds the gold DSL program, a small set of dataset-specific few-shot examples, and the model is instructed to answer all sub-questions in one response. Incremental executes the DSL line by line: at step $k$, we fill the template for sub-question $Q_k$ with answers from previous steps (e.g., ``In what city was [answer\_1] born?'' $\rightarrow$ ``In what city was Albert Einstein born?'') and issue it as a new query. Each hop is then posed as an isolated factual lookup, without access to the original top-level question or to any of the past or future steps.

\paragraph{Consistency Protocol.}
We measure factual consistency by anchoring to the Direct response, which represents the model's zero-shot answer based on its parametric knowledge. We compute two pairwise comparisons: Direct vs.\ Assistive and Direct vs.\ Incremental. An \emph{internal disagreement} is recorded whenever the answer to the decomposed approach is not semantically equivalent to the Direct answer. This definition is independent of the
actual correctness of either answer.

\label{evaluation}

\paragraph{Evaluation Measures.} We evaluate each prompting regime along two axes: accuracy, defined as agreement
with the gold answer, and consistency, defined as semantic agreement
across regimes. To address the brittleness of lexical measures such as EM or ROUGE, we employ an LLM-as-judge protocol, using \texttt{Gemini-2.5-Flash}, to
assess semantic equivalence. The judge follows a fixed rubric described in
Appendix~\ref{sec:appendix_prompts}. Namely, the rubric: (i) normalizes surface variation, including units, aliases, and abbreviations; (ii) enforces numeric tolerance and exact date matching;
and (iii) penalizes explicit contradictions. \textbf{Accuracy} is computed by
comparing a regime’s final answer to the gold reference. \textbf{Consistency} is
computed by comparing outputs from two regimes, such as Direct and
Incremental, to determine whether they express the same underlying
claim, independent of factual correctness.

\subsection{Results and Analysis}
\label{sec:results}

We analyze how LLM scale and prompting regime affect factual correctness and cross-regime consistency.
We identify three consistent patterns across benchmarks and LLMs.
First, both accuracy and consistency increase with model scale, though nontrivial inconsistency persists even in frontier models.
Second, decomposed prompting substantially improves accuracy for non-frontier models, but yields diminishing or negative returns for frontier LLMs.
Third, agreement between prompting regimes is highly indicative of accuracy, highlighting consistency
as a signal for answer confidence.

\paragraph{Scale, Accuracy and Consistency.}

\begin{figure*}[t]
  \centering
  \begin{subfigure}[b]{0.32\textwidth}
    \centering
    \includegraphics[width=\linewidth]{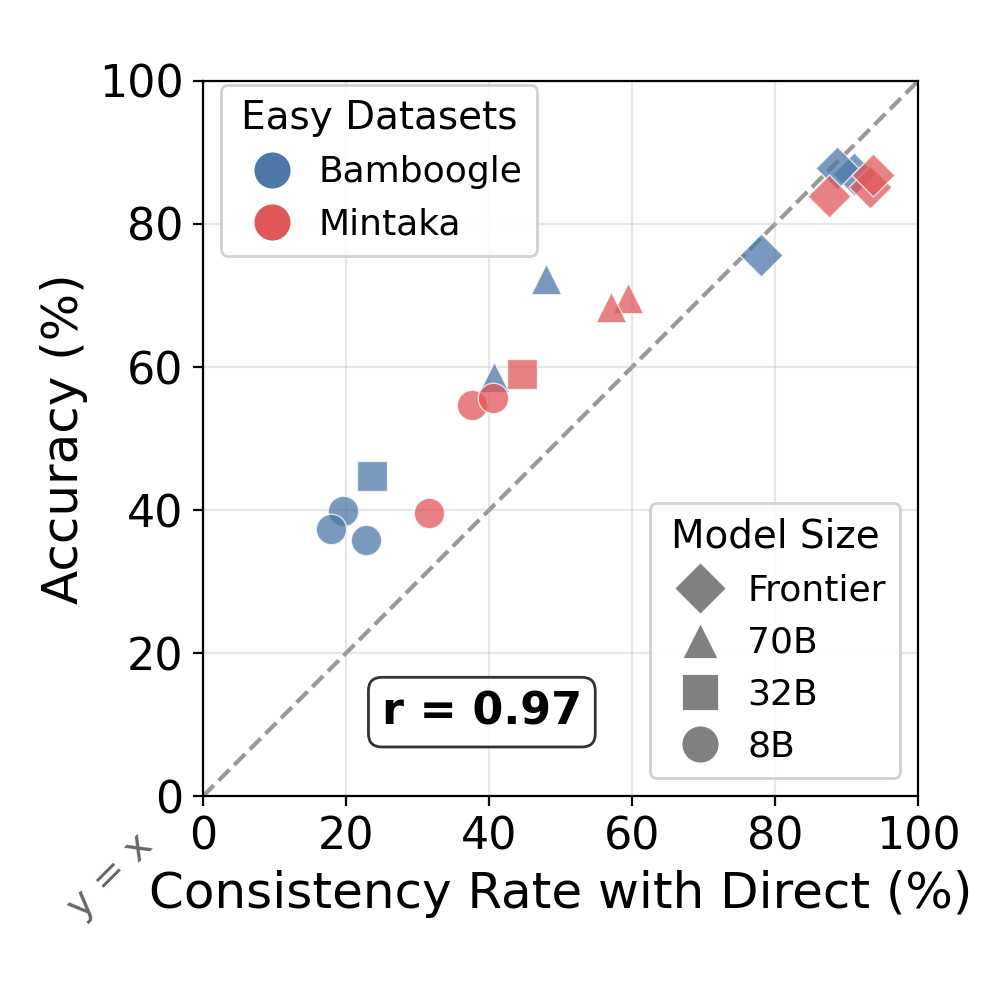}
    \label{fig:accuracy-vs-consistency-easy}
  \end{subfigure}
  \begin{subfigure}[b]{0.32\textwidth}
    \centering
    \includegraphics[width=\linewidth]{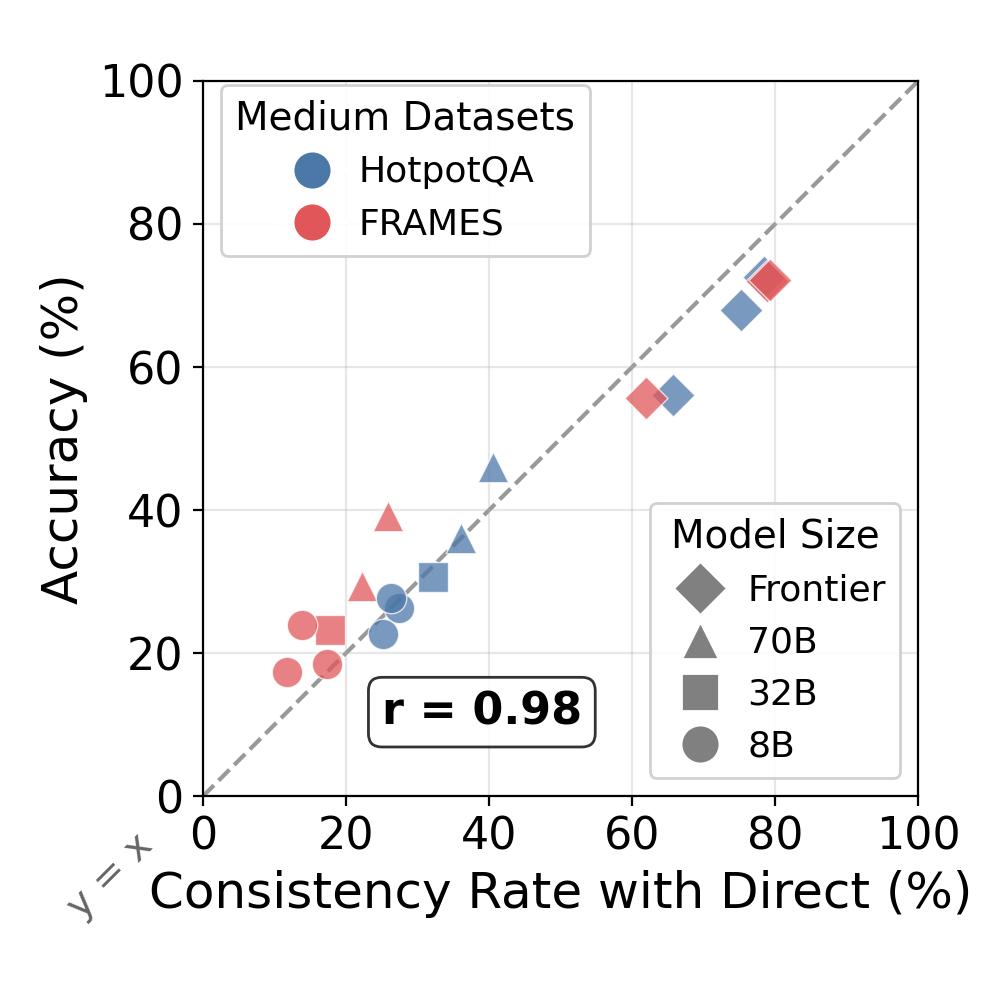}
    \label{fig:accuracy-vs-consistency-medium}
  \end{subfigure}
  \begin{subfigure}[b]{0.32\textwidth}
    \centering
    \includegraphics[width=\linewidth]{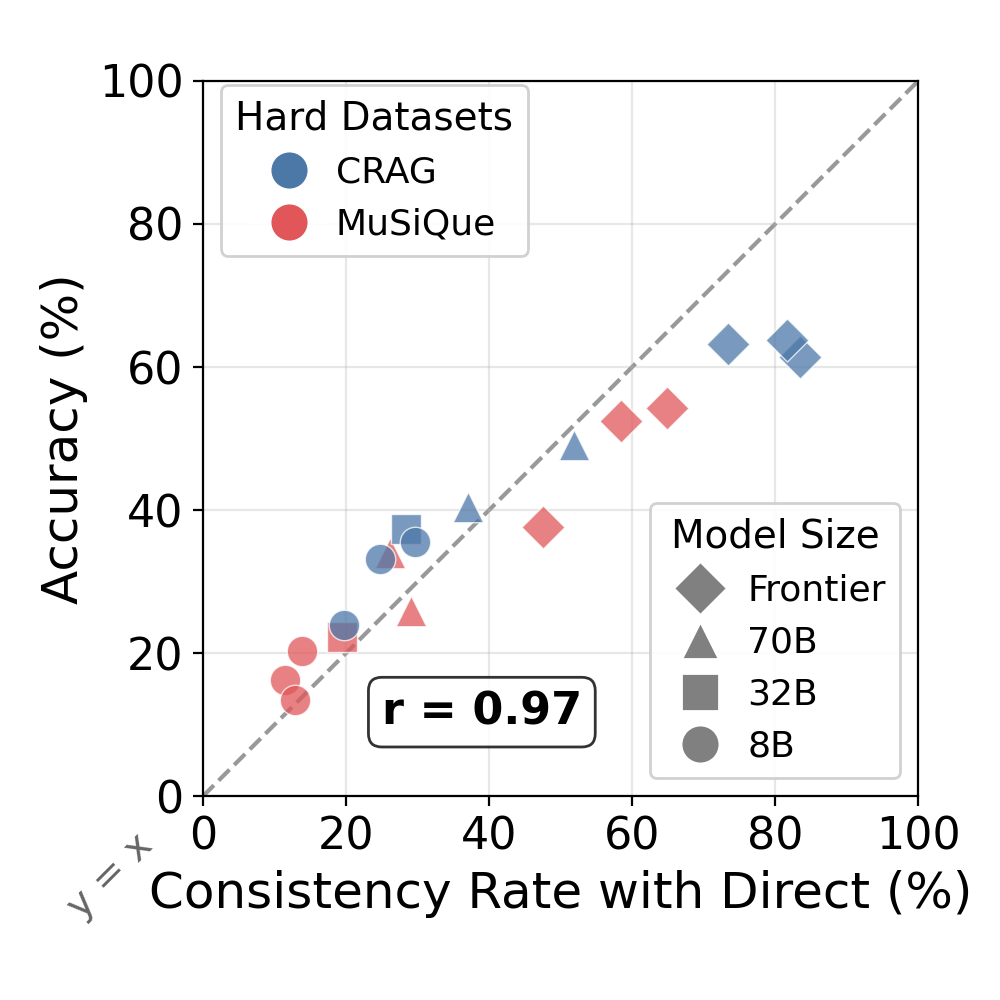}
    \label{fig:accuracy-vs-consistency-hard}
  \end{subfigure}
  \vspace{-2.5em}
  \caption{\small Accuracy vs.\ consistency rate across 9 models and 6 datasets, grouped by difficulty. Each point represents a (model, dataset) pair. Marker shape encodes LLM size (Frontier, 70B, 32B, 8B) while colors encode different evaluation datasets. }
  \label{fig:accuracy-vs-consistency-all}
  \vspace{-0.5em}
\end{figure*}


Both accuracy and cross-regime consistency increase with model scale.
Larger models are more likely to produce correct answers while
maintaining agreement across the Direct, Assistive, and Incremental
regimes, as shown in
Figure~\ref{fig:accuracy-vs-consistency-all}. The figure illustrates a
clear shift towards higher accuracy and consistency, as the LLM size increases.


However, significant inconsistency persists even among frontier models. Our most accurate LLMs still exhibit measurable cross-regime disagreement, for example, consistency on the MuSiQue benchmark reaches only 59.7\%. A comprehensive breakdown of accuracy and consistency across all models
and datasets is provided in Appendices~\ref{sec: appendix_accuracy_table}-\ref{sec: appendix_consistency_table}. This result reveals a fundamental failure of logical invariance: in a knowledge-grounded setting, semantically-equivalent queries should yield identical outputs regardless of
prompting regime. 
However, while all prompting regimes answer either the original question or a semantically equivalent decomposition, the lack of LLM consistency highlights failures in their execution or underlying parametric knowledge \cite{sinha2025illusiondiminishingreturnsmeasuring}.

\paragraph{Decomposition Gains Plateau in Frontier LLMs.}

\begin{table*}[t]
\setlength{\aboverulesep}{0.25pt}
\setlength{\belowrulesep}{0.25pt}
\setlength\tabcolsep{4pt}
\scriptsize
\begin{center}
\begin{tabular}{l ccc | ccc | ccc | ccc | ccc | ccc}
\toprule
\textbf{Model} & \multicolumn{3}{c}{\bf Bamboogle} & \multicolumn{3}{c}{\bf Mintaka} & \multicolumn{3}{c}{\bf HotpotQA} & \multicolumn{3}{c}{\bf CRAG} & \multicolumn{3}{c}{\bf FRAMES} & \multicolumn{3}{c}{\bf MuSiQue} \\ [0.1cm]
 &\bf Dir &\bf $\Delta$A &\bf $\Delta$I &\bf Dir &\bf $\Delta$A &\bf $\Delta$I &\bf Dir &\bf $\Delta$A &\bf $\Delta$I &\bf Dir &\bf $\Delta$A &\bf $\Delta$I &\bf Dir &\bf $\Delta$A &\bf $\Delta$I &\bf Dir &\bf $\Delta$A &\bf $\Delta$I \\
\toprule
Mistral 7B & 15.6 & +20.2 & +21.0 & 40.9 & +14.8 & +9.7 & 24.1 & +3.6 & +0.4 & 25.2 & +8.0 & $-$0.6 & 11.9 & +6.5 & +1.7 & 12.8 & +7.5 & +0.7 \\
Qwen 8B & 12.2 & +25.2 & +20.6 & 24.6 & +15.0 & +9.3 & 21.7 & +0.9 & $-$5.3 & 19.6 & +4.3 & +1.8 & 11.6 & +12.3 & +13.4 & 11.7 & +1.8 & +3.6 \\
Llama 8B & 21.3 & +18.5 & +26.7 & 33.3 & +21.4 & +24.7 & 25.3 & +1.0 & +2.2 & 27.6 & +8.0 & +14.1 & 8.9 & +8.5 & +9.9 & 9.2 & +7.0 & +7.2 \\
Qwen 32B & 18.7 & +26.0 & +25.2 & 38.6 & +20.5 & +14.8 & 31.4 & $-$0.7 & $-$3.7 & 27.0 & +10.4 & +8.6 & 14.3 & +8.9 & +7.2 & 16.0 & +6.2 & +3.9 \\
Qwen 72B & 31.7 & +26.8 & +29.3 & 53.7 & +14.6 & +18.1 & 34.9 & +1.2 & +1.2 & 31.9 & +8.6 & +9.2 & 17.4 & +11.9 & +11.6 & 23.0 & +2.8 & +3.2 \\
Llama 70B & 45.5 & +26.8 & +22.8 & 61.6 & +8.0 & +12.7 & 41.3 & +4.6 & +1.5 & 41.7 & +7.4 & +10.4 & 26.6 & +12.6 & +14.0 & 24.5 & +9.6 & +6.7 \\
\midrule
Gemini Flash & 83.7 & $-$8.1 & $-$3.2 & 83.2 & +0.7 & $-$0.4 & 59.7 & $-$3.7 & $-$6.4 & 64.4 & $-$1.2 & $-$2.5 & 61.4 & $-$5.8 & $-$5.8 & 37.2 & +0.4 & $-$1.4 \\
Gemini Pro & 85.4 & +1.6 & +1.5 & 85.6 & $-$0.3 & $-$2.3 & 68.1 & $-$0.1 & $-$4.3 & 64.4 & $-$3.1 & $-$4.9 & 71.7 & +0.3 & $-$6.8 & 47.5 & +5.0 & $-$2.0 \\
GPT 5.1 & 84.6 & +3.2 & +2.4 & 86.2 & +0.7 & $-$4.0 & 73.7 & $-$1.2 & $-$3.0 & 65.0 & $-$1.2 & $-$3.1 & 72.4 & $-$0.3 & $-$6.5 & 53.9 & +0.4 & $-$7.8 \\
\bottomrule
\end{tabular}
\end{center}
\vspace{-2.0em}
\caption{\small Direct accuracy (Dir, \%) and accuracy change from Direct to Assistive ($\Delta$A) and Incremental ($\Delta$I). Non-frontier models ($\le$70B) benefit substantially from decomposition; frontier models show diminishing or negative returns.}
\label{table:delta}
\end{table*}

Our results show a sharp scaling breakpoint in the efficacy of decomposed prompting (Table~\ref{table:delta}). For non-frontier models ($\leq 70$B), decomposition serves as a vital reasoning scaffold, leading to substantial accuracy gains. A striking example is Qwen-72B, where question decomposition leads to a $+26.8\%$ improvement on \textsc{Bamboogle}. These persistent, double-digit improvements ($\Delta A, \Delta I$) suggest that even high-capacity non-frontier models still benefit from explicit decomposition structure to solve complex multi-hop tasks.

However, this advantage vanishes for frontier LLMs. For Gemini-2.5 and GPT-5.1, the gains from decomposition plateau, falling to near-parity or even slipping into negative returns. We hypothesize that this transition reflects a ``ceiling effect'' where frontier models have internalized the necessary reasoning chains. In such models, explicit scaffolding, in the form of decomposed prompting, may no longer enhance, and could occasionally degrade performance.


\begin{table*}[t]
\vspace{-0.5em}
\setlength{\aboverulesep}{0.1pt}
\setlength{\belowrulesep}{0.1pt}
\setlength\tabcolsep{7.5pt}
\scriptsize
\centering
\begin{tabular}{l cc | cc | cc | cc | cc | cc}
\toprule
\textbf{Model} & \multicolumn{2}{c}{\textbf{Bamboogle}} & \multicolumn{2}{c}{\textbf{Mintaka}} & \multicolumn{2}{c}{\textbf{HotpotQA}} & \multicolumn{2}{c}{\textbf{CRAG}} & \multicolumn{2}{c}{\textbf{FRAMES}} & \multicolumn{2}{c}{\textbf{MuSiQue}} \\
 & \bf A & \bf I & \bf A & \bf I &\bf  A & \bf I & \bf A & \bf I & \bf A & \bf I & \bf A & \bf I \\
\midrule
Mistral 7B & 1.3x & 1.6x & 2.0x & 1.4x & 1.3x & 0.7x & 0.8x & 0.5x & 1.1x & 0.5x & 1.1x & 0.8x \\
Qwen 8B & 4.0x & 6.5x & 2.2x & 2.1x & 2.0x & 1.3x & 1.1x & 0.9x & 1.3x & 1.0x & 0.6x & 0.7x \\
Llama 8B & 2.2x & 3.3x & 3.5x & 3.8x & 1.8x & 1.9x & 1.5x & 1.6x & 1.0x & 0.7x & 1.6x & 0.7x \\
Qwen 32B & 3.6x & 2.8x & 5.5x & 4.1x & 2.2x & 2.2x & 2.1x & 2.1x & 1.3x & 1.0x & 2.0x & 1.4x \\
Qwen 72B & 8.8x & 12.0x & 6.2x & 7.0x & 2.0x & 2.0x & 4.8x & 4.2x & 1.4x & 2.2x & 2.2x & 1.2x \\
Llama 70B & 10.2x & 10.2x & 4.7x & 10.4x & 3.3x & 2.5x & 2.6x & 3.7x & 2.0x & 1.8x & 1.7x & 1.5x \\
\midrule
Gemini Flash & 5.4x & 8.4x & 14.6x & 13.6x & 4.9x & 2.6x & 6.1x & 5.2x & 4.7x & 3.7x & 3.6x & 2.6x \\
Gemini Pro & 34.0x & 16.5x & 50.2x & 27.4x & 13.3x & 5.4x & 10.8x & 10.8x & 10.1x & 5.0x & 5.7x & 2.8x \\
GPT 5.1 & 20.2x & 12.2x & 42.0x & 27.7x & 9.4x & 6.0x & 12.4x & 10.9x & 9.5x & 5.3x & 5.3x & 3.2x \\
\bottomrule
\end{tabular}
\vspace{-0.5em}
\caption{\small \textbf{The Reliability Multiplier.} ratio of (consistent, correct) to (inconsistent, correct) Direct answers for each dataset and decomposed prompting regime (A = Assistive, I = Incremental).}
\label{table:reliability_multiplier}
\vspace{-1.0em}
\end{table*}

\paragraph{Consistency as an Answer Reliability Signal.}

Decomposed prompting delivers large accuracy gains for non-frontier models but offers little gain for frontier LLMs (Table \ref{table:delta}). This raises a natural question: if decomposition no longer helps frontier models answer questions more accurately, what is it useful for? Our analysis suggests that its primary utility shifts from \emph{intervention} to \emph{inspection}: cross-prompt consistency becomes a direct signal of how consistent the model’s beliefs are.

\paragraph{Accuracy-Consistency Correlation.}
LLM accuracy and cross-regime consistency are closely tied across models and datasets, with Pearson correlation coefficients reaching up to
$r = 0.98$, as shown in
Figure~\ref{fig:accuracy-vs-consistency-all}. This tight coupling motivates a quantitative measure of how strongly consistency helps identify correct answers. We define the \textbf{Reliability Multiplier} $RM = \frac{\text{correct} ~\land~ \text{consistent}}{\text{correct} ~\land ~\text{inconsistent}}$, where correctness is anchored to the Direct answer. In practice, an $RM$ value of $k$ means that correct Direct answers occur $k$ times as often among consistent cases as among inconsistent cases.


\paragraph{Reliability across LLM Scales.}
The Reliability Multiplier ($RM$) varies sharply with model scale, reflecting a transition from noisy, stochastic behavior to more stable agreement across prompting regimes (Table~\ref{table:reliability_multiplier}). Smaller models ($\approx$8B) often exhibit near one or  inverted $RM$ values, consistent with a floor effect in which correct answers are too sparse for $RM$ to provide a stable signal. When correctness is sparse, $RM$ is
statistically fragile. This pattern is reflected in the near-one or inverted RM values for smaller models in Table~\ref{table:reliability_multiplier}. 
As model capacity increases into the 30B to 70B range, the multiplier
grows steadily, typically reaching values between
$2\times$ and $10\times$. At frontier scale, the effect can become much
larger, with values exceeding $50\times$ for Gemini Pro in our
measurements. These patterns indicate that, as scale increases, correct
Direct answers increasingly concentrate in the subset of cases where
regimes agree. As a result, cross-regime agreement becomes a
progressively stronger accuracy indicator.


\paragraph{Utility Shift.} These results highlight question decomposition as a dual-purpose tool, whose primary role shifts with LLM scale. For weaker models, decomposition functions primarily
as a generative aid that improves accuracy. For frontier models, where these gains saturate, decomposition serves as a diagnostic auditor: cross-regime agreement becomes a reliable signal for accuracy. While this signal is strongest for frontier LLMs, it remains useful across the full range of model sizes, with reliability multipliers greater than one in most settings.


\section{Selective Abstention Framework}
\label{sec:baselines}
\label{sec:abstention_method}
\begin{figure*}[t]  
    \centering
    \includegraphics[width=0.95\textwidth]{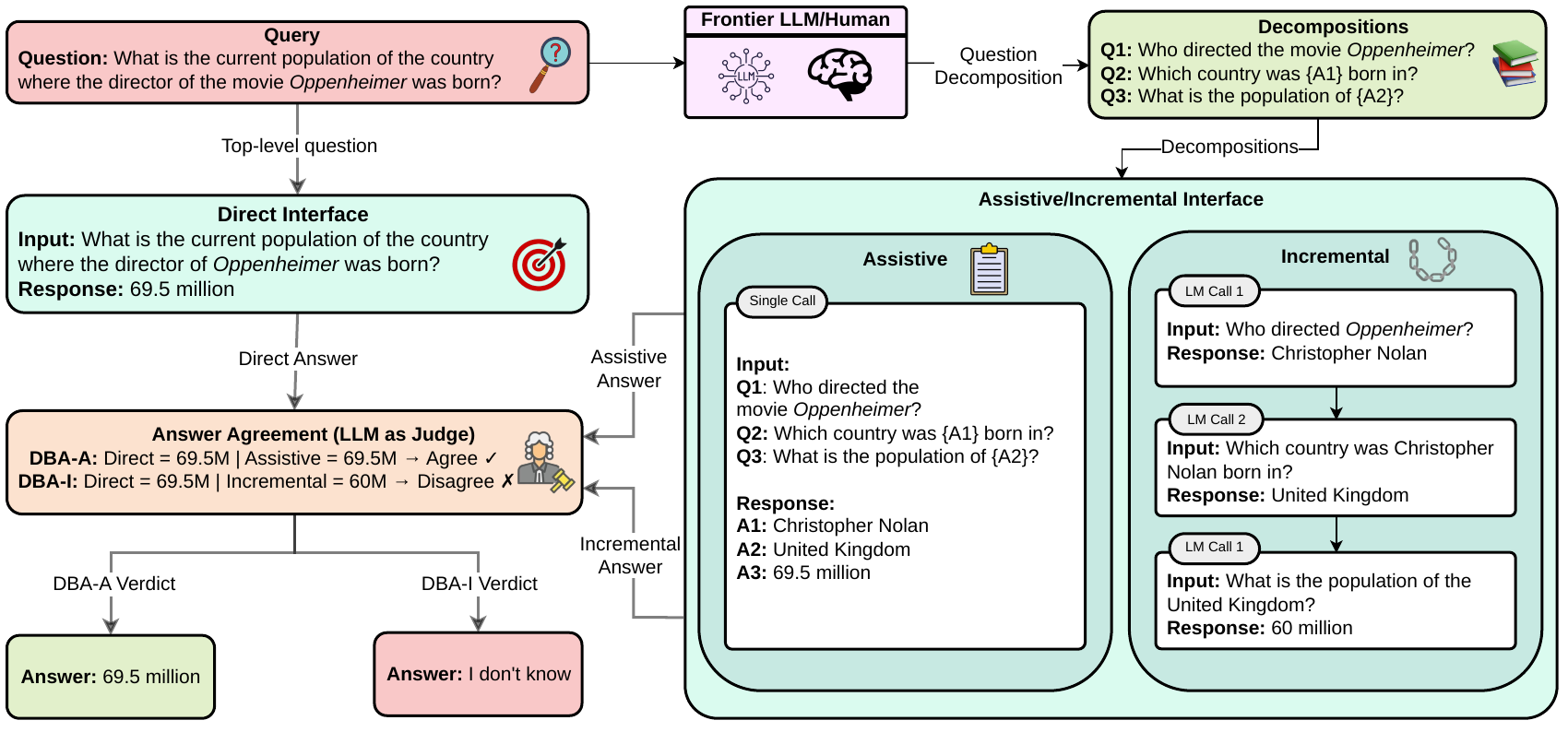} 
    \vspace{-0.5em} 
    \caption{\textbf{Disagreement-Based Abstention (DBA) Framework.} Our method compares a Direct answer against Assistive/Incremental reasoning paths, if the semantic claims disagree, the model abstains (IDK).}
    \label{fig:method}
    \vspace{-1.0em}
\end{figure*}

Our analysis shows that while Assistive and Incremental decompositions may not fundamentally expand a model’s internal knowledge, \emph{agreement} across different decomposed prompting regimes is a strong indicator of factual correctness. We exploit this cross-regime consistency to build a multi-prompt-based abstention policy. When different prompting regimes yield inconsistent final answers, we treat the instance as unreliable and output an ``I don't know'' response. Unlike other uncertainty prompting methods that rely on potentially unfaithful self-reported calibration, our approach uses answer inconsistency across decomposed prompting regimes as a direct signal of factual error.

\subsection{LLM Abstention on Factual Questions}

\paragraph{Disagreement-based Abstention (DBA).}
%


Given a question, we prompt an LLM twice, once using the Direct
regime and once using a task-equivalent decomposed regime. If the two
final answers are semantically distinct, the model abstains.
\textsc{DBA} instantiates this procedure using either an Assistive or an
Incremental decomposition, referred to as \textsc{DBA-A} and
\textsc{DBA-I}, respectively. Semantic equivalence is determined using
the same LLM-as-judge protocol described in
Section~\ref{evaluation}. The overall \textsc{DBA} pipeline is shown in
Figure~\ref{fig:method}.

\paragraph{Evaluation Metrics}
We evaluate LLM abstention policies as \emph{error detectors} where the positive class is \textbf{incorrect} Direct predictions. We treat the model’s Direct response as the primary claim-of-interest. Namely, correctness is defined by whether this Direct claim matches the ground-truth answer, and the Assistive/Incremental executions serve only as diagnostic signals for deciding whether to accept the Direct answer, or to abstain. This is in line with prior work (our abstention baselines below) which also focus on the model’s direct answer. 
 
We report four measures: \textbf{(a) Precision}, the fraction of rejected (abstained) instances whose \emph{Direct} answer is incorrect; \textbf{(b) Recall}, the fraction of all incorrect \emph{Direct} answers that are rejected; \textbf{(c) F1}, the harmonic mean of precision and recall; and \textbf{(d) AUROC}, which measures the method’s ability to separate incorrect from correct \emph{Direct} answers. 

While precision, recall, and F1 are sensitive to the base error rate, which varies widely across model scales (e.g., $\sim$10\% to $\sim$90\% on \emph{Bamboogle}), AUROC is less sensitive. In our binary setting, AUROC corresponds to the probability that an incorrect \emph{Direct} answer is rejected more strongly than a correct one, with a naive always-accept or always-reject strategy yielding AUROC $=0.50$.

\paragraph{Baselines.}
We compare \textsc{DBA} against three uncertainty-detection baselines. The first two (AYS, IC-IDK) are drawn from prior work \citep{cohen-etal-2023-lm}. Our third baseline is self-consistency, a popular prompting approach \citep{wang2023selfconsistency}.



\vspace{-0.3em}
\begin{itemize}[leftmargin=*]

\item \textbf{AYS (Are You Sure?)} first produces a Direct answer, then issues a binary
follow-up query asking the model to verify its
answer (\emph{``Are you sure regarding the correctness of <answer>?''}). The Direct answer is accepted if the model responds affirmatively
and otherwise rejected.
\vspace{-0.3em}

\item \textbf{IC-IDK (In-Context IDK)} appends an instruction to the Direct prompt that allows the model to respond with an explicit "I don't know" when uncertain \citep{cohen-etal-2023-lm}. For each model and dataset, we prepend $K=15$ demonstrations constructed from a held-out pool disjoint from the evaluation set. To ensure these demonstrations reflect each model's specific knowledge gaps, we select $D=4$ instances where that same model’s Direct answer was verified incorrect against the ground truth as IDK demonstrations, and $11$ instances where it answered correctly as standard answer demonstrations.

\vspace{-0.3em}

\item \textbf{Self-consistency} samples multiple responses from the Direct regime at
higher temperatures and predicts by majority vote
\citep{wang2023selfconsistency}. In our experiments,
a heuristic majority vote self-consistency achieves high precision but critically low recall, which
results in poor F1 performance. We therefore report its detailed
analysis in Appendix~\ref{sec:self_consistency_appendix} and focus our
main comparison on the remaining baselines.
\end{itemize}
\vspace{-0.3em}


\paragraph{Ensembles with DBA.}
We define ensemble variants that combine \textsc{AYS} with
\textsc{DBA}. \textsc{Ensemble-A} abstains if either \textsc{AYS} or
\textsc{DBA-A} abstains. \textsc{Ensemble-I} is defined analogously using
\textsc{DBA-I}.

\newcommand{\tblbest}[1]{\textbf{#1}}
\newcommand{\tblbestne}[1]{\uline{#1}}
\newcommand{\tblbestboth}[1]{\uline{\textbf{#1}}}

\begin{table*}[t]
\setlength{\aboverulesep}{0.5pt}
\setlength{\belowrulesep}{0.5pt}
\setlength{\tabcolsep}{7pt}
\scriptsize
\centering

\providecommand{\best}[1]{\textbf{#1}}
\providecommand{\bestne}[1]{\underline{#1}}
\providecommand{\bestboth}[1]{\underline{\textbf{#1}}}

\begin{tabular}{@{}p{1.1cm}l *{6}{cc}@{}}
\toprule
& & \multicolumn{2}{c}{\textbf{Bamboogle}} & \multicolumn{2}{c}{\textbf{CRAG}} & \multicolumn{2}{c}{\textbf{FRAMES}} & \multicolumn{2}{c}{\textbf{HotpotQA}} & \multicolumn{2}{c}{\textbf{Mintaka}} & \multicolumn{2}{c}{\textbf{MuSiQue}} \\
\cmidrule(lr){3-4}\cmidrule(lr){5-6}\cmidrule(lr){7-8}\cmidrule(lr){9-10}\cmidrule(lr){11-12}\cmidrule(lr){13-14}
\textbf{Model} & \textbf{Method} & \textbf{F1} & \textbf{AUC} & \textbf{F1} & \textbf{AUC} & \textbf{F1} & \textbf{AUC} & \textbf{F1} & \textbf{AUC} & \textbf{F1} & \textbf{AUC} & \textbf{F1} & \textbf{AUC} \\
\midrule

GPT 5.1 & AYS
  & 0.47 & 0.66
  & \tblbestne{0.66} & \tblbestne{0.74}
  & 0.51 & 0.67
  & 0.51 & 0.67
  & 0.42 & 0.66
  & 0.38 & 0.61 \\
& IC-IDK
  & 0.43 & 0.64
  & 0.60 & 0.70
  & 0.24 & 0.56
  & 0.58 & 0.71
  & 0.31 & 0.60
  & 0.27 & 0.57 \\
\cmidrule(lr){2-14}
& DBA-A
  & \tblbestboth{0.73} & \tblbestboth{0.81}
  & 0.53 & 0.67
  & 0.58 & 0.71
  & \tblbestne{0.70} & \tblbestne{0.79}
  & 0.43 & 0.65
  & \tblbestne{0.66} & \tblbestne{0.72} \\
& DBA-I
  & 0.56 & 0.72
  & 0.49 & 0.65
  & \tblbestboth{0.66} & \tblbestboth{0.77}
  & 0.64 & 0.75
  & \tblbestne{0.48} & \tblbestne{0.68}
  & 0.64 & 0.68 \\
\cmidrule(lr){2-14}
& Ensemble-A
  & 0.67 & 0.79
  & \tblbest{0.75} & \tblbest{0.81}
  & 0.60 & 0.73
  & \tblbest{0.73} & \tblbest{0.81}
  & 0.46 & 0.70
  & \tblbest{0.69} & \tblbest{0.73} \\
& Ensemble-I
  & 0.59 & 0.74
  & 0.74 & 0.80
  & 0.64 & \tblbest{0.77}
  & 0.68 & 0.78
  & \tblbest{0.49} & \tblbest{0.73}
  & 0.68 & 0.71 \\

\midrule

Llama 70B & AYS
  & 0.51 & 0.63
  & 0.53 & 0.64
  & 0.72 & 0.68
  & 0.53 & 0.62
  & 0.49 & 0.64
  & 0.70 & 0.69 \\
& IC-IDK
  & 0.48 & 0.60
  & \tblbestne{0.75} & 0.67
  & 0.13 & 0.53
  & 0.74 & 0.71
  & 0.62 & 0.68
  & 0.22 & 0.55 \\
\cmidrule(lr){2-14}
& DBA-A
  & \tblbestboth{0.92} & \tblbestboth{0.91}
  & 0.71 & 0.71
  & \tblbestne{0.89} & \tblbestboth{0.78}
  & \tblbestboth{0.84} & \tblbestboth{0.81}
  & 0.77 & 0.81
  & \tblbestne{0.86} & \tblbestboth{0.74} \\
& DBA-I
  & 0.84 & 0.84
  & 0.74 & \tblbestboth{0.76}
  & \tblbestne{0.89} & 0.77
  & 0.81 & 0.78
  & \tblbestboth{0.79} & \tblbestboth{0.83}
  & 0.84 & 0.71 \\
\cmidrule(lr){2-14}
& Ensemble-A
  & 0.89 & 0.87
  & 0.74 & 0.70
  & \tblbest{0.90} & 0.75
  & \tblbest{0.84} & 0.78
  & 0.77 & 0.82
  & \tblbest{0.88} & 0.72 \\
& Ensemble-I
  & 0.85 & 0.84
  & \tblbest{0.77} & 0.74
  & \tblbest{0.90} & 0.74
  & 0.81 & 0.75
  & 0.78 & 0.82
  & \tblbest{0.88} & 0.70 \\

\midrule

Qwen 8B & AYS
  & 0.73 & 0.66
  & 0.75 & 0.68
  & 0.88 & 0.71
  & 0.76 & 0.71
  & 0.70 & 0.68
  & 0.79 & 0.64 \\
& IC-IDK
  & 0.59 & 0.62
  & 0.76 & 0.61
  & 0.44 & 0.57
  & 0.62 & 0.65
  & 0.44 & 0.53
  & 0.54 & 0.62 \\
\cmidrule(lr){2-14}
& DBA-A
  & \tblbestne{0.93} & \tblbestboth{0.82}
  & \tblbestne{0.89} & \tblbestboth{0.73}
  & 0.93 & \tblbestboth{0.74}
  & \tblbestne{0.87} & \tblbestboth{0.76}
  & \tblbestne{0.87} & \tblbestboth{0.79}
  & \tblbestne{0.91} & \tblbestboth{0.65} \\
& DBA-I
  & 0.80 & 0.80
  & \tblbestne{0.89} & \tblbestboth{0.73}
  & \tblbestne{0.94} & 0.73
  & 0.84 & 0.70
  & 0.81 & 0.74
  & 0.89 & \tblbestboth{0.65} \\
\cmidrule(lr){2-14}
& Ensemble-A
  & \tblbest{0.94} & 0.80
  & \tblbest{0.91} & 0.67
  & 0.94 & 0.67
  & \tblbest{0.91} & 0.75
  & \tblbest{0.88} & 0.75
  & \tblbest{0.93} & 0.62 \\
& Ensemble-I
  & 0.90 & 0.76
  & \tblbest{0.91} & 0.70
  & \tblbest{0.95} & 0.69
  & 0.89 & 0.70
  & 0.85 & 0.69
  & 0.91 & 0.60 \\

\bottomrule
\end{tabular}
\vspace{-1.0em}
\caption{\small Per-model error-detection performance (F1 / AUROC). Prompting baselines (AYS, IC-IDK), \textsc{DBA}, and ensembles (AYS $\cup$ \textsc{DBA}). Within each model block and for each dataset/metric, \textbf{bold} indicates the best overall value (ties included) and \underline{underlined} indicates the best non-ensemble value among AYS/IC-IDK/DBA-A/DBA-I (ties included). If a non-ensemble method is also best overall, it is both bold and underlined. AUROC is reported as AUC.}
\label{tab:baseline_prf}
\vspace{-1.5em}
\end{table*}

\subsection{Results and Analysis}
\label{sec:results_and_analysis}


Table~\ref{tab:baseline_prf} presents error detection results for three
representative models spanning distinct parameter scales: GPT-5.1
(Frontier), Llama-3.3-70B (Large), and Qwen3-8B (Small). We observe
similar trends across the remaining models reported in Table \ref{tab:baseline_other_prf}. The results in both tables clearly demonstrate the efficacy
of \textsc{DBA}, showing its superior performance across nine distinct LLMs and six QA benchmarks.

 \textsc{DBA} outperforms the \textsc{AYS} baseline (in terms of F1) in 17 of the 18 evaluated model-dataset pairs. For the GPT-5.1 frontier LLM, \textsc{DBA-A} yields substantial gains on datasets where base consistency is moderate-to-high. We observe F1 improvements ranging from $+18$ to $+28$ points across \emph{Bamboogle}, \emph{HotpotQA}, and \emph{MuSiQue}. For Qwen3-8B, DBA delivers strong error detection performance across datasets, with F1 typically at or above $0.80$. For Llama-3.3-70B, DBA exceeds $0.80$ F1 on most datasets. In all cases, AUROC remains above random, ranging from the mid-$0.60s$ to low-$0.90s$. This demonstrates that the success of DBA does not merely stem from being more abstention-prone, showing that it effectively separates incorrect answers from correct ones.

\paragraph{DBA Improves Performance through Recall.} Results in Table \ref{tab:baseline_main_prf} and Table \ref{tab:baseline_other_prf} show that the main
driver of \textsc{DBA}’s F1 gains is a substantial increase in the recall
of incorrect Direct answers. Standard uncertainty baselines such as
\textsc{AYS} often exhibit overconfidence when the model hallucinates,
which leads to low recall and missed errors. \textsc{DBA} mitigates this
failure mode by treating cross-regime disagreement as a hard veto. For
open-source models and more challenging datasets, this strategy
typically increases recall by +30 to +50 points compared to
\textsc{AYS}. For example, on Bamboogle, recall for Llama-3.3-70B
improves from 37 to 90.

We also observe clear boundary conditions for frontier models such as
GPT-5.1. When base accuracy is high, as on \emph{Mintaka}, errors are rare and
disagreement signals are correspondingly sparse, which allows
\textsc{AYS} to remain competitive. Conversely, on complex datasets with
lower accuracy, such as \emph{CRAG}, frontier models can produce stable but
incorrect predictions across regimes. In these cases, \textsc{DBA}
recall decreases because consistent errors do not generate disagreement
signals that the method can exploit.

\paragraph{Complementary Failure Modes and Method Ensembling.}
To address cases where errors are consistent, we leverage an ensembling approach. This strategy combines the strengths of both paradigms: \textsc{AYS} detects stable-but-uncertain errors, while DBA flags confident-but-fragile hallucinations. The complementary nature of both approaches is most visible on GPT-5.1 when evaluated on \emph{CRAG}: while \textsc{AYS} outperforms \textsc{DBA-A} individually ($66$ vs.\ $53$ F1), their union achieves the highest overall performance ($75$ F1). This trend generalizes to other datasets like \emph{HotpotQA}, where ensembling boosts F1 to $73$.



\subsection{Efficiency and Deployment Overhead}

\begin{table}[t]
\centering
\small
\begin{tabular*}{\columnwidth}{l@{\extracolsep{\fill}}ccc}
\toprule
\textbf{Method} & \textbf{LM Calls} & \textbf{Total Tokens} & \textbf{Total$\times$} \\
\midrule
Direct & 1 & 833 & 1.00 \\
AYS & 2 & 1,054 & 1.26 \\
IC-IDK & 1 & 963 & 1.16 \\
DBA-A & 4 & 4,702 & 5.64 \\
DBA-I & 6.4 & 4,912 & 5.89 \\
Self-Consistency & 8 & 5,420 & 6.51 \\
\bottomrule
\end{tabular*}
\caption{Comparison of computational cost across abstention methods. We report the number of model calls and total token usage, along with relative cost ($\times$) normalized to Direct prompting.}
\vspace{-1em}
\label{tab:cost_main}
\end{table}
We evaluate the computational requirements of each method using call volume and token multipliers, averaged across all models and datasets (Table \ref{tab:cost_main}). IC-IDK requires a single model call, while AYS adds one verification turn for a total of two calls. IC-IDK's cost is also influenced by the length of in-context demonstrations, which vary across datasets.

Among the multi-call approaches, DBA-A and Self-Consistency (SC) allocate compute differently. DBA-A requires 4 calls and $5.64\times$ the total tokens of Direct prompting, whereas SC requires 8 calls and $6.51\times$. Notably, DBA-A generates $38\%$ fewer output tokens than SC, reflecting its focus on structural verification rather than producing multiple full-length candidate answers. Detailed token usage statistics, including a breakdown of input versus output overhead, are in Appendix \ref{sec:token_cost_appendix}.

While DBA-I issues one call per decomposition step, each hop carries only a small context, so its total token cost is only marginally higher than DBA-A despite the larger number of calls, and downstream performance is comparable. Overall, moving from lower-overhead methods such as AYS to DBA-A increases total token usage from $1.26\times$ to $5.64\times$ relative to Direct prompting, highlighting the additional cost of multi-call reliability mechanisms. However, this resource trade-off yields the significant gains in abstention accuracy and reliability metrics reported in Section \ref{sec:results_and_analysis}.

\section{Discussion}
\label{sec:discussion}


Our results reveal a persistent discrepancy between decomposed prompting regimes and direct question answering. If an LLM's reasoning were indeed consistent over stable internal knowledge, then Direct and Assistive answers would be expected to be
equivalent. Instead, even frontier LLMs exhibit only moderate
cross-prompt consistency, with agreement rates around 70\%. This
pattern suggests that LLM outputs strongly depend on the question decomposition rather than reflecting a consistent reasoning process over semantically equivalent plans. We present eight representative inconsistency cases for frontier LLMs in
Table~\ref{tab:additional_examples}, drawn from a manual error analysis of 100 random examples taken from all of our six evaluation datasets.

\paragraph{Potential for Answer Correction}

The analysis in Table~\ref{tab:inconsistency_breakdown_frontier} reveals a boundary condition where the efficacy of DBA is limited. Among the frontier-model disagreement cases summarized in Table~\ref{tab:inconsistency_breakdown_frontier}, 67\% fall into the \textit{Both} category, where both the Direct and Assistive regimes are wrong. This indicates that, for frontier models, DBA is highly effective at detecting ``reasoning shortcuts'' \citep{jiang-bansal-2019-avoiding}, but it cannot correct errors that stem from a lack of parametric knowledge.

\paragraph{Decomposition Hurts (Shortcut Hypothesis)}
Cases in which decomposition reduces performance suggest that correct
Direct answers may arise from reasoning shortcuts, as the LLM fails to reach an accurate result via a decomposed (step-by-step) execution. Example~4 in Table~\ref{tab:additional_examples} shows a case where the LLM produces the correct final answer under Direct prompting but fails to recover the required intermediate facts when executing the gold decomposition. Due to this being a fixed gold-standard decomposition, the plan is bound to be accurate. Hence, the inconsistency with the Direct answer, highlights the LLM's lack of the relevant intermediate knowledge required in the execution of the gold-standard plan.

\paragraph{Decomposition Helps (Scaffolding Hypothesis)}
Conversely, when decomposition helps performance the LLM appears to possess the knowledge of the requisite atomic facts but lacks the executive function to organize them zero-shot. In Example~2, the Direct model hallucinates the date, while the Assistive prompt successfully scaffolds the retrieval of the correct intermediate entity and final answer.


\section{Related Work}
\label{sec:related}

\paragraph{Decomposition and Planning.} Question decomposition has been formalized as a standalone question-understanding problem through QDMR and the BREAK benchmark \citep{wolfson-etal-2020-break}. More recent work has shown that smaller language models can be leveraged to generate and rank high-quality synthetic decompositions, and that fine-tuning on this data yields decomposition performance that is comparable to larger models under distribution shift \citep{han-gardent-2025-generating}. Traditional decomposition strategies function as generative interventions, modularizing execution to enhance performance on complex multi-hop queries \citep{khot2023decomposed,wolfson-etal-2022-weakly,zhou2023leasttomost,press-etal-2023-measuring,radhakrishnan2023questiondecompositionimprovesfaithfulness,wu2024_gendec}. Recent work has further leveraged these structures to improve reasoning faithfulness by enforcing structural constraints \citep{radhakrishnan2023questiondecompositionimprovesfaithfulness}. In contrast, we employ question decomposition as an experimental control: we fix a gold-standard decomposition and use it across multiple task-equivalent but structurally distinct prompting regimes, thereby decoupling planning from execution and using cross-prompt consistency as a lens on model reliability.

\paragraph{Uncertainty and Abstention.}
Our work relates to uncertainty estimation and selective prediction for deciding when a model should abstain. One approach elicits explicit self-evaluation signals, prompting models to estimate their own accuracy or internal knowledge \cite{kadavath2022languagemodelsmostlyknow}. Other methods train calibrated "I don't know" behavior through self-detection and reflection on diverse question variants \cite{zhao-etal-2024-knowing}. Additionally, verification pipelines like Chain-of-Verification utilize structured drafting and fact-checking stages to revise potential errors \cite{dhuliawala-etal-2024-chain}. These strategies generally treat uncertainty as a property of a single execution path. In contrast, we infer reliability behaviorally from the agreement across task-equivalent execution regimes, requiring no specialized supervision, auxiliary classifiers, or confidence heads.

\paragraph{Consistency as a Reliability Signal.}
Consistency across generations is a widely used indicator of correctness. Self-consistency decoding leverages majority agreement among sampled reasoning paths as a confidence heuristic \cite{wang2023selfconsistency}, which can be further refined by weighting semantic rationale similarity \cite{knappe2024enhancing}. Disagreement also helps flag hallucinations. Frameworks like SelfCheckGPT and SAC3 measure conflict among alternative continuations or perturbations \cite{manakul-etal-2023-selfcheckgpt, zhang-etal-2023-sac3}, while multi-agent systems treat inconsistencies during cross-examination as evidence of falsehood \cite{cohen-etal-2023-lm}. Beyond sampling, consistency under prompt variation serves as a core reliability axis that correlates with accuracy \cite{nalbandyan-etal-2025-score, novikova2025consistency}, and can be actively optimized  \cite{raj2025improving}. Unlike these methods based on sampling or paraphrasing, we evaluate consistency across structurally distinct yet task-equivalent reasoning interfaces. We demonstrate that disagreement between direct and decomposed regimes provides a simple, prompt-based signal that effectively tracks factual correctness.

\section{Conclusion}
\label{sec:conclusion}



In this work, we investigate the role of decomposed prompting through the lens of LLM reliability. By auditing semantically-equivalent reasoning paths across different model scales, we identify a scale-dependent shift in the efficacy of question decomposition. For weaker models, decomposition often functions as a scaffold that improves accuracy, while for frontier models these accuracy gains plateau. At the same time, LLM consistency between different prompting regimes serves as a strong signal for LLM performance.

We show that LLM inconsistency between Direct answering and decomposed prompting provides a reliable indicator of factual errors across different LLMs and outperforms self-reported confidence. Leveraging this insight, we introduce Disagreement-Based Abstention (\textsc{DBA}), a training-free approach which substantially improves error detection compared to standard uncertainty methods. Ultimately, our findings suggest that as LLMs internalize more reasoning capability, the marginal benefit of decomposed prompting shifts from expanding \textit{what can be answered} by the LLM to auditing its confidence of \textit{what is known}.

\section*{Limitations}
\label{sec:limitations}




We highlight three main limitations of our analysis and proposed
\textsc{DBA} method. First, \textsc{DBA} can only detect errors that
manifest as disagreement across prompting regimes. When a model
reproduces the same incorrect answer under Direct, Assistive, and
Incremental prompting, \textsc{DBA} treats that answer as stable and
provides no corrective signal.

Second, \textsc{DBA} relies on access to high-quality decompositions
expressed in our DSL. In the main experiments, we use decompositions
that are manually verified or produced by a strong teacher model, so
our reported setup still assumes access to reliable plans. This makes
the method less self-contained, especially for weaker models, which
often struggle to generate good multi-hop decompositions on their own.
However, this limitation is less severe for stronger models; as shown
in Appendix~\ref{app:auto-decomp-feasibility}, a 70B-scale open model
can generate usable decompositions in most manually audited cases.

Third, \textsc{DBA} incurs additional computational cost and latency,
since it requires at least one decomposed execution in addition to the
Direct call. Incremental prompting further increases this cost by
introducing one model invocation per hop. In contrast, baselines such as
\textsc{AYS} and \textsc{IC-IDK} operate within a single prompting
regime and are therefore cheaper to deploy, although they provide weaker
error-detection performance.

\section*{Ethics Statement}
\label{sec:ethics}
Our study evaluates existing, publicly available QA benchmarks and therefore inherits any biases or limitations in those datasets. We do not collect new data, solicit human subjects, or intentionally include personally identifying information beyond what may already be present in the original benchmarks. Our method is training-free and intended for research on hallucination detection; it does not guarantee correctness and may abstain unevenly across topics or domains. Because it requires additional model calls, it increases inference cost/latency, which should be considered in any deployment.

\section*{Acknowledgements}
\label{sec:acknowledgement}
This research has been supported in part by the ONR Contract N00014-23-1-2364, and conducted as a collaborative effort between \textit{Arizona State University} and the \textit{University of Pennsylvania}. We gratefully acknowledge the \textit{Complex Data Analysis and Reasoning Lab} at the \textit{School of Computing and Augmented Intelligence}, \textit{Arizona State University}, and the \textit{Cognitive Computation Group}, \textit{University of Pennsylvania}, for providing computational resources and institutional support. We also thank the anonymous reviewers for their constructive feedback and valuable suggestions.

\bibliography{anthology, custom}

@inproceedings{press-etal-2023-measuring,
    title = "Measuring and Narrowing the Compositionality Gap in Language Models",
    author = "Press, Ofir  and
      Zhang, Muru  and
      Min, Sewon  and
      Schmidt, Ludwig  and
      Smith, Noah  and
      Lewis, Mike",
    editor = "Bouamor, Houda  and
      Pino, Juan  and
      Bali, Kalika",
    booktitle = "Findings of the Association for Computational Linguistics: EMNLP 2023",
    month = dec,
    year = "2023",
    address = "Singapore",
    publisher = "Association for Computational Linguistics",
    url = "https://aclanthology.org/2023.findings-emnlp.378/",
    doi = "10.18653/v1/2023.findings-emnlp.378",
    pages = "5687--5711"
}

@inproceedings{el-asri-etal-2017-frames,
    title = "{F}rames: a corpus for adding memory to goal-oriented dialogue systems",
    author = "El Asri, Layla  and
      Schulz, Hannes  and
      Sharma, Shikhar  and
      Zumer, Jeremie  and
      Harris, Justin  and
      Fine, Emery  and
      Mehrotra, Rahul  and
      Suleman, Kaheer",
    editor = "Jokinen, Kristiina  and
      Stede, Manfred  and
      DeVault, David  and
      Louis, Annie",
    booktitle = "Proceedings of the 18th Annual {SIG}dial Meeting on Discourse and Dialogue",
    month = aug,
    year = "2017",
    address = {Saarbr{\"u}cken, Germany},
    publisher = "Association for Computational Linguistics",
    url = "https://aclanthology.org/W17-5526/",
    doi = "10.18653/v1/W17-5526",
    pages = "207--219"
}

@article{trivedi-etal-2022-musique,
    title = "{M}u{S}i{Q}ue: Multihop Questions via Single-hop Question Composition",
    author = "Trivedi, Harsh  and
      Balasubramanian, Niranjan  and
      Khot, Tushar  and
      Sabharwal, Ashish",
    editor = "Roark, Brian  and
      Nenkova, Ani",
    journal = "Transactions of the Association for Computational Linguistics",
    volume = "10",
    year = "2022",
    address = "Cambridge, MA",
    publisher = "MIT Press",
    url = "https://aclanthology.org/2022.tacl-1.31/",
    doi = "10.1162/tacl_a_00475",
    pages = "539--554"
}

@inproceedings{yang-etal-2018-hotpotqa,
    title = "{H}otpot{QA}: A Dataset for Diverse, Explainable Multi-hop Question Answering",
    author = "Yang, Zhilin  and
      Qi, Peng  and
      Zhang, Saizheng  and
      Bengio, Yoshua  and
      Cohen, William  and
      Salakhutdinov, Ruslan  and
      Manning, Christopher D.",
    editor = "Riloff, Ellen  and
      Chiang, David  and
      Hockenmaier, Julia  and
      Tsujii, Jun{'}ichi",
    booktitle = "Proceedings of the 2018 Conference on Empirical Methods in Natural Language Processing",
    month = oct # "-" # nov,
    year = "2018",
    address = "Brussels, Belgium",
    publisher = "Association for Computational Linguistics",
    url = "https://aclanthology.org/D18-1259/",
    doi = "10.18653/v1/D18-1259",
    pages = "2369--2380"
}

@inproceedings{sen-etal-2022-mintaka,
    title = "Mintaka: A Complex, Natural, and Multilingual Dataset for End-to-End Question Answering",
    author = "Sen, Priyanka  and
      Aji, Alham Fikri  and
      Saffari, Amir",
    editor = "Calzolari, Nicoletta  and
      Huang, Chu-Ren  and
      Kim, Hansaem  and
      Pustejovsky, James  and
      Wanner, Leo  and
      Choi, Key-Sun  and
      Ryu, Pum-Mo  and
      Chen, Hsin-Hsi  and
      Donatelli, Lucia  and
      Ji, Heng  and
      Kurohashi, Sadao  and
      Paggio, Patrizia  and
      Xue, Nianwen  and
      Kim, Seokhwan  and
      Hahm, Younggyun  and
      He, Zhong  and
      Lee, Tony Kyungil  and
      Santus, Enrico  and
      Bond, Francis  and
      Na, Seung-Hoon",
    booktitle = "Proceedings of the 29th International Conference on Computational Linguistics",
    month = oct,
    year = "2022",
    address = "Gyeongju, Republic of Korea",
    publisher = "International Committee on Computational Linguistics",
    url = "https://aclanthology.org/2022.coling-1.138/",
    pages = "1604--1619"
 }

@inproceedings{cohen-etal-2023-lm,
    title = "{LM} vs {LM}: Detecting Factual Errors via Cross Examination",
    author = "Cohen, Roi  and
      Hamri, May  and
      Geva, Mor  and
      Globerson, Amir",
    editor = "Bouamor, Houda  and
      Pino, Juan  and
      Bali, Kalika",
    booktitle = "Proceedings of the 2023 Conference on Empirical Methods in Natural Language Processing",
    month = dec,
    year = "2023",
    address = "Singapore",
    publisher = "Association for Computational Linguistics",
    url = "https://aclanthology.org/2023.emnlp-main.778/",
    doi = "10.18653/v1/2023.emnlp-main.778",
    pages = "12621--12640"
}

@inproceedings{zhang-etal-2023-sac3,
    title = "{SAC}$^3$: Reliable Hallucination Detection in Black-Box Language Models via Semantic-aware Cross-check Consistency",
    author = "Zhang, Jiaxin  and
      Li, Zhuohang  and
      Das, Kamalika  and
      Malin, Bradley  and
      Kumar, Sricharan",
    editor = "Bouamor, Houda  and
      Pino, Juan  and
      Bali, Kalika",
    booktitle = "Findings of the Association for Computational Linguistics: EMNLP 2023",
    month = dec,
    year = "2023",
    address = "Singapore",
    publisher = "Association for Computational Linguistics",
    url = "https://aclanthology.org/2023.findings-emnlp.1032/",
    doi = "10.18653/v1/2023.findings-emnlp.1032",
    pages = "15445--15458"
}

@inproceedings{manakul-etal-2023-selfcheckgpt,
    title = "{S}elf{C}heck{GPT}: Zero-Resource Black-Box Hallucination Detection for Generative Large Language Models",
    author = "Manakul, Potsawee  and
      Liusie, Adian  and
      Gales, Mark",
    editor = "Bouamor, Houda  and
      Pino, Juan  and
      Bali, Kalika",
    booktitle = "Proceedings of the 2023 Conference on Empirical Methods in Natural Language Processing",
    month = dec,
    year = "2023",
    address = "Singapore",
    publisher = "Association for Computational Linguistics",
    url = "https://aclanthology.org/2023.emnlp-main.557/",
    doi = "10.18653/v1/2023.emnlp-main.557",
    pages = "9004--9017"
}

@inproceedings{zhao-etal-2024-knowing,
    title = "Knowing What {LLM}s {DO} {NOT} Know: A Simple Yet Effective Self-Detection Method",
    author = "Zhao, Yukun  and
      Yan, Lingyong  and
      Sun, Weiwei  and
      Xing, Guoliang  and
      Meng, Chong  and
      Wang, Shuaiqiang  and
      Cheng, Zhicong  and
      Ren, Zhaochun  and
      Yin, Dawei",
    editor = "Duh, Kevin  and
      Gomez, Helena  and
      Bethard, Steven",
    booktitle = "Proceedings of the 2024 Conference of the North American Chapter of the Association for Computational Linguistics: Human Language Technologies (Volume 1: Long Papers)",
    month = jun,
    year = "2024",
    address = "Mexico City, Mexico",
    publisher = "Association for Computational Linguistics",
    url = "https://aclanthology.org/2024.naacl-long.390/",
    doi = "10.18653/v1/2024.naacl-long.390",
    pages = "7051--7063"
}

@inproceedings{dhuliawala-etal-2024-chain,
    title = "Chain-of-Verification Reduces Hallucination in Large Language Models",
    author = "Dhuliawala, Shehzaad  and
      Komeili, Mojtaba  and
      Xu, Jing  and
      Raileanu, Roberta  and
      Li, Xian  and
      Celikyilmaz, Asli  and
      Weston, Jason",
    editor = "Ku, Lun-Wei  and
      Martins, Andre  and
      Srikumar, Vivek",
    booktitle = "Findings of the Association for Computational Linguistics: ACL 2024",
    month = aug,
    year = "2024",
    address = "Bangkok, Thailand",
    publisher = "Association for Computational Linguistics",
    url = "https://aclanthology.org/2024.findings-acl.212/",
    doi = "10.18653/v1/2024.findings-acl.212",
    pages = "3563--3578"
}

@inproceedings{nalbandyan-etal-2025-score,
    title = "{SCORE}: Systematic {CO}nsistency and Robustness Evaluation for Large Language Models",
    author = "Nalbandyan, Grigor  and
      Shahbazyan, Rima  and
      Bakhturina, Evelina",
    editor = "Chen, Weizhu  and
      Yang, Yi  and
      Kachuee, Mohammad  and
      Fu, Xue-Yong",
    booktitle = "Proceedings of the 2025 Conference of the Nations of the Americas Chapter of the Association for Computational Linguistics: Human Language Technologies (Volume 3: Industry Track)",
    month = apr,
    year = "2025",
    address = "Albuquerque, New Mexico",
    publisher = "Association for Computational Linguistics",
    url = "https://aclanthology.org/2025.naacl-industry.39/",
    doi = "10.18653/v1/2025.naacl-industry.39",
    pages = "470--484",
    ISBN = "979-8-89176-194-0"
}

@inproceedings{jiang-bansal-2019-avoiding,
    title = "Avoiding Reasoning Shortcuts: Adversarial Evaluation, Training, and Model Development for Multi-Hop {QA}",
    author = "Jiang, Yichen  and
      Bansal, Mohit",
    editor = "Korhonen, Anna  and
      Traum, David  and
      M{\`a}rquez, Llu{\'i}s",
    booktitle = "Proceedings of the 57th Annual Meeting of the Association for Computational Linguistics",
    month = jul,
    year = "2019",
    address = "Florence, Italy",
    publisher = "Association for Computational Linguistics",
    url = "https://aclanthology.org/P19-1262/",
    doi = "10.18653/v1/P19-1262",
    pages = "2726--2736"
}

@article{
    raj2025improving,
    title={Improving Consistency in Large Language Models through Chain of Guidance},
    author={Harsh Raj and Vipul Gupta and Domenic Rosati and Subhabrata Majumdar},
    journal={Transactions on Machine Learning Research},
    issn={2835-8856},
    year={2025},
    url={https://openreview.net/forum?id=asiBW1bB9b},
    note={}
}

@inproceedings{
yang2024crag,
title={{CRAG} - Comprehensive {RAG} Benchmark},
author={Xiao Yang and Kai Sun and Hao Xin and Yushi Sun and Nikita Bhalla and Xiangsen Chen and Sajal Choudhary and Rongze Gui and Ziran Jiang and Ziyu JIANG and Lingkun Kong and Brian Moran and Jiaqi Wang and Yifan Ethan Xu and An Yan and Chenyu Yang and Eting Yuan and Hanwen Zha and Nan Tang and Lei Chen and Nicolas SCHEFFER and Yue Liu and Nirav Shah and Rakesh Wanga and Anuj Kumar and Wen-tau Yih and Xin Luna Dong},
booktitle={The Thirty-eight Conference on Neural Information Processing Systems Datasets and Benchmarks Track},
year={2024},
url={https://openreview.net/forum?id=Q7lAqY41HH}
}

@inproceedings{
zhou2023leasttomost,
title={Least-to-Most Prompting Enables Complex Reasoning in Large Language Models},
author={Denny Zhou and Nathanael Sch{\"a}rli and Le Hou and Jason Wei and Nathan Scales and Xuezhi Wang and Dale Schuurmans and Claire Cui and Olivier Bousquet and Quoc V Le and Ed H. Chi},
booktitle={The Eleventh International Conference on Learning Representations },
year={2023},
url={https://openreview.net/forum?id=WZH7099tgfM}
}

@inproceedings{wolfson-etal-2022-weakly,
    title = "Weakly Supervised Text-to-{SQL} Parsing through Question Decomposition",
    author = "Wolfson, Tomer  and
      Deutch, Daniel  and
      Berant, Jonathan",
    editor = "Carpuat, Marine  and
      de Marneffe, Marie-Catherine  and
      Meza Ruiz, Ivan Vladimir",
    booktitle = "Findings of the Association for Computational Linguistics: NAACL 2022",
    month = jul,
    year = "2022",
    address = "Seattle, United States",
    publisher = "Association for Computational Linguistics",
    url = "https://aclanthology.org/2022.findings-naacl.193/",
    doi = "10.18653/v1/2022.findings-naacl.193",
    pages = "2528--2542"
}

@inproceedings{
wang2023selfconsistency,
title={Self-Consistency Improves Chain of Thought Reasoning in Language Models},
author={Xuezhi Wang and Jason Wei and Dale Schuurmans and Quoc V Le and Ed H. Chi and Sharan Narang and Aakanksha Chowdhery and Denny Zhou},
booktitle={The Eleventh International Conference on Learning Representations },
year={2023},
url={https://openreview.net/forum?id=1PL1NIMMrw}
}

@misc{radhakrishnan2023questiondecompositionimprovesfaithfulness,
      title={Question Decomposition Improves the Faithfulness of Model-Generated Reasoning}, 
      author={Ansh Radhakrishnan and Karina Nguyen and Anna Chen and Carol Chen and Carson Denison and Danny Hernandez and Esin Durmus and Evan Hubinger and Jackson Kernion and Kamilė Lukošiūtė and Newton Cheng and Nicholas Joseph and Nicholas Schiefer and Oliver Rausch and Sam McCandlish and Sheer El Showk and Tamera Lanham and Tim Maxwell and Venkatesa Chandrasekaran and Zac Hatfield-Dodds and Jared Kaplan and Jan Brauner and Samuel R. Bowman and Ethan Perez},
      year={2023},
      eprint={2307.11768},
      archivePrefix={arXiv},
      primaryClass={cs.CL},
      url={https://arxiv.org/abs/2307.11768}, 
}

@misc{sinha2025illusiondiminishingreturnsmeasuring,
      title={The Illusion of Diminishing Returns: Measuring Long Horizon Execution in LLMs}, 
      author={Akshit Sinha and Arvindh Arun and Shashwat Goel and Steffen Staab and Jonas Geiping},
      year={2025},
      eprint={2509.09677},
      archivePrefix={arXiv},
      primaryClass={cs.AI},
      url={https://arxiv.org/abs/2509.09677}, 
}

@inproceedings{simhi-etal-2025-trust,
    title = "Trust Me, {I}{'}m Wrong: {LLM}s Hallucinate with Certainty Despite Knowing the Answer",
    author = "Simhi, Adi  and
      Itzhak, Itay  and
      Barez, Fazl  and
      Stanovsky, Gabriel  and
      Belinkov, Yonatan",
    editor = "Christodoulopoulos, Christos  and
      Chakraborty, Tanmoy  and
      Rose, Carolyn  and
      Peng, Violet",
    booktitle = "Findings of the Association for Computational Linguistics: EMNLP 2025",
    month = nov,
    year = "2025",
    address = "Suzhou, China",
    publisher = "Association for Computational Linguistics",
    url = "https://aclanthology.org/2025.findings-emnlp.792/",
    doi = "10.18653/v1/2025.findings-emnlp.792",
    pages = "14665--14688",
    ISBN = "979-8-89176-335-7",
}

@article{wolfson-etal-2026-monaco,
    title = "{M}o{N}a{C}o: More Natural and Complex Questions for Reasoning Across Dozens of Documents",
    author = "Wolfson, Tomer  and
      Trivedi, Harsh  and
      Geva, Mor  and
      Goldberg, Yoav  and
      Roth, Dan  and
      Khot, Tushar  and
      Sabharwal, Ashish  and
      Tsarfaty, Reut",
    journal = "Transactions of the Association for Computational Linguistics",
    volume = "14",
    year = "2026",
    address = "Cambridge, MA",
    publisher = "MIT Press",
    url = "https://aclanthology.org/2026.tacl-1.2/",
    doi = "10.1162/tacl.a.64",
    pages = "23--46",
}

@misc{jiang2023mistral7b,
  title         = {Mistral 7B},
  author        = {Jiang, Albert Q. and Sablayrolles, Alexandre and Mensch, Arthur and Bamford, Chris and Chaplot, Devendra Singh and de las Casas, Diego and Bressand, Florian and Lengyel, Gianna and Lample, Guillaume and Saulnier, Lucile and Renard Lavaud, L{\'e}lio and Lachaux, Marie-Anne and Stock, Pierre and Le Scao, Teven and Lavril, Thibaut and Wang, Thomas and Lacroix, Timoth{\'e}e and El Sayed, William},
  year          = {2023},
  eprint        = {2310.06825},
  archivePrefix = {arXiv},
  primaryClass  = {cs.CL},
  url           = {https://arxiv.org/abs/2310.06825}
}

@misc{grattafiori2024llama3herdmodels,
      title={The Llama 3 Herd of Models}, 
      author={Aaron Grattafiori and Abhimanyu Dubey and Abhinav Jauhri and Abhinav Pandey and Abhishek Kadian and Ahmad Al-Dahle and Aiesha Letman and Akhil Mathur and Alan Schelten and Alex Vaughan and Amy Yang and Angela Fan and Anirudh Goyal and Anthony Hartshorn and Aobo Yang and Archi Mitra and Archie Sravankumar and Artem Korenev and Arthur Hinsvark and Arun Rao and Aston Zhang and Aurelien Rodriguez and Austen Gregerson and Ava Spataru and Baptiste Roziere and Bethany Biron and Binh Tang and Bobbie Chern and Charlotte Caucheteux and Chaya Nayak and Chloe Bi and Chris Marra and Chris McConnell and Christian Keller and Christophe Touret and Chunyang Wu and Corinne Wong and Cristian Canton Ferrer and Cyrus Nikolaidis and Damien Allonsius and Daniel Song and Danielle Pintz and Danny Livshits and Danny Wyatt and David Esiobu and Dhruv Choudhary and Dhruv Mahajan and Diego Garcia-Olano and Diego Perino and Dieuwke Hupkes and Egor Lakomkin and Ehab AlBadawy and Elina Lobanova and Emily Dinan and Eric Michael Smith and Filip Radenovic and Francisco Guzmán and Frank Zhang and Gabriel Synnaeve and Gabrielle Lee and Georgia Lewis Anderson and Govind Thattai and Graeme Nail and Gregoire Mialon and Guan Pang and Guillem Cucurell and Hailey Nguyen and Hannah Korevaar and Hu Xu and Hugo Touvron and Iliyan Zarov and Imanol Arrieta Ibarra and Isabel Kloumann and Ishan Misra and Ivan Evtimov and Jack Zhang and Jade Copet and Jaewon Lee and Jan Geffert and Jana Vranes and Jason Park and Jay Mahadeokar and Jeet Shah and Jelmer van der Linde and Jennifer Billock and Jenny Hong and Jenya Lee and Jeremy Fu and Jianfeng Chi and Jianyu Huang and Jiawen Liu and Jie Wang and Jiecao Yu and Joanna Bitton and Joe Spisak and Jongsoo Park and Joseph Rocca and Joshua Johnstun and Joshua Saxe and Junteng Jia and Kalyan Vasuden Alwala and Karthik Prasad and Kartikeya Upasani and Kate Plawiak and Ke Li and Kenneth Heafield and Kevin Stone and Khalid El-Arini and Krithika Iyer and Kshitiz Malik and Kuenley Chiu and Kunal Bhalla and Kushal Lakhotia and Lauren Rantala-Yeary and Laurens van der Maaten and Lawrence Chen and Liang Tan and Liz Jenkins and Louis Martin and Lovish Madaan and Lubo Malo and Lukas Blecher and Lukas Landzaat and Luke de Oliveira and Madeline Muzzi and Mahesh Pasupuleti and Mannat Singh and Manohar Paluri and Marcin Kardas and Maria Tsimpoukelli and Mathew Oldham and Mathieu Rita and Maya Pavlova and Melanie Kambadur and Mike Lewis and Min Si and Mitesh Kumar Singh and Mona Hassan and Naman Goyal and Narjes Torabi and Nikolay Bashlykov and Nikolay Bogoychev and Niladri Chatterji and Ning Zhang and Olivier Duchenne and Onur Çelebi and Patrick Alrassy and Pengchuan Zhang and Pengwei Li and Petar Vasic and Peter Weng and Prajjwal Bhargava and Pratik Dubal and Praveen Krishnan and Punit Singh Koura and Puxin Xu and Qing He and Qingxiao Dong and Ragavan Srinivasan and Raj Ganapathy and Ramon Calderer and Ricardo Silveira Cabral and Robert Stojnic and Roberta Raileanu and Rohan Maheswari and Rohit Girdhar and Rohit Patel and Romain Sauvestre and Ronnie Polidoro and Roshan Sumbaly and Ross Taylor and Ruan Silva and Rui Hou and Rui Wang and Saghar Hosseini and Sahana Chennabasappa and Sanjay Singh and Sean Bell and Seohyun Sonia Kim and Sergey Edunov and Shaoliang Nie and Sharan Narang and Sharath Raparthy and Sheng Shen and Shengye Wan and Shruti Bhosale and Shun Zhang and Simon Vandenhende and Soumya Batra and Spencer Whitman and Sten Sootla and Stephane Collot and Suchin Gururangan and Sydney Borodinsky and Tamar Herman and Tara Fowler and Tarek Sheasha and Thomas Georgiou and Thomas Scialom and Tobias Speckbacher and Todor Mihaylov and Tong Xiao and Ujjwal Karn and Vedanuj Goswami and Vibhor Gupta and Vignesh Ramanathan and Viktor Kerkez and Vincent Gonguet and Virginie Do and Vish Vogeti and Vítor Albiero and Vladan Petrovic and Weiwei Chu and Wenhan Xiong and Wenyin Fu and Whitney Meers and Xavier Martinet and Xiaodong Wang and Xiaofang Wang and Xiaoqing Ellen Tan and Xide Xia and Xinfeng Xie and Xuchao Jia and Xuewei Wang and Yaelle Goldschlag and Yashesh Gaur and Yasmine Babaei and Yi Wen and Yiwen Song and Yuchen Zhang and Yue Li and Yuning Mao and Zacharie Delpierre Coudert and Zheng Yan and Zhengxing Chen and Zoe Papakipos and Aaditya Singh and Aayushi Srivastava and Abha Jain and Adam Kelsey and Adam Shajnfeld and Adithya Gangidi and Adolfo Victoria and Ahuva Goldstand and Ajay Menon and Ajay Sharma and Alex Boesenberg and Alexei Baevski and Allie Feinstein and Amanda Kallet and Amit Sangani and Amos Teo and Anam Yunus and Andrei Lupu and Andres Alvarado and Andrew Caples and Andrew Gu and Andrew Ho and Andrew Poulton and Andrew Ryan and Ankit Ramchandani and Annie Dong and Annie Franco and Anuj Goyal and Aparajita Saraf and Arkabandhu Chowdhury and Ashley Gabriel and Ashwin Bharambe and Assaf Eisenman and Azadeh Yazdan and Beau James and Ben Maurer and Benjamin Leonhardi and Bernie Huang and Beth Loyd and Beto De Paola and Bhargavi Paranjape and Bing Liu and Bo Wu and Boyu Ni and Braden Hancock and Bram Wasti and Brandon Spence and Brani Stojkovic and Brian Gamido and Britt Montalvo and Carl Parker and Carly Burton and Catalina Mejia and Ce Liu and Changhan Wang and Changkyu Kim and Chao Zhou and Chester Hu and Ching-Hsiang Chu and Chris Cai and Chris Tindal and Christoph Feichtenhofer and Cynthia Gao and Damon Civin and Dana Beaty and Daniel Kreymer and Daniel Li and David Adkins and David Xu and Davide Testuggine and Delia David and Devi Parikh and Diana Liskovich and Didem Foss and Dingkang Wang and Duc Le and Dustin Holland and Edward Dowling and Eissa Jamil and Elaine Montgomery and Eleonora Presani and Emily Hahn and Emily Wood and Eric-Tuan Le and Erik Brinkman and Esteban Arcaute and Evan Dunbar and Evan Smothers and Fei Sun and Felix Kreuk and Feng Tian and Filippos Kokkinos and Firat Ozgenel and Francesco Caggioni and Frank Kanayet and Frank Seide and Gabriela Medina Florez and Gabriella Schwarz and Gada Badeer and Georgia Swee and Gil Halpern and Grant Herman and Grigory Sizov and Guangyi and Zhang and Guna Lakshminarayanan and Hakan Inan and Hamid Shojanazeri and Han Zou and Hannah Wang and Hanwen Zha and Haroun Habeeb and Harrison Rudolph and Helen Suk and Henry Aspegren and Hunter Goldman and Hongyuan Zhan and Ibrahim Damlaj and Igor Molybog and Igor Tufanov and Ilias Leontiadis and Irina-Elena Veliche and Itai Gat and Jake Weissman and James Geboski and James Kohli and Janice Lam and Japhet Asher and Jean-Baptiste Gaya and Jeff Marcus and Jeff Tang and Jennifer Chan and Jenny Zhen and Jeremy Reizenstein and Jeremy Teboul and Jessica Zhong and Jian Jin and Jingyi Yang and Joe Cummings and Jon Carvill and Jon Shepard and Jonathan McPhie and Jonathan Torres and Josh Ginsburg and Junjie Wang and Kai Wu and Kam Hou U and Karan Saxena and Kartikay Khandelwal and Katayoun Zand and Kathy Matosich and Kaushik Veeraraghavan and Kelly Michelena and Keqian Li and Kiran Jagadeesh and Kun Huang and Kunal Chawla and Kyle Huang and Lailin Chen and Lakshya Garg and Lavender A and Leandro Silva and Lee Bell and Lei Zhang and Liangpeng Guo and Licheng Yu and Liron Moshkovich and Luca Wehrstedt and Madian Khabsa and Manav Avalani and Manish Bhatt and Martynas Mankus and Matan Hasson and Matthew Lennie and Matthias Reso and Maxim Groshev and Maxim Naumov and Maya Lathi and Meghan Keneally and Miao Liu and Michael L. Seltzer and Michal Valko and Michelle Restrepo and Mihir Patel and Mik Vyatskov and Mikayel Samvelyan and Mike Clark and Mike Macey and Mike Wang and Miquel Jubert Hermoso and Mo Metanat and Mohammad Rastegari and Munish Bansal and Nandhini Santhanam and Natascha Parks and Natasha White and Navyata Bawa and Nayan Singhal and Nick Egebo and Nicolas Usunier and Nikhil Mehta and Nikolay Pavlovich Laptev and Ning Dong and Norman Cheng and Oleg Chernoguz and Olivia Hart and Omkar Salpekar and Ozlem Kalinli and Parkin Kent and Parth Parekh and Paul Saab and Pavan Balaji and Pedro Rittner and Philip Bontrager and Pierre Roux and Piotr Dollar and Polina Zvyagina and Prashant Ratanchandani and Pritish Yuvraj and Qian Liang and Rachad Alao and Rachel Rodriguez and Rafi Ayub and Raghotham Murthy and Raghu Nayani and Rahul Mitra and Rangaprabhu Parthasarathy and Raymond Li and Rebekkah Hogan and Robin Battey and Rocky Wang and Russ Howes and Ruty Rinott and Sachin Mehta and Sachin Siby and Sai Jayesh Bondu and Samyak Datta and Sara Chugh and Sara Hunt and Sargun Dhillon and Sasha Sidorov and Satadru Pan and Saurabh Mahajan and Saurabh Verma and Seiji Yamamoto and Sharadh Ramaswamy and Shaun Lindsay and Shaun Lindsay and Sheng Feng and Shenghao Lin and Shengxin Cindy Zha and Shishir Patil and Shiva Shankar and Shuqiang Zhang and Shuqiang Zhang and Sinong Wang and Sneha Agarwal and Soji Sajuyigbe and Soumith Chintala and Stephanie Max and Stephen Chen and Steve Kehoe and Steve Satterfield and Sudarshan Govindaprasad and Sumit Gupta and Summer Deng and Sungmin Cho and Sunny Virk and Suraj Subramanian and Sy Choudhury and Sydney Goldman and Tal Remez and Tamar Glaser and Tamara Best and Thilo Koehler and Thomas Robinson and Tianhe Li and Tianjun Zhang and Tim Matthews and Timothy Chou and Tzook Shaked and Varun Vontimitta and Victoria Ajayi and Victoria Montanez and Vijai Mohan and Vinay Satish Kumar and Vishal Mangla and Vlad Ionescu and Vlad Poenaru and Vlad Tiberiu Mihailescu and Vladimir Ivanov and Wei Li and Wenchen Wang and Wenwen Jiang and Wes Bouaziz and Will Constable and Xiaocheng Tang and Xiaojian Wu and Xiaolan Wang and Xilun Wu and Xinbo Gao and Yaniv Kleinman and Yanjun Chen and Ye Hu and Ye Jia and Ye Qi and Yenda Li and Yilin Zhang and Ying Zhang and Yossi Adi and Youngjin Nam and Yu and Wang and Yu Zhao and Yuchen Hao and Yundi Qian and Yunlu Li and Yuzi He and Zach Rait and Zachary DeVito and Zef Rosnbrick and Zhaoduo Wen and Zhenyu Yang and Zhiwei Zhao and Zhiyu Ma},
      year={2024},
      eprint={2407.21783},
      archivePrefix={arXiv},
      primaryClass={cs.AI},
      url={https://arxiv.org/abs/2407.21783}, 
}

@misc{yang2025qwen3,
  title={Qwen3 Technical Report}, 
      author={An Yang and Anfeng Li and Baosong Yang and Beichen Zhang and Binyuan Hui and Bo Zheng and Bowen Yu and Chang Gao and Chengen Huang and Chenxu Lv and Chujie Zheng and Dayiheng Liu and Fan Zhou and Fei Huang and Feng Hu and Hao Ge and Haoran Wei and Huan Lin and Jialong Tang and Jian Yang and Jianhong Tu and Jianwei Zhang and Jianxin Yang and Jiaxi Yang and Jing Zhou and Jingren Zhou and Junyang Lin and Kai Dang and Keqin Bao and Kexin Yang and Le Yu and Lianghao Deng and Mei Li and Mingfeng Xue and Mingze Li and Pei Zhang and Peng Wang and Qin Zhu and Rui Men and Ruize Gao and Shixuan Liu and Shuang Luo and Tianhao Li and Tianyi Tang and Wenbiao Yin and Xingzhang Ren and Xinyu Wang and Xinyu Zhang and Xuancheng Ren and Yang Fan and Yang Su and Yichang Zhang and Yinger Zhang and Yu Wan and Yuqiong Liu and Zekun Wang and Zeyu Cui and Zhenru Zhang and Zhipeng Zhou and Zihan Qiu},
      year={2025},
      eprint={2505.09388},
      archivePrefix={arXiv},
      primaryClass={cs.CL},
      url={https://arxiv.org/abs/2505.09388}, 
}

@misc{qwen2024qwen25,
  title={Qwen2.5 Technical Report}, 
      author={Qwen: An Yang and Baosong Yang and Beichen Zhang and Binyuan Hui and Bo Zheng and Bowen Yu and Chengyuan Li and Dayiheng Liu and Fei Huang and Haoran Wei and Huan Lin and Jian Yang and Jianhong Tu and Jianwei Zhang and Jianxin Yang and Jiaxi Yang and Jingren Zhou and Junyang Lin and Kai Dang and Keming Lu and Keqin Bao and Kexin Yang and Le Yu and Mei Li and Mingfeng Xue and Pei Zhang and Qin Zhu and Rui Men and Runji Lin and Tianhao Li and Tianyi Tang and Tingyu Xia and Xingzhang Ren and Xuancheng Ren and Yang Fan and Yang Su and Yichang Zhang and Yu Wan and Yuqiong Liu and Zeyu Cui and Zhenru Zhang and Zihan Qiu},
      year={2025},
      eprint={2412.15115},
      archivePrefix={arXiv},
      primaryClass={cs.CL},
      url={https://arxiv.org/abs/2412.15115}, 
}

@misc{geminiteam2025gemini25,
  title={Gemini 2.5: Pushing the Frontier with Advanced Reasoning, Multimodality, Long Context, and Next Generation Agentic Capabilities}, 
      author={Gheorghe Comanici and Eric Bieber and Mike Schaekermann and Ice Pasupat and Noveen Sachdeva and Inderjit Dhillon and Marcel Blistein and Ori Ram and Dan Zhang and Evan Rosen and Luke Marris and Sam Petulla and Colin Gaffney and Asaf Aharoni and Nathan Lintz and Tiago Cardal Pais and Henrik Jacobsson and Idan Szpektor and Nan-Jiang Jiang and Krishna Haridasan and Ahmed Omran and Nikunj Saunshi and Dara Bahri and Gaurav Mishra and Eric Chu and Toby Boyd and Brad Hekman and Aaron Parisi and Chaoyi Zhang and Kornraphop Kawintiranon and Tania Bedrax-Weiss and Oliver Wang and Ya Xu and Ollie Purkiss and Uri Mendlovic and Ilaï Deutel and Nam Nguyen and Adam Langley and Flip Korn and Lucia Rossazza and Alexandre Ramé and Sagar Waghmare and Helen Miller and Nathan Byrd and Ashrith Sheshan and Raia Hadsell and Sangnie Bhardwaj and Pawel Janus and Tero Rissa and Dan Horgan and Alvin Abdagic and Lior Belenki and James Allingham and Anima Singh and Theo Guidroz and Srivatsan Srinivasan and Herman Schmit and Kristen Chiafullo and Andre Elisseeff and Nilpa Jha and Prateek Kolhar and Leonard Berrada and Frank Ding and Xiance Si and Shrestha Basu Mallick and Franz Och and Sofia Erell and Eric Ni and Tejasi Latkar and Sherry Yang and Petar Sirkovic and Ziqiang Feng and Robert Leland and Rachel Hornung and Gang Wu and Charles Blundell and Hamidreza Alvari and Po-Sen Huang and Cathy Yip and Sanja Deur and Li Liu and Gabriela Surita and Pablo Duque and Dima Damen and Johnson Jia and Arthur Guez and Markus Mircea and Animesh Sinha and Alberto Magni and Paweł Stradomski and Tal Marian and Vlado Galić and Wenhu Chen and Hisham Husain and Achintya Singhal and Dominik Grewe and François-Xavier Aubet and Shuang Song and Lorenzo Blanco and Leland Rechis and Lewis Ho and Rich Munoz and Kelvin Zheng and Jessica Hamrick and Kevin Mather and Hagai Taitelbaum and Eliza Rutherford and Yun Lei and Kuangyuan Chen and Anand Shukla and Erica Moreira and Eric Doi and Berivan Isik and Nir Shabat and Dominika Rogozińska and Kashyap Kolipaka and Jason Chang and Eugen Vušak and Srinivasan Venkatachary and Shadi Noghabi and Tarun Bharti and Younghoon Jun and Aleksandr Zaks and Simon Green and Jeshwanth Challagundla and William Wong and Muqthar Mohammad and Dean Hirsch and Yong Cheng and Iftekhar Naim and Lev Proleev and Damien Vincent and Aayush Singh and Maxim Krikun and Dilip Krishnan and Zoubin Ghahramani and Aviel Atias and Rajeev Aggarwal and Christo Kirov and Dimitrios Vytiniotis and Christy Koh and Alexandra Chronopoulou and Pawan Dogra and Vlad-Doru Ion and Gladys Tyen and Jason Lee and Felix Weissenberger and Trevor Strohman and Ashwin Balakrishna and Jack Rae and Marko Velic and Raoul de Liedekerke and Oded Elyada and Wentao Yuan and Canoee Liu and Lior Shani and Sergey Kishchenko and Bea Alessio and Yandong Li and Richard Song and Sam Kwei and Orion Jankowski and Aneesh Pappu and Youhei Namiki and Yenai Ma and Nilesh Tripuraneni and Colin Cherry and Marissa Ikonomidis and Yu-Cheng Ling and Colin Ji and Beka Westberg and Auriel Wright and Da Yu and David Parkinson and Swaroop Ramaswamy and Jerome Connor and Soheil Hassas Yeganeh and Snchit Grover and George Kenwright and Lubo Litchev and Chris Apps and Alex Tomala and Felix Halim and Alex Castro-Ros and Zefei Li and Anudhyan Boral and Pauline Sho and Michal Yarom and Eric Malmi and David Klinghoffer and Rebecca Lin and Alan Ansell and Pradeep Kumar S and Shubin Zhao and Siqi Zuo and Adam Santoro and Heng-Tze Cheng and Solomon Demmessie and Yuchi Liu and Nicole Brichtova and Allie Culp and Nathaniel Braun and Dan Graur and Will Ng and Nikhil Mehta and Aaron Phillips and Patrik Sundberg and Varun Godbole and Fangyu Liu and Yash Katariya and David Rim and Mojtaba Seyedhosseini and Sean Ammirati and Jonas Valfridsson and Mahan Malihi and Timothy Knight and Andeep Toor and Thomas Lampe and Abe Ittycheriah and Lewis Chiang and Chak Yeung and Alexandre Fréchette and Jinmeng Rao and Huisheng Wang and Himanshu Srivastava and Richard Zhang and Rocky Rhodes and Ariel Brand and Dean Weesner and Ilya Figotin and Felix Gimeno and Rachana Fellinger and Pierre Marcenac and José Leal and Eyal Marcus and Victor Cotruta and Rodrigo Cabrera and Sheryl Luo and Dan Garrette and Vera Axelrod and Sorin Baltateanu and David Barker and Dongkai Chen and Horia Toma and Ben Ingram and Jason Riesa and Chinmay Kulkarni and Yujing Zhang and Hongbin Liu and Chao Wang and Martin Polacek and Will Wu and Kai Hui and Adrian N Reyes and Yi Su and Megan Barnes and Ishaan Malhi and Anfal Siddiqui and Qixuan Feng and Mihai Damaschin and Daniele Pighin and Andreas Steiner and Samuel Yang and Ramya Sree Boppana and Simeon Ivanov and Arun Kandoor and Aditya Shah and Asier Mujika and Da Huang and Christopher A. Choquette-Choo and Mohak Patel and Tianhe Yu and Toni Creswell and Jerry and Liu and Catarina Barros and Yasaman Razeghi and Aurko Roy and Phil Culliton and Binbin Xiong and Jiaqi Pan and Thomas Strohmann and Tolly Powell and Babi Seal and Doug DeCarlo and Pranav Shyam and Kaan Katircioglu and Xuezhi Wang and Cassidy Hardin and Immanuel Odisho and Josef Broder and Oscar Chang and Arun Nair and Artem Shtefan and Maura O'Brien and Manu Agarwal and Sahitya Potluri and Siddharth Goyal and Amit Jhindal and Saksham Thakur and Yury Stuken and James Lyon and Kristina Toutanova and Fangxiaoyu Feng and Austin Wu and Ben Horn and Alek Wang and Alex Cullum and Gabe Taubman and Disha Shrivastava and Chongyang Shi and Hamish Tomlinson and Roma Patel and Tao Tu and Ada Maksutaj Oflazer and Francesco Pongetti and Mingyao Yang and Adrien Ali Taïga and Vincent Perot and Nuo Wang Pierse and Feng Han and Yoel Drori and Iñaki Iturrate and Ayan Chakrabarti and Legg Yeung and Dave Dopson and Yi-ting Chen and Apoorv Kulshreshtha and Tongfei Guo and Philip Pham and Tal Schuster and Junquan Chen and Alex Polozov and Jinwei Xing and Huanjie Zhou and Praneeth Kacham and Doron Kukliansky and Antoine Miech and Sergey Yaroshenko and Ed Chi and Sholto Douglas and Hongliang Fei and Mathieu Blondel and Preethi Myla and Lior Madmoni and Xing Wu and Daniel Keysers and Kristian Kjems and Isabela Albuquerque and Lijun Yu and Joel D'sa and Michelle Plantan and Vlad Ionescu and Jaume Sanchez Elias and Abhirut Gupta and Manish Reddy Vuyyuru and Fred Alcober and Tong Zhou and Kaiyang Ji and Florian Hartmann and Subha Puttagunta and Hugo Song and Ehsan Amid and Anca Stefanoiu and Andrew Lee and Paul Pucciarelli and Emma Wang and Amit Raul and Slav Petrov and Isaac Tian and Valentin Anklin and Nana Nti and Victor Gomes and Max Schumacher and Grace Vesom and Alex Panagopoulos and Konstantinos Bousmalis and Daniel Andor and Josh Jacob and Yuan Zhang and Bill Rosgen and Matija Kecman and Matthew Tung and Alexandra Belias and Noah Goodman and Paul Covington and Brian Wieder and Nikita Saxena and Elnaz Davoodi and Muhuan Huang and Sharath Maddineni and Vincent Roulet and Folawiyo Campbell-Ajala and Pier Giuseppe Sessa and Xintian and Wu and Guangda Lai and Paul Collins and Alex Haig and Vytenis Sakenas and Xiaowei Xu and Marissa Giustina and Laurent El Shafey and Pichi Charoenpanit and Shefali Garg and Joshua Ainslie and Boone Severson and Montse Gonzalez Arenas and Shreya Pathak and Sujee Rajayogam and Jie Feng and Michiel Bakker and Sheng Li and Nevan Wichers and Jamie Rogers and Xinyang Geng and Yeqing Li and Rolf Jagerman and Chao Jia and Nadav Olmert and David Sharon and Matthew Mauger and Sandeep Mariserla and Hongxu Ma and Megha Mohabey and Kyuyeun Kim and Alek Andreev and Scott Pollom and Juliette Love and Vihan Jain and Priyanka Agrawal and Yannick Schroecker and Alisa Fortin and Manfred Warmuth and Ji Liu and Andrew Leach and Irina Blok and Ganesh Poomal Girirajan and Roee Aharoni and Benigno Uria and Andrei Sozanschi and Dan Goldberg and Lucian Ionita and Marco Tulio Ribeiro and Martin Zlocha and Vighnesh Birodkar and Sami Lachgar and Liangzhe Yuan and Himadri Choudhury and Matt Ginsberg and Fei Zheng and Gregory Dibb and Emily Graves and Swachhand Lokhande and Gabriel Rasskin and George-Cristian Muraru and Corbin Quick and Sandeep Tata and Pierre Sermanet and Aditya Chawla and Itay Karo and Yan Wang and Susan Zhang and Orgad Keller and Anca Dragan and Guolong Su and Ian Chou and Xi Liu and Yiqing Tao and Shruthi Prabhakara and Marc Wilson and Ruibo Liu and Shibo Wang and Georgie Evans and David Du and Alfonso Castaño and Gautam Prasad and Mona El Mahdy and Sebastian Gerlach and Machel Reid and Jarrod Kahn and Amir Zait and Thanumalayan Sankaranarayana Pillai and Thatcher Ulrich and Guanyu Wang and Jan Wassenberg and Efrat Farkash and Kiran Yalasangi and Congchao Wang and Maria Bauza and Simon Bucher and Ting Liu and Jun Yan and Gary Leung and Vikas Sindhwani and Parker Barnes and Avi Singh and Ivan Jurin and Jichuan Chang and Niket Kumar Bhumihar and Sivan Eiger and Gui Citovsky and Ben Withbroe and Zhang Li and Siyang Xue and Niccolò Dal Santo and Georgi Stoyanov and Yves Raimond and Steven Zheng and Yilin Gao and Vít Listík and Sławek Kwasiborski and Rachel Saputro and Adnan Ozturel and Ganesh Mallya and Kushal Majmundar and Ross West and Paul Caron and Jinliang Wei and Lluis Castrejon and Sharad Vikram and Deepak Ramachandran and Nikhil Dhawan and Jiho Park and Sara Smoot and George van den Driessche and Yochai Blau and Chase Malik and Wei Liang and Roy Hirsch and Cicero Nogueira dos Santos and Eugene Weinstein and Aäron van den Oord and Sid Lall and Nicholas FitzGerald and Zixuan Jiang and Xuan Yang and Dale Webster and Ali Elqursh and Aedan Pope and Georges Rotival and David Raposo and Wanzheng Zhu and Jeff Dean and Sami Alabed and Dustin Tran and Arushi Gupta and Zach Gleicher and Jessica Austin and Edouard Rosseel and Megh Umekar and Dipanjan Das and Yinghao Sun and Kai Chen and Karolis Misiunas and Xiang Zhou and Yixian Di and Alyssa Loo and Josh Newlan and Bo Li and Vinay Ramasesh and Ying Xu and Alex Chen and Sudeep Gandhe and Radu Soricut and Nikita Gupta and Shuguang Hu and Seliem El-Sayed and Xavier Garcia and Idan Brusilovsky and Pu-Chin Chen and Andrew Bolt and Lu Huang and Alex Gurney and Zhiying Zhang and Alexander Pritzel and Jarek Wilkiewicz and Bryan Seybold and Bhargav Kanagal Shamanna and Felix Fischer and Josef Dean and Karan Gill and Ross Mcilroy and Abhishek Bhowmick and Jeremy Selier and Antoine Yang and Derek Cheng and Vladimir Magay and Jie Tan and Dhriti Varma and Christian Walder and Tomas Kocisky and Ryo Nakashima and Paul Natsev and Mike Kwong and Ionel Gog and Chiyuan Zhang and Sander Dieleman and Thomas Jimma and Andrey Ryabtsev and Siddhartha Brahma and David Steiner and Dayou Du and Ante Žužul and Mislav Žanić and Mukund Raghavachari and Willi Gierke and Zeyu Zheng and Dessie Petrova and Yann Dauphin and Yuchuan Liu and Ido Kessler and Steven Hand and Chris Duvarney and Seokhwan Kim and Hyo Lee and Léonard Hussenot and Jeffrey Hui and Josh Smith and Deepali Jain and Jiawei Xia and Gaurav Singh Tomar and Keyvan Amiri and Du Phan and Fabian Fuchs and Tobias Weyand and Nenad Tomasev and Alexandra Cordell and Xin Liu and Jonathan Mallinson and Pankaj Joshi and Andy Crawford and Arun Suggala and Steve Chien and Nick Fernando and Mariella Sanchez-Vargas and Duncan Williams and Phil Crone and Xiyang Luo and Igor Karpov and Jyn Shan and Terry Thurk and Robin Strudel and Paul Voigtlaender and Piyush Patil and Tim Dozat and Ali Khodaei and Sahil Singla and Piotr Ambroszczyk and Qiyin Wu and Yifan Chang and Brian Roark and Chaitra Hegde and Tianli Ding and Angelos Filos and Zhongru Wu and André Susano Pinto and Shuang Liu and Saarthak Khanna and Aditya Pandey and Siobhan Mcloughlin and Qiujia Li and Sam Haves and Allan Zhou and Elena Buchatskaya and Isabel Leal and Peter de Boursac and Nami Akazawa and Nina Anderson and Terry Chen and Krishna Somandepalli and Chen Liang and Sheela Goenka and Stephanie Winkler and Alexander Grushetsky and Yifan Ding and Jamie Smith and Fan Ye and Jordi Pont-Tuset and Eric Li and Ruichao Li and Tomer Golany and Dawid Wegner and Tao Jiang and Omer Barak and Yuan Shangguan and Eszter Vértes and Renee Wong and Jörg Bornschein and Alex Tudor and Michele Bevilacqua and Tom Schaul and Ankit Singh Rawat and Yang Zhao and Kyriakos Axiotis and Lei Meng and Cory McLean and Jonathan Lai and Jennifer Beattie and Nate Kushman and Yaxin Liu and Blair Kutzman and Fiona Lang and Jingchen Ye and Praneeth Netrapalli and Pushkar Mishra and Myriam Khan and Megha Goel and Rob Willoughby and David Tian and Honglei Zhuang and JD Chen and Zak Tsai and Tasos Kementsietsidis and Arjun Khare and James Keeling and Keyang Xu and Nathan Waters and Florent Altché and Ashok Popat and Bhavishya Mittal and David Saxton and Dalia El Badawy and Michael Mathieu and Zheng Zheng and Hao Zhou and Nishant Ranka and Richard Shin and Qingnan Duan and Tim Salimans and Ioana Mihailescu and Uri Shaham and Ming-Wei Chang and Yannis Assael and Nishanth Dikkala and Martin Izzard and Vincent Cohen-Addad and Cat Graves and Vlad Feinberg and Grace Chung and DJ Strouse and Danny Karmon and Sahand Sharifzadeh and Zoe Ashwood and Khiem Pham and Jon Blanton and Alex Vasiloff and Jarred Barber and Mark Geller and Aurick Zhou and Fedir Zubach and Tzu-Kuo Huang and Lei Zhang and Himanshu Gupta and Matt Young and Julia Proskurnia and Ronny Votel and Valentin Gabeur and Gabriel Barcik and Aditya Tripathi and Hongkun Yu and Geng Yan and Beer Changpinyo and Filip Pavetić and Amy Coyle and Yasuhisa Fujii and Jorge Gonzalez Mendez and Tianhao Zhou and Harish Rajamani and Blake Hechtman and Eddie Cao and Da-Cheng Juan and Yi-Xuan Tan and Valentin Dalibard and Yilun Du and Natalie Clay and Kaisheng Yao and Wenhao Jia and Dimple Vijaykumar and Yuxiang Zhou and Xinyi Bai and Wei-Chih Hung and Steven Pecht and Georgi Todorov and Nikhil Khadke and Pramod Gupta and Preethi Lahoti and Arnaud Autef and Karthik Duddu and James Lee-Thorp and Alexander Bykovsky and Tautvydas Misiunas and Sebastian Flennerhag and Santhosh Thangaraj and Jed McGiffin and Zack Nado and Markus Kunesch and Andreas Noever and Amir Hertz and Marco Liang and Victor Stone and Evan Palmer and Samira Daruki and Arijit Pramanik and Siim Põder and Austin Kyker and Mina Khan and Evgeny Sluzhaev and Marvin Ritter and Avraham Ruderman and Wenlei Zhou and Chirag Nagpal and Kiran Vodrahalli and George Necula and Paul Barham and Ellie Pavlick and Jay Hartford and Izhak Shafran and Long Zhao and Maciej Mikuła and Tom Eccles and Hidetoshi Shimokawa and Kanav Garg and Luke Vilnis and Hanwen Chen and Ilia Shumailov and Kuang-Huei Lee and Abdelrahman Abdelhamed and Meiyan Xie and Vered Cohen and Ester Hlavnova and Dan Malkin and Chawin Sitawarin and James Lottes and Pauline Coquinot and Tianli Yu and Sandeep Kumar and Jingwei Zhang and Aroma Mahendru and Zafarali Ahmed and James Martens and Tao Chen and Aviel Boag and Daiyi Peng and Coline Devin and Arseniy Klimovskiy and Mary Phuong and Danny Vainstein and Jin Xie and Bhuvana Ramabhadran and Nathan Howard and Xinxin Yu and Gitartha Goswami and Jingyu Cui and Sam Shleifer and Mario Pinto and Chih-Kuan Yeh and Ming-Hsuan Yang and Sara Javanmardi and Dan Ethier and Chace Lee and Jordi Orbay and Suyog Kotecha and Carla Bromberg and Pete Shaw and James Thornton and Adi Gerzi Rosenthal and Shane Gu and Matt Thomas and Ian Gemp and Aditya Ayyar and Asahi Ushio and Aarush Selvan and Joel Wee and Chenxi Liu and Maryam Majzoubi and Weiren Yu and Jake Abernethy and Tyler Liechty and Renke Pan and Hoang Nguyen and Qiong and Hu and Sarah Perrin and Abhinav Arora and Emily Pitler and Weiyi Wang and Kaushik Shivakumar and Flavien Prost and Ben Limonchik and Jing Wang and Yi Gao and Timothee Cour and Shyamal Buch and Huan Gui and Maria Ivanova and Philipp Neubeck and Kelvin Chan and Lucy Kim and Huizhong Chen and Naman Goyal and Da-Woon Chung and Lu Liu and Yao Su and Anastasia Petrushkina and Jiajun Shen and Armand Joulin and Yuanzhong Xu and Stein Xudong Lin and Yana Kulizhskaya and Ciprian Chelba and Shobha Vasudevan and Eli Collins and Vasilisa Bashlovkina and Tony Lu and Doug Fritz and Jongbin Park and Yanqi Zhou and Chen Su and Richard Tanburn and Mikhail Sushkov and Mitchelle Rasquinha and Jinning Li and Jennifer Prendki and Yiming Li and Pallavi LV and Shriya Sharma and Hen Fitoussi and Hui Huang and Andrew Dai and Phuong Dao and Mike Burrows and Henry Prior and Danfeng Qin and Golan Pundak and Lars Lowe Sjoesund and Art Khurshudov and Zhenkai Zhu and Albert Webson and Elizabeth Kemp and Tat Tan and Saurabh Agrawal and Susie Sargsyan and Liqun Cheng and Jim Stephan and Tom Kwiatkowski and David Reid and Arunkumar Byravan and Assaf Hurwitz Michaely and Nicolas Heess and Luowei Zhou and Sonam Goenka and Viral Carpenter and Anselm Levskaya and Bo Wang and Reed Roberts and Rémi Leblond and Sharat Chikkerur and Stav Ginzburg and Max Chang and Robert Riachi and Chuqiao and Xu and Zalán Borsos and Michael Pliskin and Julia Pawar and Morgane Lustman and Hannah Kirkwood and Ankit Anand and Aditi Chaudhary and Norbert Kalb and Kieran Milan and Sean Augenstein and Anna Goldie and Laurel Prince and Karthik Raman and Yanhua Sun and Vivian Xia and Aaron Cohen and Zhouyuan Huo and Josh Camp and Seher Ellis and Lukas Zilka and David Vilar Torres and Lisa Patel and Sho Arora and Betty Chan and Jonas Adler and Kareem Ayoub and Jacky Liang and Fayaz Jamil and Jiepu Jiang and Simon Baumgartner and Haitian Sun and Yael Karov and Yaroslav Akulov and Hui Zheng and Irene Cai and Claudio Fantacci and James Rubin and Alex Rav Acha and Mengchao Wang and Nina D'Souza and Rohit Sathyanarayana and Shengyang Dai and Simon Rowe and Andrey Simanovsky and Omer Goldman and Yuheng Kuang and Xiaoyue Pan and Andrew Rosenberg and Tania Rojas-Esponda and Praneet Dutta and Amy Zeng and Irina Jurenka and Greg Farquhar and Yamini Bansal and Shariq Iqbal and Becca Roelofs and Ga-Young Joung and Parker Beak and Changwan Ryu and Ryan Poplin and Yan Wu and Jean-Baptiste Alayrac and Senaka Buthpitiya and Olaf Ronneberger and Caleb Habtegebriel and Wei Li and Paul Cavallaro and Aurora Wei and Guy Bensky and Timo Denk and Harish Ganapathy and Jeff Stanway and Pratik Joshi and Francesco Bertolini and Jessica Lo and Olivia Ma and Zachary Charles and Geta Sampemane and Himanshu Sahni and Xu Chen and Harry Askham and David Gaddy and Peter Young and Jiewen Tan and Matan Eyal and Arthur Bražinskas and Li Zhong and Zhichun Wu and Mark Epstein and Kai Bailey and Andrew Hard and Kamyu Lee and Sasha Goldshtein and Alex Ruiz and Mohammed Badawi and Matthias Lochbrunner and JK Kearns and Ashley Brown and Fabio Pardo and Theophane Weber and Haichuan Yang and Pan-Pan Jiang and Berkin Akin and Zhao Fu and Marcus Wainwright and Chi Zou and Meenu Gaba and Pierre-Antoine Manzagol and Wendy Kan and Yang Song and Karina Zainullina and Rui Lin and Jeongwoo Ko and Salil Deshmukh and Apoorv Jindal and James Svensson and Divya Tyam and Heri Zhao and Christine Kaeser-Chen and Scott Baird and Pooya Moradi and Jamie Hall and Qiuchen Guo and Vincent Tsang and Bowen Liang and Fernando Pereira and Suhas Ganesh and Ivan Korotkov and Jakub Adamek and Sridhar Thiagarajan and Vinh Tran and Charles Chen and Chris Tar and Sanil Jain and Ishita Dasgupta and Taylan Bilal and David Reitter and Kai Zhao and Giulia Vezzani and Yasmin Gehman and Pulkit Mehta and Lauren Beltrone and Xerxes Dotiwalla and Sergio Guadarrama and Zaheer Abbas and Stefani Karp and Petko Georgiev and Chun-Sung Ferng and Marc Brockschmidt and Liqian Peng and Christoph Hirnschall and Vikas Verma and Yingying Bi and Ying Xiao and Avigail Dabush and Kelvin Xu and Phil Wallis and Randall Parker and Qifei Wang and Yang Xu and Ilkin Safarli and Dinesh Tewari and Yin Zhang and Seungyeon Kim and Andrea Gesmundo and Mackenzie Thomas and Sergey Levi and Ahmed Chowdhury and Kanishka Rao and Peter Garst and Sam Conway-Rahman and Helen Ran and Kay McKinney and Zhisheng Xiao and Wenhao Yu and Rohan Agrawal and Axel Stjerngren and Catalin Ionescu and Jingjing Chen and Vivek Sharma and Justin Chiu and Fei Liu and Ken Franko and Clayton Sanford and Xingyu Cai and Paul Michel and Sanjay Ganapathy and Jane Labanowski and Zachary Garrett and Ben Vargas and Sean Sun and Bryan Gale and Thomas Buschmann and Guillaume Desjardins and Nimesh Ghelani and Palak Jain and Mudit Verma and Chulayuth Asawaroengchai and Julian Eisenschlos and Jitendra Harlalka and Hideto Kazawa and Don Metzler and Joshua Howland and Ying Jian and Jake Ades and Viral Shah and Tynan Gangwani and Seungji Lee and Roman Ring and Steven M. Hernandez and Dean Reich and Amer Sinha and Ashutosh Sathe and Joe Kovac and Ashleah Gill and Ajay Kannan and Andrea D'olimpio and Martin Sevenich and Jay Whang and Been Kim and Khe Chai Sim and Jilin Chen and Jiageng Zhang and Shuba Lall and Yossi Matias and Bill Jia and Abe Friesen and Sara Nasso and Ashish Thapliyal and Bryan Perozzi and Ting Yu and Anna Shekhawat and Safeen Huda and Peter Grabowski and Eric Wang and Ashwin Sreevatsa and Hilal Dib and Mehadi Hassen and Parker Schuh and Vedrana Milutinovic and Chris Welty and Michael Quinn and Ali Shah and Bangju Wang and Gabe Barth-Maron and Justin Frye and Natalie Axelsson and Tao Zhu and Yukun Ma and Irene Giannoumis and Hanie Sedghi and Chang Ye and Yi Luan and Kevin Aydin and Bilva Chandra and Vivek Sampathkumar and Ronny Huang and Victor Lavrenko and Ahmed Eleryan and Zhi Hong and Steven Hansen and Sara Mc Carthy and Bidisha Samanta and Domagoj Ćevid and Xin Wang and Fangtao Li and Michael Voznesensky and Matt Hoffman and Andreas Terzis and Vikash Sehwag and Gil Fidel and Luheng He and Mu Cai and Yanzhang He and Alex Feng and Martin Nikoltchev and Samrat Phatale and Jason Chase and Rory Lawton and Ming Zhang and Tom Ouyang and Manuel Tragut and Mehdi Hafezi Manshadi and Arjun Narayanan and Jiaming Shen and Xu Gao and Tolga Bolukbasi and Nick Roy and Xin Li and Daniel Golovin and Liviu Panait and Zhen Qin and Guangxing Han and Thomas Anthony and Sneha Kudugunta and Viorica Patraucean and Aniket Ray and Xinyun Chen and Xiaochen Yang and Tanuj Bhatia and Pranav Talluri and Alex Morris and Andrija Ražnatović and Bethanie Brownfield and James An and Sheng Peng and Patrick Kane and Ce Zheng and Nico Duduta and Joshua Kessinger and James Noraky and Siqi Liu and Keran Rong and Petar Veličković and Keith Rush and Alex Goldin and Fanny Wei and Shiva Mohan Reddy Garlapati and Caroline Pantofaru and Okwan Kwon and Jianmo Ni and Eric Noland and Julia Di Trapani and Françoise Beaufays and Abhijit Guha Roy and Yinlam Chow and Aybuke Turker and Geoffrey Cideron and Lantao Mei and Jon Clark and Qingyun Dou and Matko Bošnjak and Ralph Leith and Yuqing Du and Amir Yazdanbakhsh and Milad Nasr and Chester Kwak and Suraj Satishkumar Sheth and Alex Kaskasoli and Ankesh Anand and Balaji Lakshminarayanan and Sammy Jerome and David Bieber and Chun-Te Chu and Alexandre Senges and Tianxiao Shen and Mukund Sridhar and Ndaba Ndebele and Benjamin Beyret and Shakir Mohamed and Mia Chen and Markus Freitag and Jiaxian Guo and Luyang Liu and Paul Roit and Heng Chen and Shen Yan and Tom Stone and JD Co-Reyes and Jeremy Cole and Salvatore Scellato and Shekoofeh Azizi and Hadi Hashemi and Alicia Jin and Anand Iyer and Marcella Valentine and András György and Arun Ahuja and Daniel Hernandez Diaz and Chen-Yu Lee and Nathan Clement and Weize Kong and Drew Garmon and Ishaan Watts and Kush Bhatia and Khyatti Gupta and Matt Miecnikowski and Hugo Vallet and Ankur Taly and Edward Loper and Saket Joshi and James Atwood and Jo Chick and Mark Collier and Fotis Iliopoulos and Ryan Trostle and Beliz Gunel and Ramiro Leal-Cavazos and Arnar Mar Hrafnkelsson and Michael Guzman and Xiaoen Ju and Andy Forbes and Jesse Emond and Kushal Chauhan and Ben Caine and Li Xiao and Wenjun Zeng and Alexandre Moufarek and Daniel Murphy and Maya Meng and Nitish Gupta and Felix Riedel and Anil Das and Elijah Lawal and Shashi Narayan and Tiberiu Sosea and James Swirhun and Linda Friso and Behnam Neyshabur and Jing Lu and Sertan Girgin and Michael Wunder and Edouard Yvinec and Aroonalok Pyne and Victor Carbune and Shruti Rijhwani and Yang Guo and Tulsee Doshi and Anton Briukhov and Max Bain and Ayal Hitron and Xuanhui Wang and Ashish Gupta and Ke Chen and Cosmo Du and Weiyang Zhang and Dhruv Shah and Arjun Akula and Max Dylla and Ashyana Kachra and Weicheng Kuo and Tingting Zou and Lily Wang and Luyao Xu and Jifan Zhu and Justin Snyder and Sachit Menon and Orhan Firat and Igor Mordatch and Yuan Yuan and Natalia Ponomareva and Rory Blevins and Lawrence Moore and Weijun Wang and Phil Chen and Martin Scholz and Artur Dwornik and Jason Lin and Sicheng Li and Diego Antognini and Te I and Xiaodan Song and Matt Miller and Uday Kalra and Adam Raveret and Oscar Akerlund and Felix Wu and Andrew Nystrom and Namrata Godbole and Tianqi Liu and Hannah DeBalsi and Jewel Zhao and Buhuang Liu and Avi Caciularu and Lauren Lax and Urvashi Khandelwal and Victoria Langston and Eric Bailey and Silvio Lattanzi and Yufei Wang and Neel Kovelamudi and Sneha Mondal and Guru Guruganesh and Nan Hua and Ofir Roval and Paweł Wesołowski and Rishikesh Ingale and Jonathan Halcrow and Tim Sohn and Christof Angermueller and Bahram Raad and Eli Stickgold and Eva Lu and Alec Kosik and Jing Xie and Timothy Lillicrap and Austin Huang and Lydia Lihui Zhang and Dominik Paulus and Clement Farabet and Alex Wertheim and Bing Wang and Rishabh Joshi and Chu-ling Ko and Yonghui Wu and Shubham Agrawal and Lily Lin and XiangHai Sheng and Peter Sung and Tyler Breland-King and Christina Butterfield and Swapnil Gawde and Sumeet Singh and Qiao Zhang and Raj Apte and Shilpa Shetty and Adrian Hutter and Tao Li and Elizabeth Salesky and Federico Lebron and Jonni Kanerva and Michela Paganini and Arthur Nguyen and Rohith Vallu and Jan-Thorsten Peter and Sarmishta Velury and David Kao and Jay Hoover and Anna Bortsova and Colton Bishop and Shoshana Jakobovits and Alessandro Agostini and Alekh Agarwal and Chang Liu and Charles Kwong and Sasan Tavakkol and Ioana Bica and Alex Greve and Anirudh GP and Jake Marcus and Le Hou and Tom Duerig and Rivka Moroshko and Dave Lacey and Andy Davis and Julien Amelot and Guohui Wang and Frank Kim and Theofilos Strinopoulos and Hui Wan and Charline Le Lan and Shankar Krishnan and Haotian Tang and Peter Humphreys and Junwen Bai and Idan Heimlich Shtacher and Diego Machado and Chenxi Pang and Ken Burke and Dangyi Liu and Renga Aravamudhan and Yue Song and Ed Hirst and Abhimanyu Singh and Brendan Jou and Liang Bai and Francesco Piccinno and Chuyuan Kelly Fu and Robin Alazard and Barak Meiri and Daniel Winter and Charlie Chen and Mingda Zhang and Jens Heitkaemper and John Lambert and Jinhyuk Lee and Alexander Frömmgen and Sergey Rogulenko and Pranav Nair and Paul Niemczyk and Anton Bulyenov and Bibo Xu and Hadar Shemtov and Morteza Zadimoghaddam and Serge Toropov and Mateo Wirth and Hanjun Dai and Sreenivas Gollapudi and Daniel Zheng and Alex Kurakin and Chansoo Lee and Kalesha Bullard and Nicolas Serrano and Ivana Balazevic and Yang Li and Johan Schalkwyk and Mark Murphy and Mingyang Zhang and Kevin Sequeira and Romina Datta and Nishant Agrawal and Charles Sutton and Nithya Attaluri and Mencher Chiang and Wael Farhan and Gregory Thornton and Kate Lin and Travis Choma and Hung Nguyen and Kingshuk Dasgupta and Dirk Robinson and Iulia Comşa and Michael Riley and Arjun Pillai and Basil Mustafa and Ben Golan and Amir Zandieh and Jean-Baptiste Lespiau and Billy Porter and David Ross and Sujeevan Rajayogam and Mohit Agarwal and Subhashini Venugopalan and Bobak Shahriari and Qiqi Yan and Hao Xu and Taylor Tobin and Pavel Dubov and Hongzhi Shi and Adrià Recasens and Anton Kovsharov and Sebastian Borgeaud and Lucio Dery and Shanthal Vasanth and Elena Gribovskaya and Linhai Qiu and Mahdis Mahdieh and Wojtek Skut and Elizabeth Nielsen and CJ Zheng and Adams Yu and Carrie Grimes Bostock and Shaleen Gupta and Aaron Archer and Chris Rawles and Elinor Davies and Alexey Svyatkovskiy and Tomy Tsai and Yoni Halpern and Christian Reisswig and Bartek Wydrowski and Bo Chang and Joan Puigcerver and Mor Hazan Taege and Jian Li and Eva Schnider and Xinjian Li and Dragos Dena and Yunhan Xu and Umesh Telang and Tianze Shi and Heiga Zen and Kyle Kastner and Yeongil Ko and Neesha Subramaniam and Aviral Kumar and Pete Blois and Zhuyun Dai and John Wieting and Yifeng Lu and Yoel Zeldes and Tian Xie and Anja Hauth and Alexandru Ţifrea and Yuqi Li and Sam El-Husseini and Dan Abolafia and Howard Zhou and Wen Ding and Sahra Ghalebikesabi and Carlos Guía and Andrii Maksai and Ágoston Weisz and Sercan Arik and Nick Sukhanov and Aga Świetlik and Xuhui Jia and Luo Yu and Weiyue Wang and Mark Brand and Dawn Bloxwich and Sean Kirmani and Zhe Chen and Alec Go and Pablo Sprechmann and Nithish Kannen and Alen Carin and Paramjit Sandhu and Isabel Edkins and Leslie Nooteboom and Jai Gupta and Loren Maggiore and Javad Azizi and Yael Pritch and Pengcheng Yin and Mansi Gupta and Danny Tarlow and Duncan Smith and Desi Ivanov and Mohammad Babaeizadeh and Ankita Goel and Satish Kambala and Grace Chu and Matej Kastelic and Michelle Liu and Hagen Soltau and Austin Stone and Shivani Agrawal and Min Kim and Kedar Soparkar and Srinivas Tadepalli and Oskar Bunyan and Rachel Soh and Arvind Kannan and DY Kim and Blake JianHang Chen and Afief Halumi and Sudeshna Roy and Yulong Wang and Olcan Sercinoglu and Gena Gibson and Sijal Bhatnagar and Motoki Sano and Daniel von Dincklage and Qingchun Ren and Blagoj Mitrevski and Mirek Olšák and Jennifer She and Carl Doersch and Jilei and Wang and Bingyuan Liu and Qijun Tan and Tamar Yakar and Tris Warkentin and Alex Ramirez and Carl Lebsack and Josh Dillon and Rajiv Mathews and Tom Cobley and Zelin Wu and Zhuoyuan Chen and Jon Simon and Swaroop Nath and Tara Sainath and Alexei Bendebury and Ryan Julian and Bharath Mankalale and Daria Ćurko and Paulo Zacchello and Adam R. Brown and Kiranbir Sodhia and Heidi Howard and Sergi Caelles and Abhinav Gupta and Gareth Evans and Anna Bulanova and Lesley Katzen and Roman Goldenberg and Anton Tsitsulin and Joe Stanton and Benoit Schillings and Vitaly Kovalev and Corey Fry and Rushin Shah and Kuo Lin and Shyam Upadhyay and Cheng Li and Soroush Radpour and Marcello Maggioni and Jing Xiong and Lukas Haas and Jenny Brennan and Aishwarya Kamath and Nikolay Savinov and Arsha Nagrani and Trevor Yacovone and Ryan Kappedal and Kostas Andriopoulos and Li Lao and YaGuang Li and Grigory Rozhdestvenskiy and Kazuma Hashimoto and Andrew Audibert and Sophia Austin and Daniel Rodriguez and Anian Ruoss and Garrett Honke and Deep Karkhanis and Xi Xiong and Qing Wei and James Huang and Zhaoqi Leng and Vittal Premachandran and Stan Bileschi and Georgios Evangelopoulos and Thomas Mensink and Jay Pavagadhi and Denis Teplyashin and Paul Chang and Linting Xue and Garrett Tanzer and Sally Goldman and Kaushal Patel and Shixin Li and Jeremy Wiesner and Ivy Zheng and Ian Stewart-Binks and Jie Han and Zhi Li and Liangchen Luo and Karel Lenc and Mario Lučić and Fuzhao Xue and Ryan Mullins and Alexey Guseynov and Chung-Ching Chang and Isaac Galatzer-Levy and Adam Zhang and Garrett Bingham and Grace Hu and Ale Hartman and Yue Ma and Jordan Griffith and Alex Irpan and Carey Radebaugh and Summer Yue and Lijie Fan and Victor Ungureanu and Christina Sorokin and Hannah Teufel and Peiran Li and Rohan Anil and Dimitris Paparas and Todd Wang and Chu-Cheng Lin and Hui Peng and Megan Shum and Goran Petrovic and Demetra Brady and Richard Nguyen and Klaus Macherey and Zhihao Li and Harman Singh and Madhavi Yenugula and Mariko Iinuma and Xinyi Chen and Kavya Kopparapu and Alexey Stern and Shachi Dave and Chandu Thekkath and Florence Perot and Anurag Kumar and Fangda Li and Yang Xiao and Matthew Bilotti and Mohammad Hossein Bateni and Isaac Noble and Lisa Lee and Amelio Vázquez-Reina and Julian Salazar and Xiaomeng Yang and Boyu Wang and Ela Gruzewska and Anand Rao and Sindhu Raghuram and Zheng Xu and Eyal Ben-David and Jieru Mei and Sid Dalmia and Zhaoyi Zhang and Yuchen Liu and Gagan Bansal and Helena Pankov and Steven Schwarcz and Andrea Burns and Christine Chan and Sumit Sanghai and Ricky Liang and Ethan Liang and Antoine He and Amy Stuart and Arun Narayanan and Yukun Zhu and Christian Frank and Bahar Fatemi and Amit Sabne and Oran Lang and Indro Bhattacharya and Shane Settle and Maria Wang and Brendan McMahan and Andrea Tacchetti and Livio Baldini Soares and Majid Hadian and Serkan Cabi and Timothy Chung and Nikita Putikhin and Gang Li and Jeremy Chen and Austin Tarango and Henryk Michalewski and Mehran Kazemi and Hussain Masoom and Hila Sheftel and Rakesh Shivanna and Archita Vadali and Ramona Comanescu and Doug Reid and Joss Moore and Arvind Neelakantan and Michaël Sander and Jonathan Herzig and Aviv Rosenberg and Mostafa Dehghani and JD Choi and Michael Fink and Reid Hayes and Eric Ge and Shitao Weng and Chia-Hua Ho and John Karro and Kalpesh Krishna and Lam Nguyen Thiet and Amy Skerry-Ryan and Daniel Eppens and Marco Andreetto and Navin Sarma and Silvano Bonacina and Burcu Karagol Ayan and Megha Nawhal and Zhihao Shan and Mike Dusenberry and Shantanu Thakoor and Sagar Gubbi and Duc Dung Nguyen and Reut Tsarfaty and Samuel Albanie and Jovana Mitrović and Meet Gandhi and Bo-Juen Chen and Alessandro Epasto and Georgi Stephanov and Ye Jin and Samuel Gehman and Aida Amini and Jack Weber and Feryal Behbahani and Shawn Xu and Miltos Allamanis and Xi Chen and Myle Ott and Claire Sha and Michal Jastrzebski and Hang Qi and David Greene and Xinyi Wu and Abodunrinwa Toki and Daniel Vlasic and Jane Shapiro and Ragha Kotikalapudi and Zhe Shen and Takaaki Saeki and Sirui Xie and Albin Cassirer and Shikhar Bharadwaj and Tatsuya Kiyono and Srinadh Bhojanapalli and Elan Rosenfeld and Sam Ritter and Jieming Mao and João Gabriel Oliveira and Zoltan Egyed and Bernd Bandemer and Emilio Parisotto and Keisuke Kinoshita and Juliette Pluto and Petros Maniatis and Steve Li and Yaohui Guo and Golnaz Ghiasi and Jean Tarbouriech and Srimon Chatterjee and Julie Jin and Katrina and Xu and Jennimaria Palomaki and Séb Arnold and Madhavi Sewak and Federico Piccinini and Mohit Sharma and Ben Albrecht and Sean Purser-haskell and Ashwin Vaswani and Chongyan Chen and Matheus Wisniewski and Qin Cao and John Aslanides and Nguyet Minh Phu and Maximilian Sieb and Lauren Agubuzu and Anne Zheng and Daniel Sohn and Marco Selvi and Anders Andreassen and Krishan Subudhi and Prem Eruvbetine and Oliver Woodman and Tomas Mery and Sebastian Krause and Xiaoqi Ren and Xiao Ma and Jincheng Luo and Dawn Chen and Wei Fan and Henry Griffiths and Christian Schuler and Alice Li and Shujian Zhang and Jean-Michel Sarr and Shixin Luo and Riccardo Patana and Matthew Watson and Dani Naboulsi and Michael Collins and Sailesh Sidhwani and Emiel Hoogeboom and Sharon Silver and Emily Caveness and Xiaokai Zhao and Mikel Rodriguez and Maxine Deines and Libin Bai and Patrick Griffin and Marco Tagliasacchi and Emily Xue and Spandana Raj Babbula and Bo Pang and Nan Ding and Gloria Shen and Elijah Peake and Remi Crocker and Shubha Srinivas Raghvendra and Danny Swisher and Woohyun Han and Richa Singh and Ling Wu and Vladimir Pchelin and Tsendsuren Munkhdalai and Dana Alon and Geoff Bacon and Efren Robles and Jannis Bulian and Melvin Johnson and George Powell and Felipe Tiengo Ferreira and Yaoyiran Li and Frederik Benzing and Mihajlo Velimirović and Hubert Soyer and William Kong and Tony and Nguyên and Zhen Yang and Jeremiah Liu and Joost van Amersfoort and Daniel Gillick and Baochen Sun and Nathalie Rauschmayr and Katie Zhang and Serena Zhan and Tao Zhou and Alexey Frolov and Chengrun Yang and Denis Vnukov and Louis Rouillard and Hongji Li and Amol Mandhane and Nova Fallen and Rajesh Venkataraman and Clara Huiyi Hu and Jennifer Brennan and Jenny Lee and Jerry Chang and Martin Sundermeyer and Zhufeng Pan and Rosemary Ke and Simon Tong and Alex Fabrikant and William Bono and Jindong Gu and Ryan Foley and Yiran Mao and Manolis Delakis and Dhruva Bhaswar and Roy Frostig and Nick Li and Avital Zipori and Cath Hope and Olga Kozlova and Swaroop Mishra and Josip Djolonga and Craig Schiff and Majd Al Merey and Eleftheria Briakou and Peter Morgan and Andy Wan and Avinatan Hassidim and RJ Skerry-Ryan and Kuntal Sengupta and Mary Jasarevic and Praveen Kallakuri and Paige Kunkle and Hannah Brennan and Tom Lieber and Hassan Mansoor and Julian Walker and Bing Zhang and Annie Xie and Goran Žužić and Adaeze Chukwuka and Alex Druinsky and Donghyun Cho and Rui Yao and Ferjad Naeem and Shiraz Butt and Eunyoung Kim and Zhipeng Jia and Mandy Jordan and Adam Lelkes and Mark Kurzeja and Sophie Wang and James Zhao and Andrew Over and Abhishek Chakladar and Marcel Prasetya and Neha Jha and Sriram Ganapathy and Yale Cong and Prakash Shroff and Carl Saroufim and Sobhan Miryoosefi and Mohamed Hammad and Tajwar Nasir and Weijuan Xi and Yang Gao and Young Maeng and Ben Hora and Chin-Yi Cheng and Parisa Haghani and Yoad Lewenberg and Caden Lu and Martin Matysiak and Naina Raisinghani and Huiyu Wang and Lexi Baugher and Rahul Sukthankar and Minh Giang and John Schultz and Noah Fiedel and Minmin Chen and Cheng-Chun Lee and Tapomay Dey and Hao Zheng and Shachi Paul and Celine Smith and Andy Ly and Yicheng Wang and Rishabh Bansal and Bartek Perz and Susanna Ricco and Stasha Blank and Vaishakh Keshava and Deepak Sharma and Marvin Chow and Kunal Lad and Komal Jalan and Simon Osindero and Craig Swanson and Jacob Scott and Anastasija Ilić and Xiaowei Li and Siddhartha Reddy Jonnalagadda and Afzal Shama Soudagar and Yan Xiong and Bat-Orgil Batsaikhan and Daniel Jarrett and Naveen Kumar and Maulik Shah and Matt Lawlor and Austin Waters and Mark Graham and Rhys May and Sabela Ramos and Sandra Lefdal and Zeynep Cankara and Nacho Cano and Brendan O'Donoghue and Jed Borovik and Frederick Liu and Jordan Grimstad and Mahmoud Alnahlawi and Katerina Tsihlas and Tom Hudson and Nikolai Grigorev and Yiling Jia and Terry Huang and Tobenna Peter Igwe and Sergei Lebedev and Xiaodan Tang and Igor Krivokon and Frankie Garcia and Melissa Tan and Eric Jia and Peter Stys and Shikhar Vashishth and Yu Liang and Balaji Venkatraman and Chenjie Gu and Anastasios Kementsietsidis and Chen Zhu and Junehyuk Jung and Yunfei Bai and Mohammad Javad Hosseini and Faruk Ahmed and Aditya Gupta and Xin Yuan and Shereen Ashraf and Shitij Nigam and Gautam Vasudevan and Pranjal Awasthi and Adi Mayrav Gilady and Zelda Mariet and Ramy Eskander and Haiguang Li and Hexiang Hu and Guillermo Garrido and Philippe Schlattner and George Zhang and Rohun Saxena and Petar Dević and Kritika Muralidharan and Ashwin Murthy and Yiqian Zhou and Min Choi and Arissa Wongpanich and Zhengdong Wang and Premal Shah and Yuntao Xu and Yiling Huang and Stephen Spencer and Alice Chen and James Cohan and Junjie Wang and Jonathan Tompson and Junru Wu and Ruba Haroun and Haiqiong Li and Blanca Huergo and Fan Yang and Tongxin Yin and James Wendt and Michael Bendersky and Rahma Chaabouni and Javier Snaider and Johan Ferret and Abhishek Jindal and Tara Thompson and Andrew Xue and Will Bishop and Shubham Milind Phal and Archit Sharma and Yunhsuan Sung and Prabakar Radhakrishnan and Mo Shomrat and Reeve Ingle and Roopali Vij and Justin Gilmer and Mihai Dorin Istin and Sam Sobell and Yang Lu and Emily Nottage and Dorsa Sadigh and Jeremiah Willcock and Tingnan Zhang and Steve Xu and Sasha Brown and Katherine Lee and Gary Wang and Yun Zhu and Yi Tay and Cheolmin Kim and Audrey Gutierrez and Abhanshu Sharma and Yongqin Xian and Sungyong Seo and Claire Cui and Elena Pochernina and Cip Baetu and Krzysztof Jastrzębski and Mimi Ly and Mohamed Elhawaty and Dan Suh and Eren Sezener and Pidong Wang and Nancy Yuen and George Tucker and Jiahao Cai and Zuguang Yang and Cindy Wang and Alex Muzio and Hai Qian and Jae Yoo and Derek Lockhart and Kevin R. McKee and Mandy Guo and Malika Mehrotra and Artur Mendonça and Sanket Vaibhav Mehta and Sherry Ben and Chetan Tekur and Jiaqi Mu and Muye Zhu and Victoria Krakovna and Hongrae Lee and AJ Maschinot and Sébastien Cevey and HyunJeong Choe and Aijun Bai and Hansa Srinivasan and Derek Gasaway and Nick Young and Patrick Siegler and Dan Holtmann-Rice and Vihari Piratla and Kate Baumli and Roey Yogev and Alex Hofer and Hado van Hasselt and Svetlana Grant and Yuri Chervonyi and David Silver and Andrew Hogue and Ayushi Agarwal and Kathie Wang and Preeti Singh and Four Flynn and Josh Lipschultz and Robert David and Lizzetth Bellot and Yao-Yuan Yang and Long Le and Filippo Graziano and Kate Olszewska and Kevin Hui and Akanksha Maurya and Nikos Parotsidis and Weijie Chen and Tayo Oguntebi and Joe Kelley and Anirudh Baddepudi and Johannes Mauerer and Gregory Shaw and Alex Siegman and Lin Yang and Shravya Shetty and Subhrajit Roy and Yunting Song and Wojciech Stokowiec and Ryan Burnell and Omkar Savant and Robert Busa-Fekete and Jin Miao and Samrat Ghosh and Liam MacDermed and Phillip Lippe and Mikhail Dektiarev and Zach Behrman and Fabian Mentzer and Kelvin Nguyen and Meng Wei and Siddharth Verma and Chris Knutsen and Sudeep Dasari and Zhipeng Yan and Petr Mitrichev and Xingyu Wang and Virat Shejwalkar and Jacob Austin and Srinivas Sunkara and Navneet Potti and Yan Virin and Christian Wright and Gaël Liu and Oriana Riva and Etienne Pot and Greg Kochanski and Quoc Le and Gargi Balasubramaniam and Arka Dhar and Yuguo Liao and Adam Bloniarz and Divyansh Shukla and Elizabeth Cole and Jong Lee and Sheng Zhang and Sushant Kafle and Siddharth Vashishtha and Parsa Mahmoudieh and Grace Chen and Raphael Hoffmann and Pranesh Srinivasan and Agustin Dal Lago and Yoav Ben Shalom and Zi Wang and Michael Elabd and Anuj Sharma and Junhyuk Oh and Suraj Kothawade and Maigo Le and Marianne Monteiro and Shentao Yang and Kaiz Alarakyia and Robert Geirhos and Diana Mincu and Håvard Garnes and Hayato Kobayashi and Soroosh Mariooryad and Kacper Krasowiak and Zhixin and Lai and Shibl Mourad and Mingqiu Wang and Fan Bu and Ophir Aharoni and Guanjie Chen and Abhimanyu Goyal and Vadim Zubov and Ankur Bapna and Elahe Dabir and Nisarg Kothari and Kay Lamerigts and Nicola De Cao and Jeremy Shar and Christopher Yew and Nitish Kulkarni and Dre Mahaarachchi and Mandar Joshi and Zhenhai Zhu and Jared Lichtarge and Yichao Zhou and Hannah Muckenhirn and Vittorio Selo and Oriol Vinyals and Peter Chen and Anthony Brohan and Vaibhav Mehta and Sarah Cogan and Ruth Wang and Ty Geri and Wei-Jen Ko and Wei Chen and Fabio Viola and Keshav Shivam and Lisa Wang and Madeleine Clare Elish and Raluca Ada Popa and Sébastien Pereira and Jianqiao Liu and Raphael Koster and Donnie Kim and Gufeng Zhang and Sayna Ebrahimi and Partha Talukdar and Yanyan Zheng and Petra Poklukar and Ales Mikhalap and Dale Johnson and Anitha Vijayakumar and Mark Omernick and Matt Dibb and Ayush Dubey and Qiong Hu and Apurv Suman and Vaibhav Aggarwal and Ilya Kornakov and Fei Xia and Wing Lowe and Alexey Kolganov and Ted Xiao and Vitaly Nikolaev and Steven Hemingray and Bonnie Li and Joana Iljazi and Mikołaj Rybiński and Ballie Sandhu and Peggy Lu and Thang Luong and Rodolphe Jenatton and Vineetha Govindaraj and Hui and Li and Gabriel Dulac-Arnold and Wonpyo Park and Henry Wang and Abhinit Modi and Jean Pouget-Abadie and Kristina Greller and Rahul Gupta and Robert Berry and Prajit Ramachandran and Jinyu Xie and Liam McCafferty and Jianling Wang and Kilol Gupta and Hyeontaek Lim and Blaž Bratanič and Andy Brock and Ilia Akolzin and Jim Sproch and Dan Karliner and Duhyeon Kim and Adrian Goedeckemeyer and Noam Shazeer and Cordelia Schmid and Daniele Calandriello and Parul Bhatia and Krzysztof Choromanski and Ceslee Montgomery and Dheeru Dua and Ana Ramalho and Helen King and Yue Gao and Lynn Nguyen and David Lindner and Divya Pitta and Oleaser Johnson and Khalid Salama and Diego Ardila and Michael Han and Erin Farnese and Seth Odoom and Ziyue Wang and Xiangzhuo Ding and Norman Rink and Ray Smith and Harshal Tushar Lehri and Eden Cohen and Neera Vats and Tong He and Parthasarathy Gopavarapu and Adam Paszke and Miteyan Patel and Wouter Van Gansbeke and Lucia Loher and Luis Castro and Maria Voitovich and Tamara von Glehn and Nelson George and Simon Niklaus and Zach Eaton-Rosen and Nemanja Rakićević and Erik Jue and Sagi Perel and Carrie Zhang and Yuval Bahat and Angéline Pouget and Zhi Xing and Fantine Huot and Ashish Shenoy and Taylor Bos and Vincent Coriou and Bryan Richter and Natasha Noy and Yaqing Wang and Santiago Ontanon and Siyang Qin and Gleb Makarchuk and Demis Hassabis and Zhuowan Li and Mandar Sharma and Kumaran Venkatesan and Iurii Kemaev and Roxanne Daniel and Shiyu Huang and Saloni Shah and Octavio Ponce and Warren and Chen and Manaal Faruqui and Jialin Wu and Slavica Andačić and Szabolcs Payrits and Daniel McDuff and Tom Hume and Yuan Cao and MH Tessler and Qingze Wang and Yinan Wang and Ivor Rendulic and Eirikur Agustsson and Matthew Johnson and Tanya Lando and Andrew Howard and Sri Gayatri Sundara Padmanabhan and Mayank Daswani and Andrea Banino and Michael Kilgore and Jonathan Heek and Ziwei Ji and Alvaro Caceres and Conglong Li and Nora Kassner and Alexey Vlaskin and Zeyu Liu and Alex Grills and Yanhan Hou and Roykrong Sukkerd and Gowoon Cheon and Nishita Shetty and Larisa Markeeva and Piotr Stanczyk and Tejas Iyer and Yuan Gong and Shawn Gao and Keerthana Gopalakrishnan and Tim Blyth and Malcolm Reynolds and Avishkar Bhoopchand and Misha Bilenko and Dero Gharibian and Vicky Zayats and Aleksandra Faust and Abhinav Singh and Min Ma and Hongyang Jiao and Sudheendra Vijayanarasimhan and Lora Aroyo and Vikas Yadav and Sarah Chakera and Ashwin Kakarla and Vilobh Meshram and Karol Gregor and Gabriela Botea and Evan Senter and Dawei Jia and Geza Kovacs and Neha Sharma and Sebastien Baur and Kai Kang and Yifan He and Lin Zhuo and Marija Kostelac and Itay Laish and Songyou Peng and Louis O'Bryan and Daniel Kasenberg and Girish Ramchandra Rao and Edouard Leurent and Biao Zhang and Sage Stevens and Ana Salazar and Ye Zhang and Ivan Lobov and Jake Walker and Allen Porter and Morgan Redshaw and Han Ke and Abhishek Rao and Alex Lee and Hoi Lam and Michael Moffitt and Jaeyoun Kim and Siyuan Qiao and Terry Koo and Robert Dadashi and Xinying Song and Mukund Sundararajan and Peng Xu and Chizu Kawamoto and Yan Zhong and Clara Barbu and Apoorv Reddy and Mauro Verzetti and Leon Li and George Papamakarios and Hanna Klimczak-Plucińska and Mary Cassin and Koray Kavukcuoglu and Rigel Swavely and Alain Vaucher and Jeffrey Zhao and Ross Hemsley and Michael Tschannen and Heming Ge and Gaurav Menghani and Yang Yu and Natalie Ha and Wei He and Xiao Wu and Maggie Song and Rachel Sterneck and Stefan Zinke and Dan A. Calian and Annie Marsden and Alejandro Cruzado Ruiz and Matteo Hessel and Almog Gueta and Benjamin Lee and Brian Farris and Manish Gupta and Yunjie Li and Mohammad Saleh and Vedant Misra and Kefan Xiao and Piermaria Mendolicchio and Gavin Buttimore and Varvara Krayvanova and Nigamaa Nayakanti and Matthew Wiethoff and Yash Pande and Azalia Mirhoseini and Ni Lao and Jasmine Liu and Yiqing Hua and Angie Chen and Yury Malkov and Dmitry Kalashnikov and Shubham Gupta and Kartik Audhkhasi and Yuexiang Zhai and Sudhindra Kopalle and Prateek Jain and Eran Ofek and Clemens Meyer and Khuslen Baatarsukh and Hana Strejček and Jun Qian and James Freedman and Ricardo Figueira and Michal Sokolik and Olivier Bachem and Raymond Lin and Dia Kharrat and Chris Hidey and Pingmei Xu and Dennis Duan and Yin Li and Muge Ersoy and Richard Everett and Kevin Cen and Rebeca Santamaria-Fernandez and Amir Taubenfeld and Ian Mackinnon and Linda Deng and Polina Zablotskaia and Shashank Viswanadha and Shivanker Goel and Damion Yates and Yunxiao Deng and Peter Choy and Mingqing Chen and Abhishek Sinha and Alex Mossin and Yiming Wang and Arthur Szlam and Susan Hao and Paul Kishan Rubenstein and Metin Toksoz-Exley and Miranda Aperghis and Yin Zhong and Junwhan Ahn and Michael Isard and Olivier Lacombe and Florian Luisier and Chrysovalantis Anastasiou and Yogesh Kalley and Utsav Prabhu and Emma Dunleavy and Shaan Bijwadia and Justin Mao-Jones and Kelly Chen and Rama Pasumarthi and Emily Wood and Adil Dostmohamed and Nate Hurley and Jiri Simsa and Alicia Parrish and Mantas Pajarskas and Matt Harvey and Ondrej Skopek and Yony Kochinski and Javier Rey and Verena Rieser and Denny Zhou and Sun Jae Lee and Trilok Acharya and Guowang Li and Joe Jiang and Xiaofan Zhang and Bryant Gipson and Ethan Mahintorabi and Marco Gelmi and Nima Khajehnouri and Angel Yeh and Kayi Lee and Loic Matthey and Leslie Baker and Trang Pham and Han Fu and Alex Pak and Prakhar Gupta and Cristina Vasconcelos and Adam Sadovsky and Brian Walker and Sissie Hsiao and Patrik Zochbauer and Andreea Marzoca and Noam Velan and Junhao Zeng and Gilles Baechler and Danny Driess and Divya Jain and Yanping Huang and Lizzie Tao and John Maggs and Nir Levine and Jon Schneider and Erika Gemzer and Samuel Petit and Shan Han and Zach Fisher and Dustin Zelle and Courtney Biles and Eugene Ie and Asya Fadeeva and Casper Liu and Juliana Vicente Franco and Adrian Collister and Hao Zhang and Renshen Wang and Ruizhe Zhao and Leandro Kieliger and Kurt Shuster and Rui Zhu and Boqing Gong and Lawrence Chan and Ruoxi Sun and Sujoy Basu and Roland Zimmermann and Jamie Hayes and Abhishek Bapna and Jasper Snoek and Weel Yang and Puranjay Datta and Jad Al Abdallah and Kevin Kilgour and Lu Li and SQ Mah and Yennie Jun and Morgane Rivière and Abhijit Karmarkar and Tammo Spalink and Tao Huang and Lucas Gonzalez and Duc-Hieu Tran and Averi Nowak and John Palowitch and Martin Chadwick and Ellie Talius and Harsh Mehta and Thibault Sellam and Philipp Fränken and Massimo Nicosia and Kyle He and Aditya Kini and David Amos and Sugato Basu and Harrison Jobe and Eleni Shaw and Qiantong Xu and Colin Evans and Daisuke Ikeda and Chaochao Yan and Larry Jin and Lun Wang and Sachin Yadav and Ilia Labzovsky and Ramesh Sampath and Ada Ma and Candice Schumann and Aditya Siddhant and Rohin Shah and John Youssef and Rishabh Agarwal and Natalie Dabney and Alessio Tonioni and Moran Ambar and Jing Li and Isabelle Guyon and Benny Li and David Soergel and Boya Fang and Georgi Karadzhov and Cristian Udrescu and Trieu Trinh and Vikas Raunak and Seb Noury and Dee Guo and Sonal Gupta and Mara Finkelstein and Denis Petek and Lihao Liang and Greg Billock and Pei Sun and David Wood and Yiwen Song and Xiaobin Yu and Tatiana Matejovicova and Regev Cohen and Kalyan Andra and David D'Ambrosio and Zhiwei Deng and Vincent Nallatamby and Ebrahim Songhori and Rumen Dangovski and Andrew Lampinen and Pankil Botadra and Adam Hillier and Jiawei Cao and Nagabhushan Baddi and Adhi Kuncoro and Toshihiro Yoshino and Ankit Bhagatwala and Marcáurelio Ranzato and Rylan Schaeffer and Tianlin Liu and Shuai Ye and Obaid Sarvana and John Nham and Chenkai Kuang and Isabel Gao and Jinoo Baek and Shubham Mittal and Ayzaan Wahid and Anita Gergely and Bin Ni and Josh Feldman and Carrie Muir and Pascal Lamblin and Wolfgang Macherey and Ethan Dyer and Logan Kilpatrick and Víctor Campos and Mukul Bhutani and Stanislav Fort and Yanif Ahmad and Aliaksei Severyn and Kleopatra Chatziprimou and Oleksandr Ferludin and Mason Dimarco and Aditya Kusupati and Joe Heyward and Dan Bahir and Kevin Villela and Katie Millican and Dror Marcus and Sanaz Bahargam and Caglar Unlu and Nicholas Roth and Zichuan Wei and Siddharth Gopal and Deepanway Ghoshal and Edward Lee and Sharon Lin and Jennie Lees and Dayeong Lee and Anahita Hosseini and Connie Fan and Seth Neel and Marcus Wu and Yasemin Altun and Honglong Cai and Enrique Piqueras and Josh Woodward and Alessandro Bissacco and Salem Haykal and Mahyar Bordbar and Prasha Sundaram and Sarah Hodkinson and Daniel Toyama and George Polovets and Austin Myers and Anu Sinha and Tomer Levinboim and Kashyap Krishnakumar and Rachita Chhaparia and Tatiana Sholokhova and Nitesh Bharadwaj Gundavarapu and Ganesh Jawahar and Haroon Qureshi and Jieru Hu and Nikola Momchev and Matthew Rahtz and Renjie Wu and Aishwarya P S and Kedar Dhamdhere and Meiqi Guo and Umang Gupta and Ali Eslami and Mariano Schain and Michiel Blokzijl and David Welling and Dave Orr and Levent Bolelli and Nicolas Perez-Nieves and Mikhail Sirotenko and Aman Prasad and Arjun Kar and Borja De Balle Pigem and Tayfun Terzi and Gellért Weisz and Dipankar Ghosh and Aditi Mavalankar and Dhruv Madeka and Kaspar Daugaard and Hartwig Adam and Viraj Shah and Dana Berman and Maggie Tran and Steven Baker and Ewa Andrejczuk and Grishma Chole and Ganna Raboshchuk and Mahdi Mirzazadeh and Thais Kagohara and Shimu Wu and Christian Schallhart and Bernett Orlando and Chen Wang and Alban Rrustemi and Hao Xiong and Hao Liu and Arpi Vezer and Nolan Ramsden and Shuo-yiin Chang and Sidharth Mudgal and Yan Li and Nino Vieillard and Yedid Hoshen and Farooq Ahmad and Ambrose Slone and Amy Hua and Natan Potikha and Mirko Rossini and Jon Stritar and Sushant Prakash and Zifeng Wang and Xuanyi Dong and Alireza Nazari and Efrat Nehoran and Kaan Tekelioglu and Yinxiao Li and Kartikeya Badola and Tom Funkhouser and Yuanzhen Li and Varun Yerram and Ramya Ganeshan and Daniel Formoso and Karol Langner and Tian Shi and Huijian Li and Yumeya Yamamori and Amayika Panda and Alaa Saade and Angelo Scorza Scarpati and Chris Breaux and CJ Carey and Zongwei Zhou and Cho-Jui Hsieh and Sophie Bridgers and Alena Butryna and Nishesh Gupta and Vaibhav Tulsyan and Sanghyun Woo and Evgenii Eltyshev and Will Grathwohl and Chanel Parks and Seth Benjamin and Rina Panigrahy and Shenil Dodhia and Daniel De Freitas and Chris Sauer and Will Song and Ferran Alet and Jackson Tolins and Cosmin Paduraru and Xingyi Zhou and Brian Albert and Zizhao Zhang and Lei Shu and Mudit Bansal and Sarah Nguyen and Amir Globerson and Owen Xiao and James Manyika and Tom Hennigan and Rong Rong and Josip Matak and Anton Bakalov and Ankur Sharma and Danila Sinopalnikov and Andrew Pierson and Stephen Roller and Geoff Brown and Mingcen Gao and Toshiyuki Fukuzawa and Amin Ghafouri and Kenny Vassigh and Iain Barr and Zhicheng Wang and Anna Korsun and Rajesh Jayaram and Lijie Ren and Tim Zaman and Samira Khan and Yana Lunts and Dan Deutsch and Dave Uthus and Nitzan Katz and Masha Samsikova and Amr Khalifa and Nikhil Sethi and Jiao Sun and Luming Tang and Uri Alon and Xianghong Luo and Dian Yu and Abhishek Nayyar and Bryce Petrini and Will Truong and Vincent Hellendoorn and Nikolai Chinaev and Chris Alberti and Wei Wang and Jingcao Hu and Vahab Mirrokni and Ananth Balashankar and Avia Aharon and Aahil Mehta and Ahmet Iscen and Joseph Kready and Lucas Manning and Anhad Mohananey and Yuankai Chen and Anshuman Tripathi and Allen Wu and Igor Petrovski and Dawsen Hwang and Martin Baeuml and Shreyas Chandrakaladharan and Yuan Liu and Rey Coaguila and Maxwell Chen and Sally Ma and Pouya Tafti and Susheel Tatineni and Terry Spitz and Jiayu Ye and Paul Vicol and Mihaela Rosca and Adrià Puigdomènech and Zohar Yahav and Sanjay Ghemawat and Hanzhao Lin and Phoebe Kirk and Zaid Nabulsi and Sergey Brin and Bernd Bohnet and Ken Caluwaerts and Aditya Srikanth Veerubhotla and Dan Zheng and Zihang Dai and Petre Petrov and Yichong Xu and Ramin Mehran and Zhuo Xu and Luisa Zintgraf and Jiho Choi and Spurthi Amba Hombaiah and Romal Thoppilan and Sashank Reddi and Lukasz Lew and Li Li and Kellie Webster and KP Sawhney and Lampros Lamprou and Siamak Shakeri and Mayank Lunayach and Jianmin Chen and Sumit Bagri and Alex Salcianu and Ying Chen and Yani Donchev and Charlotte Magister and Signe Nørly and Vitor Rodrigues and Tomas Izo and Hila Noga and Joe Zou and Thomas Köppe and Wenxuan Zhou and Kenton Lee and Xiangzhu Long and Danielle Eisenbud and Anthony Chen and Connor Schenck and Chi Ming To and Peilin Zhong and Emanuel Taropa and Minh Truong and Omer Levy and Danilo Martins and Zhiyuan Zhang and Christopher Semturs and Kelvin Zhang and Alex Yakubovich and Pol Moreno and Lara McConnaughey and Di Lu and Sam Redmond and Lotte Weerts and Yonatan Bitton and Tiziana Refice and Nicolas Lacasse and Arthur Conmy and Corentin Tallec and Julian Odell and Hannah Forbes-Pollard and Arkadiusz Socala and Jonathan Hoech and Pushmeet Kohli and Alanna Walton and Rui Wang and Mikita Sazanovich and Kexin Zhu and Andrei Kapishnikov and Rich Galt and Matthew Denton and Ben Murdoch and Caitlin Sikora and Kareem Mohamed and Wei Wei and Uri First and Tim McConnell and Luis C. Cobo and James Qin and Thi Avrahami and Daniel Balle and Yu Watanabe and Annie Louis and Adam Kraft and Setareh Ariafar and Yiming Gu and Eugénie Rives and Charles Yoon and Andrei Rusu and James Cobon-Kerr and Chris Hahn and Jiaming Luo and Yuvein and Zhu and Niharika Ahuja and Rodrigo Benenson and Raphaël Lopez Kaufman and Honglin Yu and Lloyd Hightower and Junlin Zhang and Darren Ni and Lisa Anne Hendricks and Gabby Wang and Gal Yona and Lalit Jain and Pablo Barrio and Surya Bhupatiraju and Siva Velusamy and Allan Dafoe and Sebastian Riedel and Tara Thomas and Zhe Yuan and Mathias Bellaiche and Sheena Panthaplackel and Klemen Kloboves and Sarthak Jauhari and Canfer Akbulut and Todor Davchev and Evgeny Gladchenko and David Madras and Aleksandr Chuklin and Tyrone Hill and Quan Yuan and Mukundan Madhavan and Luke Leonhard and Dylan Scandinaro and Qihang Chen and Ning Niu and Arthur Douillard and Bogdan Damoc and Yasumasa Onoe and Fabian Pedregosa and Fred Bertsch and Chas Leichner and Joseph Pagadora and Jonathan Malmaud and Sameera Ponda and Andy Twigg and Oleksii Duzhyi and Jingwei Shen and Miaosen Wang and Roopal Garg and Jing Chen and Utku Evci and Jonathan Lee and Leon Liu and Koji Kojima and Masa Yamaguchi and Arunkumar Rajendran and AJ Piergiovanni and Vinodh Kumar Rajendran and Marco Fornoni and Gabriel Ibagon and Harry Ragan and Sadh MNM Khan and John Blitzer and Andrew Bunner and Guan Sun and Takahiro Kosakai and Scott Lundberg and Ndidi Elue and Kelvin Guu and SK Park and Jane Park and Arunachalam Narayanaswamy and Chengda Wu and Jayaram Mudigonda and Trevor Cohn and Hairong Mu and Ravi Kumar and Laura Graesser and Yichi Zhang and Richard Killam and Vincent Zhuang and Mai Giménez and Wael Al Jishi and Ruy Ley-Wild and Alex Zhai and Kazuki Osawa and Diego Cedillo and Jialu Liu and Mayank Upadhyay and Marcin Sieniek and Roshan Sharma and Tom Paine and Anelia Angelova and Sravanti Addepalli and Carolina Parada and Kingshuk Majumder and Avery Lamp and Sanjiv Kumar and Xiang Deng and Artiom Myaskovsky and Tea Sabolić and Jeffrey Dudek and Sarah York and Félix de Chaumont Quitry and Jiazhong Nie and Dee Cattle and Alok Gunjan and Bilal Piot and Waleed Khawaja and Seojin Bang and Simon Wang and Siavash Khodadadeh and Raghavender R and Praynaa Rawlani and Richard Powell and Kevin Lee and Johannes Griesser and GS Oh and Cesar Magalhaes and Yujia Li and Simon Tokumine and Hadas Natalie Vogel and Dennis Hsu and Arturo BC and Disha Jindal and Matan Cohen and Zi Yang and Junwei Yuan and Dario de Cesare and Tony Bruguier and Jun Xu and Monica Roy and Alon Jacovi and Dan Belov and Rahul Arya and Phoenix Meadowlark and Shlomi Cohen-Ganor and Wenting Ye and Patrick Morris-Suzuki and Praseem Banzal and Gan Song and Pranavaraj Ponnuramu and Fred Zhang and George Scrivener and Salah Zaiem and Alif Raditya Rochman and Kehang Han and Badih Ghazi and Kate Lee and Shahar Drath and Daniel Suo and Antonious Girgis and Pradeep Shenoy and Duy Nguyen and Douglas Eck and Somit Gupta and Le Yan and Joao Carreira and Anmol Gulati and Ruoxin Sang and Daniil Mirylenka and Emma Cooney and Edward Chou and Mingyang Ling and Cindy Fan and Ben Coleman and Guilherme Tubone and Ravin Kumar and Jason Baldridge and Felix Hernandez-Campos and Angeliki Lazaridou and James Besley and Itay Yona and Neslihan Bulut and Quentin Wellens and AJ Pierigiovanni and Jasmine George and Richard Green and Pu Han and Connie Tao and Geoff Clark and Chong You and Abbas Abdolmaleki and Justin Fu and Tongzhou Chen and Ashwin Chaugule and Angad Chandorkar and Altaf Rahman and Will Thompson and Penporn Koanantakool and Mike Bernico and Jie Ren and Andrey Vlasov and Sergei Vassilvitskii and Maciej Kula and Yizhong Liang and Dahun Kim and Yangsibo Huang and Chengxi Ye and Dmitry Lepikhin and Wesley Helmholz},
      year={2025},
      eprint={2507.06261},
      archivePrefix={arXiv},
      primaryClass={cs.CL},
      url={https://arxiv.org/abs/2507.06261}, 
}

@misc{openai2025gpt51systemcard,
  title  = {GPT-5.1 Instant and GPT-5.1 Thinking System Card Addendum},
  author = {{OpenAI}},
  year   = {2025},
  month  = nov,
  url    = {https://cdn.openai.com/pdf/4173ec8d-1229-47db-96de-06d87147e07e/5_1_system_card.pdf},
  note   = {Accessed 2026-01-03}
}

@inproceedings{
novikova2025consistency,
title={Consistency in Language Models: Current Landscape, Challenges, and Future Directions},
author={Jekaterina Novikova and Carol Myrick Anderson and Borhane Blili-Hamelin and Subhabrata Majumdar},
booktitle={ICML 2025 Workshop on Reliable and Responsible Foundation Models},
year={2025},
url={https://openreview.net/forum?id=ejvvhJZJSf}
}

@misc{wu2024_gendec,
  title={GenDec: A robust generative Question-decomposition method for Multi-hop reasoning}, 
      author={Jian Wu and Linyi Yang and Yuliang Ji and Wenhao Huang and Börje F. Karlsson and Manabu Okumura},
      year={2024},
      eprint={2402.11166},
      archivePrefix={arXiv},
      primaryClass={cs.CL},
      url={https://arxiv.org/abs/2402.11166}, 
}

@inproceedings{knappe2024enhancing,
title={Enhancing Language Model Reasoning via Weighted Reasoning in Self-Consistency},
author={Tim Knappe and Ryan Luo Li and Ayush Chauhan and Kaylee Chhua and Kevin Zhu and Sean O'Brien},
booktitle={The 4th Workshop on Mathematical Reasoning and AI at NeurIPS'24},
year={2024},
url={https://openreview.net/forum?id=2w0CIzWlle}
}

@misc{kadavath2022languagemodelsmostlyknow,
      title={Language Models (Mostly) Know What They Know}, 
      author={Saurav Kadavath and Tom Conerly and Amanda Askell and Tom Henighan and Dawn Drain and Ethan Perez and Nicholas Schiefer and Zac Hatfield-Dodds and Nova DasSarma and Eli Tran-Johnson and Scott Johnston and Sheer El-Showk and Andy Jones and Nelson Elhage and Tristan Hume and Anna Chen and Yuntao Bai and Sam Bowman and Stanislav Fort and Deep Ganguli and Danny Hernandez and Josh Jacobson and Jackson Kernion and Shauna Kravec and Liane Lovitt and Kamal Ndousse and Catherine Olsson and Sam Ringer and Dario Amodei and Tom Brown and Jack Clark and Nicholas Joseph and Ben Mann and Sam McCandlish and Chris Olah and Jared Kaplan},
      year={2022},
      eprint={2207.05221},
      archivePrefix={arXiv},
      primaryClass={cs.CL},
      url={https://arxiv.org/abs/2207.05221}, 
}

@inproceedings{
khot2023decomposed,
title={Decomposed Prompting: A Modular Approach for Solving Complex Tasks},
author={Tushar Khot and Harsh Trivedi and Matthew Finlayson and Yao Fu and Kyle Richardson and Peter Clark and Ashish Sabharwal},
booktitle={The Eleventh International Conference on Learning Representations },
year={2023},
url={https://openreview.net/forum?id=_nGgzQjzaRy}
}

@article{wolfson-etal-2020-break,
    title = "Break It Down: A Question Understanding Benchmark",
    author = "Wolfson, Tomer  and
      Geva, Mor  and
      Gupta, Ankit  and
      Gardner, Matt  and
      Goldberg, Yoav  and
      Deutch, Daniel  and
      Berant, Jonathan",
    editor = "Johnson, Mark  and
      Roark, Brian  and
      Nenkova, Ani",
    journal = "Transactions of the Association for Computational Linguistics",
    volume = "8",
    year = "2020",
    address = "Cambridge, MA",
    publisher = "MIT Press",
    url = "https://aclanthology.org/2020.tacl-1.13",
    doi = "10.1162/tacl_a_00309",
    pages = "183--198",
}

@inproceedings{min-etal-2019-multi,
    title = "Multi-hop Reading Comprehension through Question Decomposition and Rescoring",
    author = "Min, Sewon  and
      Zhong, Victor  and
      Zettlemoyer, Luke  and
      Hajishirzi, Hannaneh",
    editor = "Korhonen, Anna  and
      Traum, David  and
      M{\`a}rquez, Llu{\'\i}s",
    booktitle = "Proceedings of the 57th Annual Meeting of the Association for Computational Linguistics",
    month = jul,
    year = "2019",
    address = "Florence, Italy",
    publisher = "Association for Computational Linguistics",
    url = "https://aclanthology.org/P19-1613",
    doi = "10.18653/v1/P19-1613",
    pages = "6097--6109",
}

@article{WANG2025100755,
title = {Empowering large language models to edge intelligence: A survey of edge efficient LLMs and techniques},
journal = {Computer Science Review},
volume = {57},
pages = {100755},
year = {2025},
issn = {1574-0137},
doi = {https://doi.org/10.1016/j.cosrev.2025.100755},
url = {https://www.sciencedirect.com/science/article/pii/S1574013725000310},
author = {Rui Wang and Zhiyong Gao and Liuyang Zhang and Shuaibing Yue and Ziyi Gao}
}

@misc{urlana2025llmsindustriallensdeciphering,
      title={LLMs with Industrial Lens: Deciphering the Challenges and Prospects -- A Survey}, 
      author={Ashok Urlana and Charaka Vinayak Kumar and Ajeet Kumar Singh and Bala Mallikarjunarao Garlapati and Srinivasa Rao Chalamala and Rahul Mishra},
      year={2025},
      eprint={2402.14558},
      archivePrefix={arXiv},
      primaryClass={cs.CL},
      url={https://arxiv.org/abs/2402.14558}, 
}

@article{qu2025mobileedgeLLM,
  title   = {Mobile Edge Intelligence for Large Language Models: A Contemporary Survey},
  author  = {Qu, Guanqiao and Chen, Qiyuan and Wei, Wei and Lin, Zheng and Chen, Xianhao and Huang, Kaibin},
  journal = {IEEE Communications Surveys \& Tutorials},
  year    = {2025},
  doi     = {10.1109/COMST.2025.3527641},
}

@inproceedings{han-gardent-2025-generating,
    title = "Generating Complex Question Decompositions in the Face of Distribution Shifts",
    author = "Han, Kelvin  and
      Gardent, Claire",
    editor = "Chiruzzo, Luis  and
      Ritter, Alan  and
      Wang, Lu",
    booktitle = "Proceedings of the 2025 Conference of the Nations of the Americas Chapter of the Association for Computational Linguistics: Human Language Technologies (Volume 1: Long Papers)",
    month = apr,
    year = "2025",
    address = "Albuquerque, New Mexico",
    publisher = "Association for Computational Linguistics",
    url = "https://aclanthology.org/2025.naacl-long.55/",
    doi = "10.18653/v1/2025.naacl-long.55",
    pages = "1189--1211",
    ISBN = "979-8-89176-189-6"
}

\appendix
\clearpage

\section{Complete Results Tables}
\label{sec:appendix_tables}

This appendix provides the complete accuracy and consistency tables across all models, datasets, and reasoning interfaces.

\subsection{Full Accuracy Results}
\label{sec: appendix_accuracy_table}

Table~\ref{table:accuracy_full} reports accuracy (\%) for all nine models
across six datasets and three reasoning interfaces (Direct, Assistive, and Incremental).

\begin{table*}[t]
\setlength\tabcolsep{2.5pt}
\small
\begin{center}
\begin{tabular}{l ccc | ccc | ccc | ccc | ccc | ccc}
\textbf{Model} & \multicolumn{3}{c}{Bamboogle} & \multicolumn{3}{c}{Mintaka} & \multicolumn{3}{c}{HotpotQA} & \multicolumn{3}{c}{CRAG} & \multicolumn{3}{c}{FRAMES} & \multicolumn{3}{c}{MuSiQue} \\ [0.1cm]
 & D & A & I & D & A & I & D & A & I & D & A & I & D & A & I & D & A & I \\
\toprule
Mistral 7B & 15.6 & 35.8 & 36.6 & 40.9 & 55.7 & 50.7 & 24.1 & 27.7 & 24.4 & 25.1 & 33.1 & 24.5 & 11.9 & 18.5 & 13.7 & 12.8 & 20.3 & 13.5 \\
Qwen 8B & 12.2 & 37.4 & 32.8 & 24.6 & 39.6 & 33.9 & 21.7 & 22.6 & 16.4 & 19.6 & 23.9 & 21.5 & 11.6 & 23.9 & 25.0 & 11.7 & 13.5 & 15.2 \\
Llama 8B & 21.3 & 39.8 & 48.0 & 33.3 & 54.7 & 58.0 & 25.3 & 26.3 & 27.5 & 27.6 & 35.6 & 41.7 & 8.9 & 17.4 & 18.8 & 9.2 & 16.2 & 16.4 \\
Qwen 32B & 18.7 & 44.7 & 43.9 & 38.6 & 59.1 & 53.4 & 31.4 & 30.7 & 27.7 & 27.0 & 37.4 & 35.6 & 14.3 & 23.2 & 21.5 & 16.0 & 22.2 & 19.9 \\
Qwen 72B & 31.7 & 58.5 & 61.0 & 53.7 & 68.3 & 71.8 & 34.9 & 36.1 & 36.1 & 31.9 & 40.5 & 41.1 & 17.4 & 29.4 & 29.0 & 23.1 & 25.9 & 26.2 \\
Llama 70B & 45.5 & 72.4 & 68.3 & 61.6 & 69.6 & 74.3 & 41.3 & 46.0 & 42.8 & 41.7 & 49.1 & 52.1 & 26.6 & 39.2 & 40.6 & 24.5 & 34.0 & 31.2 \\
Gemini Flash & 83.7 & 75.6 & 80.5 & 83.2 & 83.9 & 82.8 & 59.7 & 56.0 & 53.3 & 64.4 & 63.2 & 62.0 & 61.4 & 55.7 & 55.6 & 37.2 & 37.6 & 35.8 \\
Gemini Pro & 85.4 & 87.0 & 86.9 & 85.6 & 85.2 & 83.2 & 68.1 & 68.0 & 63.9 & 64.4 & 61.4 & 59.5 & 71.7 & 72.0 & 64.8 & 47.5 & 52.5 & 45.5 \\
GPT 5.1 & 84.5 & 87.8 & 87.0 & 86.2 & 86.9 & 82.2 & 73.7 & 72.5 & 70.7 & 65.0 & 63.8 & 62.0 & 72.3 & 72.1 & 65.9 & 53.9 & 54.3 & 46.1 \\
\bottomrule
\end{tabular}
\end{center}
\caption{Accuracy (\%) by model, dataset, and reasoning approach. D=Direct, A=Assistive, I=Incremental.}
\label{table:accuracy_full}
\end{table*}

\subsection{Full Consistency Results}
\label{sec: appendix_consistency_table}

Table~\ref{table:consistency_full} reports the consistency rate (\%)
between the Direct interface and each structured interface
(Assistive and Incremental), measuring how often
the two interfaces produce semantically equivalent answers.

\begin{table*}[t]
\setlength\tabcolsep{3.0pt}
\small
\begin{center}
\begin{tabular}{l cc | cc | cc | cc | cc | cc}
\textbf{Model} & \multicolumn{2}{c}{Bamboogle} & \multicolumn{2}{c}{Mintaka} & \multicolumn{2}{c}{HotpotQA} & \multicolumn{2}{c}{CRAG} & \multicolumn{2}{c}{FRAMES} & \multicolumn{2}{c}{MuSiQue} \\ [0.1cm]
 & A & I & A & I & A & I & A & I & A & I & A & I \\
\toprule
Mistral 7B & 22.8 & 20.3 & 40.6 & 36.2 & 26.3 & 20.8 & 24.7 & 21.6 & 17.4 & 13.0 & 13.8 & 13.1 \\
Qwen 8B & 17.9 & 14.6 & 31.5 & 28.5 & 25.2 & 19.7 & 19.8 & 19.1 & 13.9 & 10.0 & 12.8 & 16.7 \\
Llama 8B & 19.5 & 26.0 & 37.6 & 38.3 & 27.4 & 26.3 & 29.6 & 27.8 & 11.7 & 10.6 & 11.4 & 12.1 \\
Qwen 32B & 23.6 & 22.8 & 44.6 & 43.0 & 32.1 & 32.9 & 28.4 & 28.4 & 17.7 & 12.7 & 19.4 & 21.4 \\
Qwen 72B & 40.6 & 39.8 & 57.0 & 58.7 & 36.1 & 34.7 & 37.0 & 37.0 & 22.2 & 20.9 & 29.1 & 23.8 \\
Llama 70B & 48.0 & 50.4 & 59.4 & 66.1 & 40.5 & 39.4 & 51.9 & 56.2 & 25.8 & 24.0 & 26.2 & 28.4 \\
Gemini Flash & 78.0 & 83.7 & 87.6 & 88.9 & 65.7 & 57.7 & 73.5 & 72.8 & 61.9 & 58.4 & 47.5 & 43.3 \\
Gemini Pro & 91.1 & 89.4 & 93.3 & 93.0 & 75.2 & 70.4 & 83.4 & 83.4 & 78.8 & 71.6 & 58.5 & 55.2 \\
GPT 5.1 & 88.6 & 89.4 & 93.6 & 92.3 & 78.5 & 76.3 & 81.6 & 82.8 & 79.3 & 69.7 & 64.9 & 59.7 \\
\bottomrule
\end{tabular}
\end{center}
\caption{Consistency rate (\%) with direct answers by model, dataset, and reasoning approach. A=Assistive, I=Incremental.}
\label{table:consistency_full}
\end{table*}

\subsection{Cross-Regime Consistency by Dataset}

Table~\ref{tab:consistency_by_regime} summarizes cross-regime
consistency rates, averaged across all models, for each dataset.

\begin{table}[t]
\centering
\caption{Cross-regime consistency (equivalence rate with Direct), averaged across all models per dataset. Higher indicates more stable answers across reasoning approaches.}
\label{tab:consistency_by_regime}
\begin{tabular}{lcc}
\toprule
\textbf{Dataset} & \textbf{Assistive} & \textbf{Incremental} \\
\midrule
Bamboogle & 0.439 & \textbf{0.452} \\
CRAG & 0.556 & \textbf{0.572} \\
FRAMES & \textbf{0.334} & 0.295 \\
HotpotQA & 0.496 & \textbf{0.508} \\
Mintaka & \textbf{0.573} & 0.562 \\
MuSiQue & \textbf{0.290} & 0.273 \\
\midrule
\textit{Average} & \textbf{0.421} & 0.418 \\
\bottomrule
\end{tabular}
\end{table}

\subsection{Complete Baseline Results}
\label{sec:appendix_baseline_models}

Table~\ref{tab:baseline_main_prf} reports complete precision, recall,
and F1 scores for all error detection methods across the three primary
models (GPT-5.1, Llama-3.3-70B, and Qwen3-8B) discussed in
Section~\ref{sec:baselines}.

\providecommand{\tblbest}[1]{\textbf{#1}}
\providecommand{\tblbestne}[1]{\underline{#1}}
\providecommand{\tblbestboth}[1]{\underline{\textbf{#1}}}
\begin{table*}[t]
\centering
\tiny
\setlength{\tabcolsep}{2pt}
\renewcommand{\arraystretch}{1.05}
\begin{tabular}{@{}ll *{6}{cccc}@{}}
\toprule
& & \multicolumn{4}{c}{\textbf{Bamboogle}} & \multicolumn{4}{c}{\textbf{CRAG}} & \multicolumn{4}{c}{\textbf{FRAMES}} & \multicolumn{4}{c}{\textbf{HotpotQA}} & \multicolumn{4}{c}{\textbf{Mintaka}} & \multicolumn{4}{c}{\textbf{MuSiQue}} \\
\cmidrule(lr){3-6}\cmidrule(lr){7-10}\cmidrule(lr){11-14}\cmidrule(lr){15-18}\cmidrule(lr){19-22}\cmidrule(lr){23-26}
\textbf{Model} & \textbf{Method} & \textbf{P} & \textbf{R} & \textbf{F1} & \textbf{AUC} & \textbf{P} & \textbf{R} & \textbf{F1} & \textbf{AUC} & \textbf{P} & \textbf{R} & \textbf{F1} & \textbf{AUC} & \textbf{P} & \textbf{R} & \textbf{F1} & \textbf{AUC} & \textbf{P} & \textbf{R} & \textbf{F1} & \textbf{AUC} & \textbf{P} & \textbf{R} & \textbf{F1} & \textbf{AUC} \\
\midrule
GPT 5.1 & AYS & 0.64 & 0.37 & 0.47 & 0.67 & 0.75 & \tblbestne{0.59} & \tblbestne{0.66} & \tblbestne{0.74} & 0.64 & 0.42 & 0.51 & 0.67 & \tblbestboth{0.87} & 0.36 & 0.51 & 0.67 & 0.42 & \tblbestne{0.41} & 0.42 & 0.66 & \tblbestboth{0.91} & 0.24 & 0.38 & 0.61 \\
 & IC-IDK & 0.67 & 0.32 & 0.43 & 0.64 & 0.71 & 0.52 & 0.60 & 0.70 & \tblbestboth{0.73} & 0.14 & 0.24 & 0.56 & 0.74 & 0.47 & 0.58 & 0.71 & 0.37 & 0.27 & 0.31 & 0.60 & \tblbestboth{0.91} & 0.16 & 0.27 & 0.57 \\
\cmidrule(lr){2-26}
 & DBA-A & \tblbestboth{0.86} & \tblbestboth{0.63} & \tblbestboth{0.73} & \tblbestboth{0.81} & \tblbestboth{0.77} & 0.40 & 0.53 & 0.67 & 0.66 & 0.52 & 0.58 & 0.71 & 0.78 & \tblbestne{0.64} & \tblbestne{0.70} & \tblbestne{0.79} & \tblbestboth{0.68} & 0.32 & 0.43 & 0.65 & 0.75 & 0.59 & \tblbestne{0.66} & \tblbestne{0.72} \\
 & DBA-I & 0.69 & 0.47 & 0.56 & 0.72 & 0.75 & 0.37 & 0.49 & 0.65 & 0.62 & \tblbestne{0.70} & \tblbestboth{0.66} & \tblbestboth{0.77} & 0.68 & 0.61 & 0.64 & 0.75 & 0.64 & 0.39 & \tblbestne{0.48} & \tblbestne{0.68} & 0.68 & \tblbestne{0.60} & 0.64 & 0.68 \\
\cmidrule(lr){2-26}
 & Ensemble-A & 0.71 & \tblbest{0.63} & 0.67 & 0.79 & 0.73 & \tblbest{0.77} & \tblbest{0.75} & \tblbest{0.81} & 0.59 & 0.60 & 0.60 & 0.73 & 0.77 & \tblbest{0.69} & \tblbest{0.73} & \tblbest{0.81} & 0.42 & 0.51 & 0.46 & 0.70 & 0.75 & 0.65 & \tblbest{0.69} & \tblbest{0.73} \\
 & Ensemble-I & 0.67 & 0.53 & 0.59 & 0.74 & 0.73 & 0.75 & 0.74 & 0.80 & 0.56 & \tblbest{0.74} & 0.64 & \tblbest{0.77} & 0.67 & 0.68 & 0.68 & 0.78 & 0.43 & \tblbest{0.59} & \tblbest{0.49} & \tblbest{0.73} & 0.69 & \tblbest{0.66} & 0.68 & 0.71 \\
\midrule
Llama 70B & AYS & 0.81 & 0.37 & 0.51 & 0.63 & 0.82 & 0.39 & 0.53 & 0.64 & 0.87 & 0.61 & 0.72 & 0.68 & 0.79 & 0.40 & 0.53 & 0.62 & 0.69 & 0.38 & 0.49 & 0.64 & 0.90 & 0.57 & 0.70 & 0.69 \\
 & IC-IDK & 0.73 & 0.36 & 0.48 & 0.60 & 0.70 & \tblbestboth{0.81} & \tblbestne{0.75} & 0.67 & \tblbestboth{0.94} & 0.07 & 0.13 & 0.53 & 0.78 & 0.71 & 0.74 & 0.71 & 0.59 & 0.65 & 0.62 & 0.68 & \tblbestboth{0.96} & 0.12 & 0.22 & 0.55 \\
\cmidrule(lr){2-26}
 & DBA-A & \tblbestboth{0.94} & \tblbestne{0.90} & \tblbestboth{0.92} & \tblbestboth{0.91} & 0.79 & 0.65 & 0.71 & 0.71 & 0.88 & 0.89 & \tblbestne{0.89} & \tblbestboth{0.78} & \tblbestboth{0.85} & \tblbestne{0.84} & \tblbestboth{0.84} & \tblbestboth{0.81} & \tblbestboth{0.83} & 0.72 & 0.77 & 0.81 & 0.88 & \tblbestne{0.85} & \tblbestne{0.86} & \tblbestboth{0.74} \\
 & DBA-I & 0.89 & 0.81 & 0.84 & 0.84 & \tblbestboth{0.87} & 0.65 & 0.74 & \tblbestboth{0.76} & 0.87 & \tblbestne{0.91} & \tblbestne{0.89} & 0.77 & 0.83 & 0.80 & 0.81 & 0.78 & 0.82 & \tblbestne{0.77} & \tblbestboth{0.79} & \tblbestboth{0.83} & 0.86 & 0.82 & 0.84 & 0.71 \\
\cmidrule(lr){2-26}
 & Ensemble-A & 0.87 & \tblbest{0.91} & 0.89 & 0.87 & 0.76 & 0.72 & 0.74 & 0.70 & 0.85 & \tblbest{0.95} & \tblbest{0.90} & 0.75 & 0.79 & \tblbest{0.90} & \tblbest{0.84} & 0.78 & 0.73 & 0.82 & 0.77 & 0.82 & 0.85 & \tblbest{0.92} & \tblbest{0.88} & 0.72 \\
 & Ensemble-I & 0.85 & 0.85 & 0.85 & 0.84 & 0.80 & 0.74 & \tblbest{0.77} & 0.74 & 0.85 & \tblbest{0.95} & \tblbest{0.90} & 0.74 & 0.78 & 0.85 & 0.81 & 0.75 & 0.73 & \tblbest{0.84} & 0.78 & 0.82 & 0.85 & 0.91 & \tblbest{0.88} & 0.70 \\
\midrule
Qwen3 8B & AYS & 0.94 & 0.59 & 0.73 & 0.66 & \tblbestboth{0.90} & 0.64 & 0.75 & 0.68 & \tblbestboth{0.94} & 0.83 & 0.88 & 0.71 & \tblbestboth{0.91} & 0.65 & 0.76 & 0.71 & 0.90 & 0.57 & 0.70 & 0.68 & 0.93 & 0.68 & 0.79 & 0.64 \\
 & IC-IDK & 0.94 & 0.43 & 0.59 & 0.62 & 0.86 & 0.68 & 0.76 & 0.61 & \tblbestboth{0.94} & 0.29 & 0.44 & 0.57 & \tblbestboth{0.91} & 0.47 & 0.62 & 0.65 & 0.80 & 0.30 & 0.44 & 0.53 & \tblbestboth{0.95} & 0.38 & 0.54 & 0.61 \\
\cmidrule(lr){2-26}
 & DBA-A & 0.96 & \tblbestne{0.90} & \tblbestne{0.93} & \tblbestboth{0.82} & 0.89 & \tblbestne{0.89} & \tblbestne{0.89} & \tblbestboth{0.73} & \tblbestboth{0.94} & 0.92 & 0.93 & \tblbestboth{0.74} & \tblbestboth{0.91} & \tblbestne{0.83} & \tblbestne{0.87} & \tblbestboth{0.76} & \tblbestboth{0.92} & \tblbestne{0.82} & \tblbestne{0.87} & \tblbestboth{0.79} & 0.92 & \tblbestne{0.91} & \tblbestne{0.91} & \tblbestboth{0.65} \\
 & DBA-I & \tblbestboth{0.99} & 0.67 & 0.80 & 0.80 & 0.89 & \tblbestne{0.89} & \tblbestne{0.89} & \tblbestboth{0.73} & \tblbestboth{0.94} & \tblbestne{0.95} & \tblbestne{0.94} & 0.73 & 0.88 & 0.80 & 0.84 & 0.70 & 0.90 & 0.74 & 0.81 & 0.74 & 0.92 & 0.87 & 0.89 & \tblbestboth{0.65} \\
\cmidrule(lr){2-26}
 & Ensemble-A & 0.95 & \tblbest{0.94} & \tblbest{0.94} & 0.80 & 0.87 & \tblbest{0.95} & \tblbest{0.91} & 0.67 & 0.92 & 0.96 & 0.94 & 0.67 & 0.89 & \tblbest{0.93} & \tblbest{0.91} & 0.75 & 0.88 & \tblbest{0.89} & \tblbest{0.88} & 0.75 & 0.91 & \tblbest{0.95} & \tblbest{0.93} & 0.62 \\
 & Ensemble-I & 0.95 & 0.85 & 0.90 & 0.76 & 0.88 & \tblbest{0.95} & \tblbest{0.91} & 0.70 & 0.92 & \tblbest{0.99} & \tblbest{0.95} & 0.69 & 0.87 & \tblbest{0.93} & 0.89 & 0.70 & 0.85 & 0.84 & 0.85 & 0.69 & 0.91 & 0.90 & 0.91 & 0.60 \\
\bottomrule
\end{tabular}
\caption{\small Complete error-detection results (Precision, Recall, F1, AUROC; AUROC is reported as AUC in the table) for the three main models across all datasets. Within each model block and dataset/metric pair, \textbf{bold} marks the best overall value across all methods (ties included), while \underline{underlined} marks the best non-ensemble value among AYS, IC-IDK, DBA-A, and DBA-I (ties included). When a non-ensemble method is also best overall, it is both \textbf{bold} and \underline{underlined}.}
\label{tab:baseline_main_prf}
\end{table*}

Table~\ref{tab:baseline_other_prf} reports complete precision, recall,
and F1 scores for all remaining models across all datasets.

\providecommand{\tblbest}[1]{\textbf{#1}}
\providecommand{\tblbestne}[1]{\underline{#1}}
\providecommand{\tblbestboth}[1]{\underline{\textbf{#1}}}
\begin{table*}[t]
\centering
\tiny
\setlength{\tabcolsep}{2pt}
\renewcommand{\arraystretch}{1.05}
\begin{tabular}{@{}ll *{6}{cccc}@{}}
\toprule
& & \multicolumn{4}{c}{\textbf{Bamboogle}} & \multicolumn{4}{c}{\textbf{CRAG}} & \multicolumn{4}{c}{\textbf{FRAMES}} & \multicolumn{4}{c}{\textbf{HotpotQA}} & \multicolumn{4}{c}{\textbf{Mintaka}} & \multicolumn{4}{c}{\textbf{MuSiQue}} \\
\cmidrule(lr){3-6}\cmidrule(lr){7-10}\cmidrule(lr){11-14}\cmidrule(lr){15-18}\cmidrule(lr){19-22}\cmidrule(lr){23-26}
\textbf{Model} & \textbf{Method} & \textbf{P} & \textbf{R} & \textbf{F1} & \textbf{AUC} & \textbf{P} & \textbf{R} & \textbf{F1} & \textbf{AUC} & \textbf{P} & \textbf{R} & \textbf{F1} & \textbf{AUC} & \textbf{P} & \textbf{R} & \textbf{F1} & \textbf{AUC} & \textbf{P} & \textbf{R} & \textbf{F1} & \textbf{AUC} & \textbf{P} & \textbf{R} & \textbf{F1} & \textbf{AUC} \\
\midrule
Mistral 7B & AYS & \tblbestboth{0.95} & 0.38 & 0.54 & 0.63 & \tblbestboth{0.92} & 0.20 & 0.33 & 0.58 & 0.91 & 0.41 & 0.57 & 0.56 & 0.90 & 0.46 & 0.60 & 0.64 & 0.81 & 0.30 & 0.43 & 0.60 & 0.92 & 0.42 & 0.57 & 0.58 \\
 & IC-IDK & 0.92 & 0.22 & 0.36 & 0.56 & 0.90 & 0.51 & 0.65 & 0.67 & \tblbestboth{0.96} & 0.37 & 0.53 & 0.63 & \tblbestboth{0.95} & 0.10 & 0.18 & 0.54 & \tblbestboth{0.88} & 0.21 & 0.34 & 0.58 & 0.92 & 0.58 & 0.71 & 0.61 \\
\cmidrule(lr){2-26}
 & DBA-A & 0.94 & \tblbestne{0.43} & \tblbestne{0.59} & \tblbestne{0.64} & 0.88 & 0.81 & 0.84 & \tblbestne{0.75} & 0.93 & 0.87 & \tblbestne{0.90} & \tblbestboth{0.69} & 0.89 & \tblbestne{0.60} & \tblbestne{0.72} & \tblbestne{0.69} & 0.85 & \tblbestne{0.38} & \tblbestne{0.52} & \tblbestne{0.64} & \tblbestboth{0.93} & \tblbestne{0.92} & \tblbestboth{0.93} & \tblbestboth{0.72} \\
 & DBA-I & 0.91 & 0.40 & 0.56 & 0.60 & 0.87 & \tblbestboth{0.85} & \tblbestne{0.86} & 0.73 & 0.91 & \tblbestne{0.90} & \tblbestne{0.90} & 0.62 & 0.85 & 0.55 & 0.67 & 0.62 & 0.74 & 0.36 & 0.48 & 0.59 & 0.92 & \tblbestne{0.92} & 0.92 & 0.68 \\
\cmidrule(lr){2-26}
 & Ensemble-A & 0.93 & \tblbest{0.62} & \tblbest{0.75} & \tblbest{0.68} & 0.89 & 0.84 & \tblbest{0.87} & \tblbest{0.76} & 0.92 & 0.89 & 0.91 & 0.66 & 0.88 & \tblbest{0.78} & \tblbest{0.83} & \tblbest{0.71} & 0.82 & \tblbest{0.57} & \tblbest{0.67} & \tblbest{0.69} & 0.92 & \tblbest{0.94} & \tblbest{0.93} & 0.67 \\
 & Ensemble-I & 0.92 & 0.57 & 0.70 & 0.65 & 0.86 & \tblbest{0.85} & 0.86 & 0.71 & 0.91 & \tblbest{0.93} & \tblbest{0.92} & 0.61 & 0.85 & 0.72 & 0.78 & 0.65 & 0.76 & 0.55 & 0.64 & 0.65 & 0.90 & \tblbest{0.94} & 0.92 & 0.61 \\
\midrule
Llama 3.1 8B & AYS & 0.92 & 0.50 & 0.65 & 0.68 & 0.86 & 0.54 & 0.66 & 0.66 & \tblbestboth{0.97} & 0.40 & 0.57 & 0.64 & 0.89 & 0.52 & 0.66 & 0.66 & 0.86 & 0.55 & 0.67 & 0.68 & 0.95 & 0.45 & 0.61 & 0.61 \\
 & IC-IDK & 0.83 & 0.05 & 0.10 & 0.51 & \tblbestboth{0.91} & 0.25 & 0.40 & 0.59 & \tblbestboth{0.97} & 0.12 & 0.21 & 0.54 & \tblbestboth{1.00} & 0.02 & 0.04 & 0.51 & 0.82 & 0.07 & 0.13 & 0.52 & \tblbestboth{1.00} & 0.10 & 0.19 & 0.55 \\
\cmidrule(lr){2-26}
 & DBA-A & \tblbestboth{0.93} & \tblbestne{0.83} & \tblbestne{0.88} & \tblbestboth{0.80} & 0.88 & 0.84 & 0.86 & 0.76 & 0.95 & 0.92 & \tblbestne{0.94} & \tblbestboth{0.71} & 0.91 & 0.77 & \tblbestne{0.83} & \tblbestboth{0.77} & \tblbestboth{0.91} & \tblbestne{0.76} & \tblbestne{0.83} & \tblbestboth{0.80} & 0.96 & \tblbestne{0.93} & \tblbestboth{0.95} & \tblbestboth{0.77} \\
 & DBA-I & \tblbestboth{0.93} & 0.69 & 0.79 & 0.75 & 0.89 & \tblbestne{0.87} & \tblbestboth{0.88} & \tblbestboth{0.79} & 0.94 & \tblbestne{0.93} & 0.93 & 0.67 & 0.88 & \tblbestne{0.78} & \tblbestne{0.83} & 0.73 & \tblbestboth{0.91} & 0.73 & 0.81 & 0.79 & 0.94 & 0.91 & 0.92 & 0.67 \\
\cmidrule(lr){2-26}
 & Ensemble-A & 0.89 & \tblbest{0.89} & \tblbest{0.89} & 0.76 & 0.84 & 0.88 & 0.86 & 0.71 & 0.95 & 0.94 & 0.94 & 0.70 & 0.87 & 0.85 & 0.86 & 0.74 & 0.85 & \tblbest{0.85} & \tblbest{0.85} & 0.77 & 0.95 & \tblbest{0.96} & \tblbest{0.95} & 0.67 \\
 & Ensemble-I & 0.91 & 0.84 & 0.88 & 0.77 & 0.84 & \tblbest{0.91} & 0.87 & 0.71 & 0.94 & \tblbest{0.96} & \tblbest{0.95} & 0.65 & 0.86 & \tblbest{0.89} & \tblbest{0.88} & 0.73 & 0.85 & \tblbest{0.85} & \tblbest{0.85} & 0.77 & 0.93 & 0.93 & 0.93 & 0.64 \\
\midrule
Qwen3 32B & AYS & 0.90 & 0.56 & 0.69 & 0.65 & \tblbestboth{0.90} & 0.61 & 0.72 & 0.71 & \tblbestboth{0.93} & 0.74 & 0.82 & 0.71 & \tblbestboth{0.94} & 0.48 & 0.63 & 0.70 & 0.84 & 0.52 & 0.65 & 0.68 & \tblbestboth{0.94} & 0.64 & 0.76 & 0.71 \\
 & IC-IDK & 0.94 & 0.15 & 0.26 & 0.55 & 0.86 & 0.32 & 0.46 & 0.58 & 0.92 & 0.10 & 0.17 & 0.52 & 0.92 & 0.06 & 0.12 & 0.52 & \tblbestboth{1.00} & 0.10 & 0.17 & 0.55 & \tblbestboth{0.94} & 0.20 & 0.33 & 0.57 \\
\cmidrule(lr){2-26}
 & DBA-A & \tblbestboth{0.95} & \tblbestne{0.89} & \tblbestboth{0.92} & \tblbestboth{0.84} & 0.89 & \tblbestne{0.87} & \tblbestboth{0.88} & 0.79 & \tblbestboth{0.93} & 0.89 & 0.91 & \tblbestboth{0.73} & 0.86 & 0.83 & \tblbestne{0.85} & \tblbestboth{0.77} & 0.88 & \tblbestne{0.83} & \tblbestne{0.86} & \tblbestboth{0.83} & 0.93 & \tblbestne{0.90} & \tblbestboth{0.92} & \tblbestboth{0.78} \\
 & DBA-I & 0.94 & \tblbestne{0.89} & 0.91 & 0.81 & \tblbestboth{0.90} & 0.86 & \tblbestboth{0.88} & \tblbestboth{0.80} & 0.92 & \tblbestne{0.94} & \tblbestne{0.93} & 0.72 & 0.85 & \tblbestne{0.84} & \tblbestne{0.85} & 0.76 & 0.87 & \tblbestne{0.83} & 0.85 & 0.82 & 0.91 & 0.86 & 0.88 & 0.72 \\
\cmidrule(lr){2-26}
 & Ensemble-A & 0.90 & 0.90 & 0.90 & 0.73 & 0.85 & 0.89 & 0.87 & 0.74 & 0.91 & 0.96 & 0.93 & 0.68 & 0.85 & 0.89 & 0.87 & \tblbest{0.77} & 0.84 & \tblbest{0.92} & \tblbest{0.88} & 0.82 & 0.92 & \tblbest{0.93} & \tblbest{0.92} & 0.74 \\
 & Ensemble-I & 0.90 & \tblbest{0.91} & 0.91 & 0.74 & 0.86 & \tblbest{0.91} & \tblbest{0.88} & 0.74 & 0.91 & \tblbest{0.97} & \tblbest{0.94} & 0.69 & 0.84 & \tblbest{0.91} & \tblbest{0.88} & 0.76 & 0.82 & 0.91 & 0.86 & 0.78 & 0.91 & 0.90 & 0.90 & 0.70 \\
\midrule
Qwen 2.5 72B & AYS & 0.87 & 0.63 & 0.73 & 0.71 & \tblbestboth{0.91} & 0.72 & 0.80 & 0.78 & 0.91 & 0.74 & 0.81 & 0.69 & 0.84 & 0.69 & 0.76 & 0.72 & 0.70 & 0.66 & 0.68 & 0.70 & 0.82 & 0.74 & 0.78 & 0.60 \\
 & IC-IDK & 0.79 & \tblbestne{0.87} & 0.82 & 0.68 & 0.84 & 0.47 & 0.60 & 0.64 & \tblbestboth{0.97} & 0.25 & 0.40 & 0.61 & \tblbestboth{0.90} & 0.41 & 0.57 & 0.66 & 0.74 & 0.26 & 0.39 & 0.59 & \tblbestboth{0.92} & 0.42 & 0.58 & 0.65 \\
\cmidrule(lr){2-26}
 & DBA-A & 0.94 & 0.82 & \tblbestne{0.88} & \tblbestboth{0.86} & \tblbestboth{0.91} & \tblbestne{0.83} & \tblbestboth{0.87} & \tblbestboth{0.83} & 0.91 & 0.85 & 0.88 & 0.72 & 0.84 & \tblbestne{0.84} & \tblbestne{0.84} & \tblbestboth{0.77} & 0.86 & 0.77 & 0.81 & 0.83 & 0.90 & 0.83 & \tblbestne{0.86} & \tblbestboth{0.76} \\
 & DBA-I & \tblbestboth{0.97} & 0.73 & 0.83 & 0.84 & 0.90 & 0.81 & 0.85 & 0.81 & 0.93 & \tblbestne{0.89} & \tblbestne{0.91} & \tblbestboth{0.79} & 0.84 & \tblbestne{0.84} & \tblbestne{0.84} & \tblbestboth{0.77} & \tblbestboth{0.87} & \tblbestne{0.78} & \tblbestboth{0.82} & \tblbestboth{0.84} & 0.86 & \tblbestne{0.86} & \tblbestne{0.86} & 0.71 \\
\cmidrule(lr){2-26}
 & Ensemble-A & 0.87 & \tblbest{0.90} & \tblbest{0.89} & 0.81 & 0.87 & 0.87 & \tblbest{0.87} & 0.80 & 0.90 & 0.94 & 0.92 & 0.70 & 0.80 & \tblbest{0.91} & \tblbest{0.85} & 0.74 & 0.74 & \tblbest{0.90} & 0.81 & 0.80 & 0.83 & 0.93 & \tblbest{0.88} & 0.66 \\
 & Ensemble-I & 0.90 & 0.85 & 0.87 & 0.82 & 0.86 & \tblbest{0.88} & \tblbest{0.87} & 0.78 & 0.90 & \tblbest{0.96} & \tblbest{0.93} & 0.72 & 0.80 & 0.90 & \tblbest{0.85} & 0.74 & 0.72 & 0.89 & 0.79 & 0.78 & 0.81 & \tblbest{0.95} & \tblbest{0.88} & 0.61 \\
\midrule
Gemini Flash & AYS & 0.67 & 0.40 & 0.50 & 0.68 & \tblbestboth{0.85} & \tblbestne{0.67} & \tblbestne{0.75} & \tblbestne{0.80} & 0.68 & 0.50 & 0.57 & 0.68 & 0.73 & 0.29 & 0.42 & 0.61 & 0.56 & 0.43 & 0.49 & 0.68 & 0.86 & 0.27 & 0.41 & 0.60 \\
 & IC-IDK & \tblbestboth{1.00} & 0.05 & 0.10 & 0.53 & 0.64 & 0.28 & 0.39 & 0.59 & 0.70 & 0.12 & 0.21 & 0.55 & \tblbestboth{0.87} & 0.23 & 0.37 & 0.60 & 0.43 & 0.12 & 0.19 & 0.54 & \tblbestboth{0.87} & 0.27 & 0.41 & 0.60 \\
\cmidrule(lr){2-26}
 & DBA-A & 0.43 & \tblbestne{0.60} & 0.50 & 0.72 & 0.77 & 0.57 & 0.65 & 0.74 & \tblbestboth{0.72} & 0.71 & \tblbestne{0.71} & \tblbestne{0.77} & 0.74 & 0.64 & \tblbestne{0.69} & \tblbestne{0.74} & \tblbestboth{0.63} & \tblbestne{0.44} & \tblbestne{0.52} & \tblbestne{0.69} & 0.84 & 0.71 & 0.77 & \tblbestboth{0.74} \\
 & DBA-I & 0.57 & \tblbestne{0.60} & \tblbestne{0.58} & \tblbestne{0.76} & 0.73 & 0.57 & 0.64 & 0.73 & 0.69 & \tblbestne{0.74} & \tblbestne{0.71} & 0.76 & 0.66 & \tblbestne{0.70} & 0.68 & 0.73 & 0.58 & 0.35 & 0.44 & 0.65 & 0.82 & \tblbestne{0.74} & \tblbestne{0.78} & 0.73 \\
\cmidrule(lr){2-26}
 & Ensemble-A & 0.45 & \tblbest{0.75} & 0.57 & \tblbest{0.79} & 0.75 & \tblbest{0.82} & \tblbest{0.78} & \tblbest{0.83} & 0.67 & \tblbest{0.86} & \tblbest{0.75} & \tblbest{0.79} & 0.70 & 0.69 & \tblbest{0.70} & \tblbest{0.75} & 0.52 & \tblbest{0.62} & \tblbest{0.56} & \tblbest{0.75} & 0.83 & 0.75 & \tblbest{0.79} & \tblbest{0.74} \\
 & Ensemble-I & 0.52 & 0.70 & \tblbest{0.60} & \tblbest{0.79} & 0.72 & \tblbest{0.82} & 0.77 & 0.82 & 0.66 & 0.83 & 0.73 & 0.78 & 0.62 & \tblbest{0.73} & 0.67 & 0.71 & 0.48 & 0.58 & 0.53 & 0.73 & 0.81 & \tblbest{0.78} & \tblbest{0.79} & 0.73 \\
\midrule
Gemini Pro & AYS & 0.78 & 0.39 & 0.52 & 0.68 & \tblbestboth{0.80} & \tblbestne{0.56} & \tblbestne{0.66} & \tblbestne{0.74} & 0.55 & 0.22 & 0.31 & 0.57 & 0.74 & 0.22 & 0.34 & 0.59 & 0.44 & \tblbestne{0.32} & 0.37 & 0.62 & 0.74 & 0.21 & 0.33 & 0.56 \\
 & IC-IDK & \tblbestboth{1.00} & 0.06 & 0.10 & 0.53 & 0.77 & 0.30 & 0.43 & 0.62 & 0.67 & 0.05 & 0.09 & 0.52 & \tblbestboth{0.86} & 0.20 & 0.32 & 0.59 & \tblbestboth{0.71} & 0.27 & 0.39 & 0.62 & \tblbestboth{0.93} & 0.18 & 0.29 & 0.58 \\
\cmidrule(lr){2-26}
 & DBA-A & 0.57 & \tblbestne{0.44} & 0.50 & 0.69 & 0.70 & 0.33 & 0.45 & 0.63 & \tblbestboth{0.69} & 0.52 & \tblbestne{0.59} & \tblbestne{0.71} & 0.81 & \tblbestne{0.66} & \tblbestboth{0.73} & \tblbestboth{0.79} & 0.52 & 0.30 & 0.38 & 0.62 & 0.83 & \tblbestne{0.66} & \tblbestne{0.73} & \tblbestboth{0.75} \\
 & DBA-I & 0.67 & \tblbestne{0.44} & \tblbestne{0.53} & \tblbestne{0.70} & 0.65 & 0.30 & 0.41 & 0.61 & 0.58 & \tblbestne{0.58} & 0.58 & \tblbestne{0.71} & 0.72 & 0.59 & 0.65 & 0.74 & 0.65 & \tblbestne{0.32} & \tblbestne{0.43} & \tblbestne{0.64} & 0.72 & 0.61 & 0.66 & 0.68 \\
\cmidrule(lr){2-26}
 & Ensemble-A & 0.63 & \tblbest{0.67} & \tblbest{0.65} & \tblbest{0.80} & 0.76 & \tblbest{0.70} & \tblbest{0.73} & \tblbest{0.79} & 0.62 & 0.58 & \tblbest{0.60} & \tblbest{0.72} & 0.77 & \tblbest{0.67} & 0.72 & 0.78 & 0.42 & 0.47 & 0.44 & 0.67 & 0.79 & \tblbest{0.70} & \tblbest{0.74} & \tblbest{0.75} \\
 & Ensemble-I & 0.67 & 0.56 & 0.61 & 0.75 & 0.75 & 0.68 & 0.71 & 0.77 & 0.55 & \tblbest{0.65} & \tblbest{0.60} & \tblbest{0.72} & 0.72 & 0.62 & 0.66 & 0.74 & 0.47 & \tblbest{0.49} & \tblbest{0.48} & \tblbest{0.69} & 0.72 & 0.67 & 0.69 & 0.69 \\
\bottomrule
\end{tabular}
\caption{\small Complete error-detection results (Precision, Recall, F1, AUROC; AUROC is reported as AUC in the table) for all remaining models. Within each model block and dataset/metric pair, \textbf{bold} marks the best overall value across all methods (ties included), while \underline{underlined} marks the best non-ensemble value among AYS, IC-IDK, DBA-A, and DBA-I (ties included). When a non-ensemble method is also best overall, it is both \textbf{bold} and \underline{underlined}.}
\label{tab:baseline_other_prf}
\end{table*}

\subsection{Incremental Consistency Analysis}
\label{sec:appendix_incremental_consistency}

\begin{figure*}[t]
  \centering
  \begin{subfigure}[b]{0.32\textwidth}
    \centering
    \includegraphics[width=\linewidth]{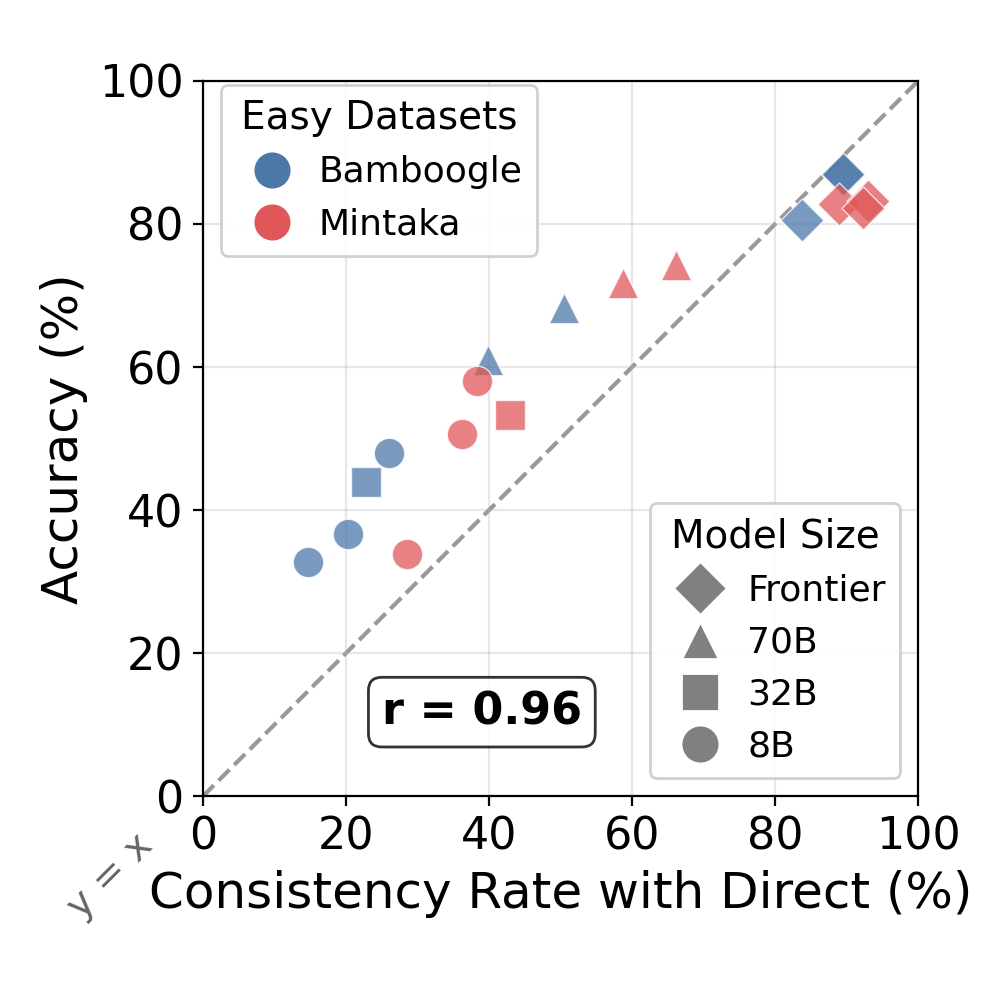}
    \caption{Easy (Bamboogle, Mintaka)}
  \end{subfigure}
  \hfill
  \begin{subfigure}[b]{0.32\textwidth}
    \centering
    \includegraphics[width=\linewidth]{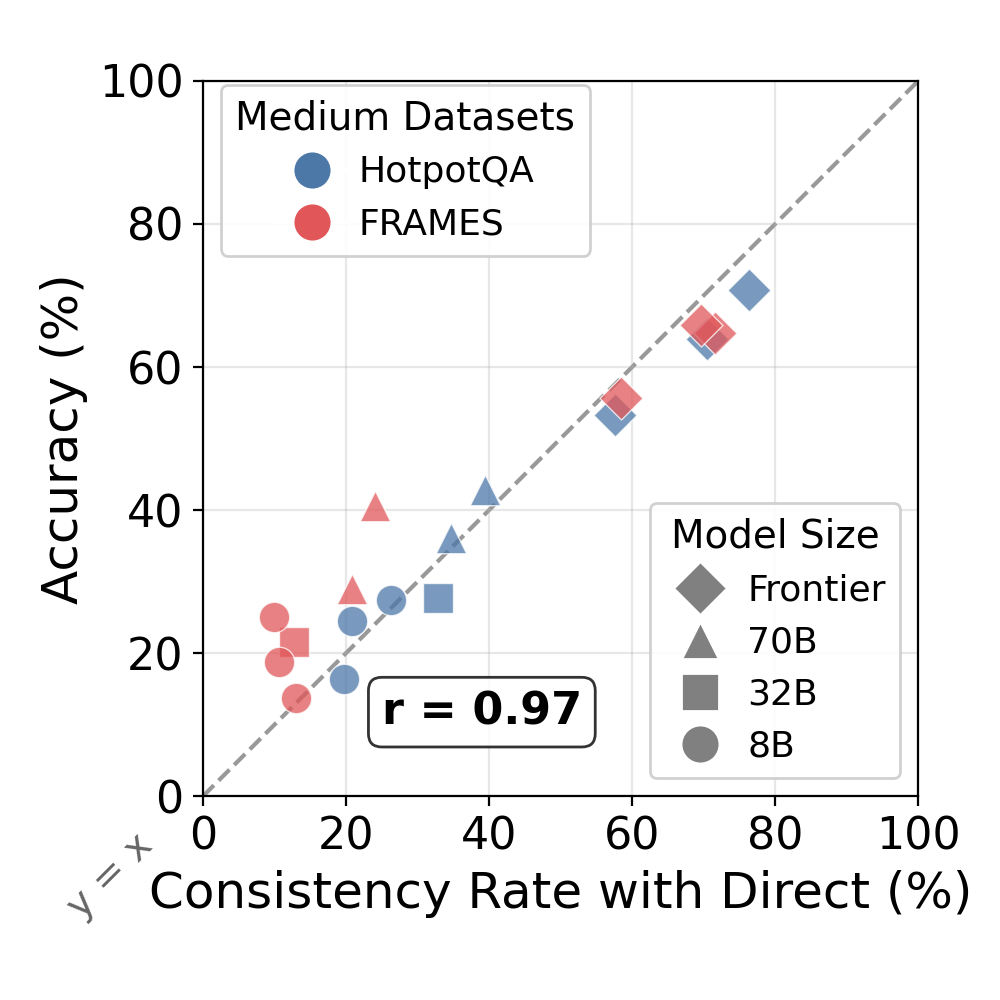}
    \caption{Medium (HotpotQA, FRAMES)}
  \end{subfigure}
  \hfill
  \begin{subfigure}[b]{0.32\textwidth}
    \centering
    \includegraphics[width=\linewidth]{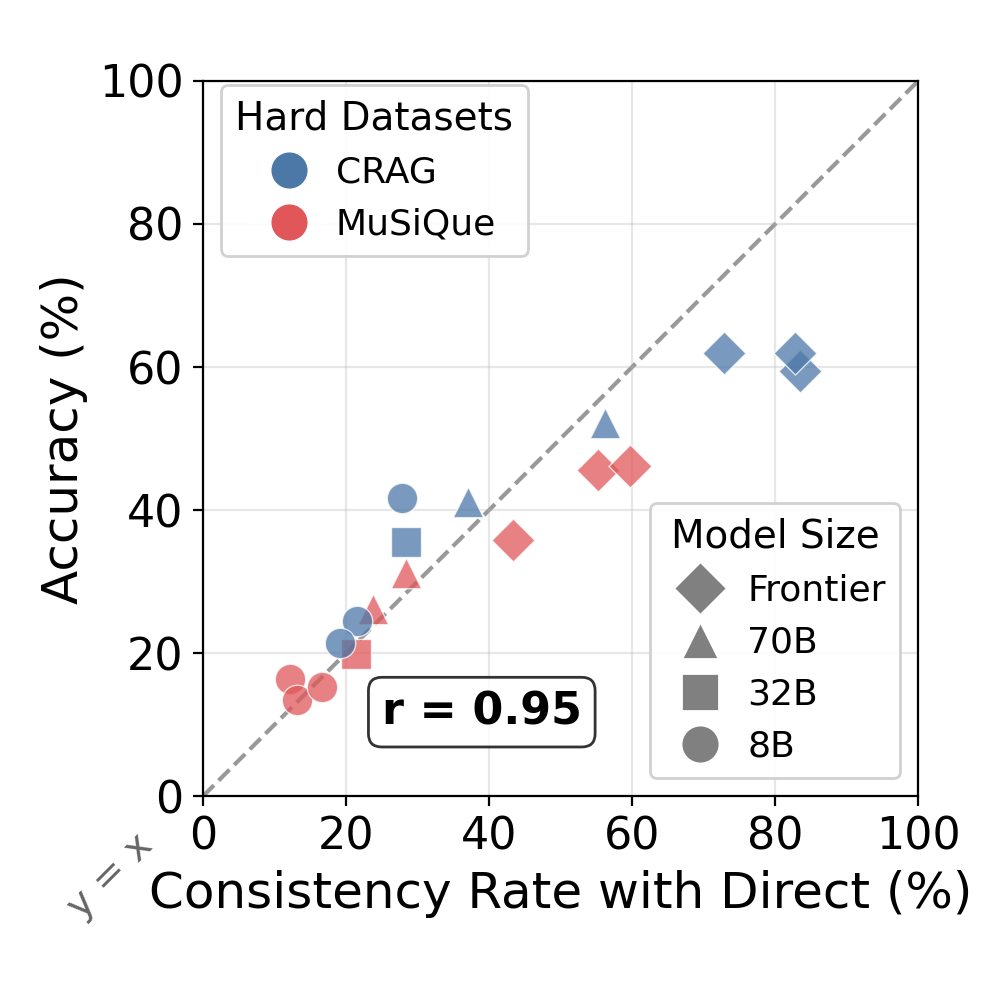}
    \caption{Hard (CRAG, MuSiQue)}
  \end{subfigure}
  \caption{Incremental accuracy vs.\ Incremental consistency rate (Direct-Incremental agreement) across 9 models and 6 datasets. Similar to the Assistive case (Figure~\ref{fig:accuracy-vs-consistency-all}), we observe strong positive correlations, confirming that the relationship between consistency and correctness generalizes across reasoning interfaces.}
  \label{fig:accuracy-vs-consistency-incremental}
\end{figure*}

Figure~\ref{fig:accuracy-vs-consistency-incremental} shows the same
accuracy--consistency relationship as
Figure~\ref{fig:accuracy-vs-consistency-all} in the main paper, but
using Incremental consistency (Direct--Incremental agreement) and
Incremental accuracy. We observe similarly strong positive correlations
across all difficulty levels, confirming that the relationship between
consistency and correctness is robust to the choice of reasoning
interface.

\subsection{Token Cost and Call Overhead}
\label{sec:token_cost_appendix}
\begin{table*}[!h]
\centering
\small
\label{tab:detailed_cost_analysis}
\begin{tabular*}{\textwidth}{l @{\extracolsep{\fill}} ccccccc}
\toprule
\textbf{Method} & \textbf{LLM Calls} & \textbf{Input Tokens} & \textbf{Output Tokens} & \textbf{Total Tokens} & \textbf{Input$\times$} & \textbf{Output$\times$} & \textbf{Total$\times$} \\
\midrule
Direct & 1 & 634 & 199 & 833 & 1.00 & 1.00 & 1.00 \\
AYS & 2 & 706 & 348 & 1,054 & 1.11 & 1.74 & 1.26 \\
IC-IDK & 1 & 753 & 210 & 963 & 1.19 & 1.06 & 1.16 \\
DBA-A & 4 & 4,197 & 505 & 4,702 & 6.61 & 2.53 & 5.64 \\
DBA-I & 6.4 & 4,374 & 538 & 4,912 & 6.89 & 2.70 & 5.89 \\                             
Self-Consistency & 8 & 4,610 & 810 & 5,420 & 7.27 & 4.07 & 6.51 \\
\bottomrule
\end{tabular*}
\caption{Detailed comparison of computational cost and token distribution across different prompting and abstention methods. All token counts and multipliers are normalized against the Direct prompting baseline.}
\label{tab:detailed_cost_analysis}
\end{table*}
Table~\ref{tab:detailed_cost_analysis} reports the full call and token overhead averaged across all evaluated methods, models, and datasets.


\section{Data Filtering and Quality Control}
\label{sec:appendix_filtering}

Table~\ref{tab:data_filtering} summarizes the dataset splits used in our
experiments, along with dataset sizes before and after filtering.

\begin{table}[!h]
\small
\centering
\begin{tabular}{l l r r r r}
\toprule
\textbf{Dataset} & \textbf{Split} & \textbf{Init.} & \textbf{Final} & \textbf{Rmvd.} & \textbf{\% Lost} \\
\midrule
Bamboogle & Full & 125 & 123 & 2 & 1.6\% \\
CRAG$^\dagger$ & train & 163 & 163 & 0 & 0\% \\
FRAMES & test & 300 & 293 & 7 & 2.3\% \\
HotpotQA & val & 300 & 274 & 26 & 8.7\% \\
Mintaka & val & 300 & 298 & 2 & 0.7\% \\
MuSiQue & val & 300 & 282 & 18 & 6.0\% \\
\midrule
\textbf{Total} & -- & \textbf{1488} & \textbf{1433} & \textbf{55} & \textbf{3.7\%} \\
\bottomrule
\end{tabular}
\caption{Dataset splits and sizes before and after filtering.
$^\dagger$CRAG uses the multi-hop subset from the original dataset; single-hop questions were excluded by design.
Other datasets use the standard evaluation split and were filtered for temporal stability, semantic clarity, and DSL validity.}
\label{tab:data_filtering}
\end{table}

\subsection{Filtering Criteria}

We manually verified all datasets and their gold decompositions and
applied the following filters to ensure high-quality, unambiguous
evaluation instances.

\paragraph{Bamboogle.}
We used 123 out of 125 of all questions in the Bamboogle dataset, as all questions are
multi-hop by design.

\paragraph{CRAG.}
The CRAG dataset contains a mix of single-hop and multi-hop questions, as
well as questions whose answers change over time. We retained only
questions that are both multi-hop and temporally stable (static or
slow-changing), as defined in the original dataset, yielding 163
questions.

\paragraph{Other Datasets.}
For HotpotQA, FRAMES, Mintaka, and MuSiQue, we began with 300 randomly
sampled questions from the standard evaluation split and applied three
successive filters:

\begin{enumerate}
    \item \textbf{Temporal Stability:}
    We excluded questions containing explicit time-dependent markers
    (e.g., ``currently'', ``now'', ``as of today'') to ensure answers
    remain valid regardless of evaluation date.

    \item \textbf{Semantic Clarity:}
    We removed ambiguous, underspecified, or logically flawed questions
    identified during manual verification of the decompositions.

    \item \textbf{DSL Validity:}
    We excluded questions whose gold decompositions were malformed,
    truncated, or did not logically entail the final answer.
\end{enumerate}

\section{Evaluation and Generation Prompts}
\label{sec:appendix_prompts}

This appendix documents the prompt templates and configurations used for
answer generation, decomposition, and evaluation across all experiments.

\subsection{LLM-as-Judge Configuration}
\label{sec:judge_config}

We evaluate answer correctness using an LLM-as-judge protocol configured
as follows:

\begin{itemize}
    \item \textbf{Judge model:} Gemini-2.5-Flash (\texttt{google/gemini-2.5-flash})
    \item \textbf{Decoding:} Greedy decoding (temperature $=0$)
    \item \textbf{Few-shot examples:} Four examples per dataset
\end{itemize}



















\subsection{Direct Answering Prompt}
\label{sec:direct_prompt}

The Direct interface uses the following zero-shot prompt for open-ended
question answering (Table~\ref{tab:prompt_direct}).

\begin{table*}[h]
\small
\begin{tabular}{p{0.97\textwidth}}
\toprule \textbf{System prompt:}\\
You are a precise question answering assistant.
Always answer the question; if unsure, provide your best single answer.
Do not expose hidden reasoning or use \texttt{<think>} tags.
\\\vspace{0.5mm}

\underline{GENERAL GUIDELINES}:
\\\vspace{0.5mm}

$\bullet$ Always return your best single answer even if uncertain.

$\bullet$ Keep prose minimal---prefer the shortest phrasing that still answers the question.
\\\vspace{0.5mm}

\underline{REFERENCE DATE}:

Treat all time-relative terms (e.g., ``now'', ``currently'') and age calculations as referring to \{reference\_date\} unless the question specifies another date.
\\\vspace{0.5mm}

\underline{NORMALIZATION}:
\\\vspace{0.5mm}

$\bullet$ Dates: if day known $\to$ YYYY-MM-DD; if only month+year $\to$ YYYY-MM; if only year $\to$ YYYY.

$\bullet$ Units: prefer symbols (km, m, ft, kg, mi$^2$, km$^2$). Keep one consistent unit.

$\bullet$ Multi-item strings: separate with semicolons exactly like ``A; B; C''.

$\bullet$ Names: prefer canonical full names (e.g., ``Thomas Edison'' not ``Edison'').
\\\vspace{0.5mm}

\underline{OUTPUT}:

\texttt{answer:}\\
\texttt{<one line with only the normalized final answer>}\\
\bottomrule
\end{tabular}
\caption{Direct answering prompt used for open-ended question answering in zero-shot setting.}
\label{tab:prompt_direct}
\end{table*}

\subsection{Correctness Evaluation Prompt}
\label{sec:correctness_prompt}

The LLM-as-judge prompt used to evaluate answer correctness is shown in
Table~\ref{tab:prompt_correctness}.

\begin{table*}[!h]
\small
\begin{tabular}{p{0.97\textwidth}}
\toprule \textbf{System prompt:}\\
You are an impartial string evaluator.
\\\vspace{0.5mm}

Given the original question, the gold (ground-truth) answer, and the prediction, decide whether the prediction communicates the same essential fact(s).
\\\vspace{0.5mm}

[question]: \{question\}

[gold]: \{gold\_answer\}

[prediction]: \{prediction\}
\\\vspace{0.5mm}

\underline{CRITICAL TOLERANCE RULES}:
\\\vspace{0.5mm}

1) \textbf{Normalization} --- Ignore case, leading/trailing spaces, and punctuation. Treat common aliases/abbreviations as equivalent (e.g., ``NYC'' $\equiv$ ``New York City'').
\\\vspace{0.5mm}

2) \textbf{Numeric expressions} --- Words vs digits are equivalent (``ten'' $\equiv$ ``10''). Accept $|a - b| \leq \max(0.5, 5\%$ of reference$)$ except for calendar dates/years, which must match exactly.
\\\vspace{0.5mm}

3) \textbf{Lists/Sets} --- If the question does not require order, treat unordered lists as equivalent when they contain the same unique items.
\\\vspace{0.5mm}

4) \textbf{Units \& formatting} --- Normalize common unit spellings/symbols (``km'' $\equiv$ ``kilometers'') and ignore thousands separators.
\\\vspace{0.5mm}

5) \textbf{Contradictions} --- If any part of the prediction contradicts the gold fact, it is incorrect.
\\\vspace{0.5mm}

6) \textbf{Extra detail} --- Extra information is acceptable if it does not contradict the gold answer.
\\\vspace{0.5mm}

Return exactly two lines: \\
Line 1: \texttt{correct: 1} or \texttt{0}\\
Line 2: \texttt{reasoning: <brief explanation>}\\
\bottomrule
\end{tabular}
\caption{LLM-as-judge prompt for evaluating answer correctness against ground truth.}
\label{tab:prompt_correctness}
\end{table*}

\subsection{Consistency Evaluation Prompt}
\label{sec:consistency_prompt}

The LLM-as-judge prompt used to evaluate pairwise answer equivalence
(consistency) is shown in Table~\ref{tab:prompt_consistency}.

\begin{table*}[h]
\small
\begin{tabular}{p{0.97\textwidth}}
\toprule \textbf{System prompt:}\\
You are an impartial pairwise string evaluator.
\\\vspace{0.5mm}

Given an original question (for context) and two candidate answers (A and B), decide if they convey the same essential facts with respect to the question, subject to the CRITICAL TOLERANCE RULES.
\\\vspace{0.5mm}

[question]: \{question\}

[answer\_a]: \{answer\_a\}

[answer\_b]: \{answer\_b\}
\\\vspace{0.5mm}

\underline{CRITICAL TOLERANCE RULES}:
\\\vspace{0.5mm}

1) \textbf{Normalization} --- Ignore case, leading/trailing spaces, punctuation, diacritics, and English articles (``the'', ``a'', ``an''). Treat common aliases/abbreviations as equivalent.
\\\vspace{0.5mm}

2) \textbf{Numeric expressions} --- Words vs digits are equivalent (``ten'' $\equiv$ ``10''). Accept $|a - b| \leq \max(0.5, 5\%$ of reference$)$ except for calendar dates/years, which must match exactly.
\\\vspace{0.5mm}

3) \textbf{Lists/Sets} --- If the question does not require order (``in order'', ``ranked'', ``first/second/third''), treat unordered lists as equivalent when they contain the same unique items.
\\\vspace{0.5mm}

4) \textbf{Mappings/Pairs} --- Treat ``A $\to$ B'', ``A: B'', ``A = B'', and ``A (B)'' as equivalent notations for the same pair.
\\\vspace{0.5mm}

5) \textbf{Units \& formatting} --- Normalize common unit spellings/symbols (``km'' $\equiv$ ``kilometers'') and ignore thousands separators.
\\\vspace{0.5mm}

6) \textbf{Yes/No} --- ``yes/true/correct'' $\equiv$ ``no/false/incorrect'' only within their respective groups.
\\\vspace{0.5mm}

7) \textbf{Contradictions} --- If any part of one answer contradicts the other regarding the same fact, they are not equivalent.
\\\vspace{0.5mm}

8) \textbf{Extra detail} --- Extra descriptive text is acceptable if it does not change or contradict the shared fact(s).
\\\vspace{0.5mm}

9) \textbf{Subset allowance} — If the question does not request a specific count (e.g., ``top k'' or ``all''), an answer that is a subset of another is acceptable provided it does not contradict any requested facts.
\\\vspace{0.5mm}

10) \textbf{Alternatives} --- If answer\_b contains semicolon-separated alternatives, compare answer\_a independently to each alternative. If any single alternative is equivalent $\to$ output equivalent: 1.
\\\vspace{0.5mm}

Return exactly two lines:\\
Line 1: \texttt{equivalent: 1} or \texttt{0}\\
Line 2: \texttt{reasoning: <brief explanation>}\\
\bottomrule
\end{tabular}
\caption{LLM-as-judge prompt for evaluating pairwise answer equivalence (consistency).}
\label{tab:prompt_consistency}
\end{table*}

\subsection{Assistive Execution Prompt}
\label{sec:assistive_prompt}

The Assistive interface receives a DSL decomposition and executes it
step-by-step using the prompt shown in
Table~\ref{tab:prompt_assistive}.

\begin{table*}[h]
\small
\begin{tabular}{p{0.97\textwidth}}
\toprule \textbf{System prompt:}\\
You will be given a short Python-flavored DSL and must execute it step by step to produce intermediate answers.
\\\vspace{0.5mm}

Each line has the form:

\texttt{answer\_k: TYPE = qa\_model("QUESTION WITH \{placeholders\}")}
\\\vspace{0.5mm}

\underline{EXECUTION RULES}:
\\\vspace{0.5mm}

$\bullet$ Execute strictly in sequence (answer\_1, answer\_2, \ldots). Later answers may reference earlier ones.

$\bullet$ Multi-item requests: return ONE string with items separated by semicolons (e.g., ``A; B; C'').

$\bullet$ Concision: values only---no extra prose; no terminal period.

$\bullet$ Always answer; if unsure, provide your best answer.
\\\vspace{0.5mm}

\underline{OUTPUT FORMAT (MANDATORY)}:

Your entire reply MUST be a single valid JSON object:

\texttt{\{``answer\_1'': ``<value>'', ``answer\_2'': ``<value>'', ...\}}\\
\bottomrule
\end{tabular}
\caption{Assistive execution prompt for DSL-guided step-by-step reasoning.}
\label{tab:prompt_assistive}
\end{table*}

\subsection{Incremental Execution Prompt}
\label{sec:incremental_prompt}

The Incremental interface processes each subquestion independently using
the prompt shown in Table~\ref{tab:prompt_incremental}.

\begin{table*}[h]
\small
\begin{tabular}{p{0.97\textwidth}}
\toprule \textbf{Subquestion system prompt:}\\
Answer the following question concisely (prefer 1--3 words when natural).
\\\vspace{0.5mm}

$\bullet$ No extra prose.

$\bullet$ Always answer; if unsure, provide your best single answer.

$\bullet$ Multi-item answers: return one string with items separated by semicolons (A; B; C).
\\\vspace{0.5mm}

Output: [Only the answer text]\\
\midrule
\textbf{Aggregation system prompt:}\\
You will be given a reasoning chain (sequence of sub-questions and answers) that builds up to a complex answer.
\\\vspace{0.5mm}

$\bullet$ Use ONLY the provided chain to answer the original question.

$\bullet$ Answer concisely (prefer 1--3 words when natural).

$\bullet$ Return only the final answer text; do not include an explanation.\\
\bottomrule
\end{tabular}
\caption{Incremental execution prompts: subquestion prompt for each hop, and aggregation prompt for final answer synthesis.}
\label{tab:prompt_incremental}
\end{table*}

\subsection{Baseline Abstention Prompts}
\label{sec:baseline_prompts}

The instructions appended to standard prompts for the abstention
baselines are shown in Table~\ref{tab:prompt_baselines}.

\begin{table*}[h]
\small
\begin{tabular}{p{0.97\textwidth}}
\toprule \textbf{IC-IDK (I Don't Know) instruction:}\\
If you are not sure you know the answer, answer with ``I don't know'' only.\\
\midrule
\textbf{AYS (Are-You-Sure) follow-up prompt:}\\
After receiving the model's initial answer, the following follow-up is issued:\\
\vspace{1mm}
Question: \{question\}\\
Claim: \{initial\_answer\}\\
\vspace{1mm}
Are you sure regarding the correctness of your claim? Please answer with Yes or No.\\
\bottomrule
\end{tabular}
\caption{Baseline abstention instructions. IC-IDK is appended to the original question prompt; AYS is a separate follow-up turn after the model provides an initial answer.}
\label{tab:prompt_baselines}
\end{table*}


\section{Extended Baseline Analysis}
\label{sec:appendix_baselines}

\subsection{Per-Model Baseline Comparison}

Tables~\ref{tab:baseline_main_prf} and~\ref{tab:baseline_other_prf} provide the complete per-model baseline results, including precision, recall, and F1 scores for all error detection methods across all models and datasets.

\subsection{Self-Consistency vs. DBA}
\label{sec:self_consistency_appendix}
We compare DBA against self-consistency \citep{wang2023selfconsistency}, a standard method for reliability estimation. We report two complementary self-consistency analyses: the original majority-vote protocol and a continuous-score evaluation based on agreement across sampled answers.

\subsubsection{Majority-vote analysis}
We compare DBA against self-consistency \citep{wang2023selfconsistency}, a standard method for reliability estimation. While DBA measures agreement across \emph{different prompting regimes} (Direct vs. Decomposed), self-consistency measures \emph{within-prompt} agreement across stochastic samples.

For each question, we generate seven candidate answers using the Direct interface: six sampled generations at temperature $T=0.7$ and one deterministic generation at $T=0$.\footnote{For GPT-5.1, we sampled 7 responses via medium reasoning effort as temperature control was not exposed.} We apply a majority vote threshold: if a single semantic claim appears in at least four of the seven candidates, the model returns that answer; otherwise, it abstains (flags the instance as an error).

We ran self-consistency in 4 datasets and present the results in Table~\ref{tab:self_consistency_comparison}.

\subsubsection{Continuous-score AUROC evaluation}

\begin{table*}[!h]
\centering
\small
\setlength{\tabcolsep}{3.5pt}
\begin{tabular}{llcccc}
\toprule
\textbf{Model} & \textbf{Method} & \textbf{Prec.} & \textbf{Rec.} & \textbf{F1} & \textbf{AUROC} \\
\midrule
\multirow{2}{*}{Qwen3-8B}
& SC    & 0.91 & 0.76 & 0.83 & 0.72 \\
& DBA-A & \textbf{0.94} & \textbf{0.91} & \textbf{0.92} & \textbf{0.74} \\
\midrule
\multirow{2}{*}{Qwen3-32B}
& SC    & 0.93 & 0.85 & 0.89 & 0.75 \\
& DBA-A & \textbf{0.94} & \textbf{0.89} & \textbf{0.92} & \textbf{0.78} \\
\midrule
\multirow{2}{*}{Llama-3.3-70B}
& SC    & 0.83 & 0.55 & 0.67 & 0.65 \\
& DBA-A & \textbf{0.90} & \textbf{0.88} & \textbf{0.89} & \textbf{0.81} \\
\midrule
\multirow{2}{*}{Gemini-2.5-Pro}
& SC    & 0.67 & \textbf{0.68} & \textbf{0.67} & \textbf{0.76} \\
& DBA-A & \textbf{0.71} & 0.55 & 0.61 & 0.73 \\
\bottomrule
\end{tabular}
\caption{Continuous-score evaluation of self-consistency.}
\label{tab:self_consistency_auc}
\end{table*}

We also re-evaluate self-consistency by treating the number of semantically equivalent answers (out of 7) as a continuous score and computing AUROC. For each question, we identify the dominant semantic claim among the seven generations and use its support count as the abstention score. We then sweep thresholds from 1 to 7, abstaining when the support count falls below the threshold, and report Precision, Recall, and F1 at the F1-optimal threshold.

We evaluate this protocol on three datasets spanning easy, medium, and hard settings (Bamboogle, FRAMES, MuSiQue) and across four representative model scales (Qwen3-8B, Qwen3-32B, Llama-3.3-70B, Gemini-2.5-Pro) as defined in Section~\ref{sec:models_experiments}. Table~\ref{tab:self_consistency_auc} reports metrics averaged across the three datasets.

Across all settings, the F1-optimal threshold is consistently high (6 or 7), suggesting that self-consistency is most reliable under near-unanimous agreement. While self-consistency achieves competitive AUROC, its optimal-threshold F1 is generally lower than DBA-A. Additionally, it requires seven sampled generations, semantic aggregation with an LLM judge, and explicit threshold tuning.



\subsection{Comparative Advantage of DBA}
Table~\ref{tab:self_consistency_comparison} demonstrates that DBA, which varies the decomposed prompting regime, is a far more effective probe for error detection than varying the \emph{decoding parameters} (Self-Consistency).

\begin{enumerate}
    \item \textbf{Recall Disparity:} \textsc{DBA-A} and \textsc{DBA-I} consistently achieve Recall scores 4--5$\times$ higher than Self-Consistency. For Qwen3-8B on \emph{MuSiQue}, Self-Consistency Recall is 0.46, whereas \textsc{DBA-A} Recall is 0.91.
    \item \textbf{Comparable Precision:} Despite the massive gain in Recall, DBA maintains Precision levels comparable to Self-Consistency (e.g., for Mistral 7B on \emph{FRAMES}, Self-Consistency Precision is 1.00 vs. \textsc{DBA-A} Precision of 0.87).
    \item \textbf{F1 Score Dominance:} Consequently, DBA dominates on F1 score. For Gemini Pro on Mintaka, Self-Consistency achieves an F1 of 0.16, while DBA-A achieves 0.38.
\end{enumerate}

\begin{table*}[t]
\setlength\tabcolsep{2pt}
\small
\centering
\begin{tabular}{ll | ccc | ccc | ccc | ccc }
\toprule
 &  & \multicolumn{3}{c}{\textbf{Bamboogle}} & \multicolumn{3}{c}{\textbf{FRAMES}} & \multicolumn{3}{c}{\textbf{Mintaka}} & \multicolumn{3}{c}{\textbf{MuSiQue}} \\
\textbf{Model} & \textbf{Method} & P & R & F1 & P & R & F1 & P & R & F1 & P & R & F1 \\
\midrule
Gemini Flash & self-consistency & 0.50 & 0.35 & 0.41 & \textbf{0.85} & 0.49 & 0.62 & 0.28 & 0.10 & 0.15 & \textbf{0.93} & 0.32 & 0.48 \\
 & DBA-A & 0.43 & \textbf{0.60} & 0.50 & 0.72 & 0.71 & 0.71 & \textbf{0.63} & \textbf{0.44} & \textbf{0.52} & 0.85 & \textbf{0.71} & \textbf{0.77} \\
 & DBA-I & \textbf{0.57} & \textbf{0.60} & \textbf{0.59} & 0.69 & \textbf{0.74} & \textbf{0.71} & 0.58 & 0.35 & 0.44 & 0.82 & \textbf{0.74} & \textbf{0.78} \\
\midrule
Gemini Pro & self-consistency & 0.50 & 0.17 & 0.25 & \textbf{0.90} & 0.34 & 0.49 & \textbf{0.80} & 0.09 & 0.16 & \textbf{0.85} & 0.26 & 0.40 \\
 & DBA-A & 0.57 & \textbf{0.44} & \textbf{0.50} & 0.69 & 0.52 & 0.59 & 0.52 & \textbf{0.30} & \textbf{0.38} & 0.83 & \textbf{0.66} & \textbf{0.73} \\
 & DBA-I & \textbf{0.67} & \textbf{0.44} & 0.53 & 0.58 & \textbf{0.58} & \textbf{0.58} & 0.65 & \textbf{0.32} & \textbf{0.43} & 0.72 & 0.62 & 0.66 \\
\midrule
GPT 5.1 & self-consistency & 0.50 & 0.18 & 0.26 & \textbf{0.91} & 0.12 & 0.22 & 0.43 & 0.15 & 0.22 & \textbf{0.86} & 0.05 & 0.09 \\
 & DBA-A & \textbf{0.86} & \textbf{0.63} & \textbf{0.73} & 0.66 & 0.52 & 0.58 & \textbf{0.68} & 0.32 & 0.43 & 0.75 & \textbf{0.59} & \textbf{0.66} \\
 & DBA-I & 0.69 & 0.47 & 0.56 & 0.62 & \textbf{0.70} & \textbf{0.66} & 0.64 & \textbf{0.39} & \textbf{0.49} & 0.68 & 0.60 & 0.64 \\
\midrule
Llama 3.1 8B & self-consistency & \textbf{1.00} & 0.07 & 0.14 & \textbf{1.00} & 0.05 & 0.09 & \textbf{1.00} & 0.08 & 0.15 & \textbf{1.00} & 0.04 & 0.08 \\
 & DBA-A & 0.93 & \textbf{0.84} & \textbf{0.88} & 0.95 & 0.92 & \textbf{0.94} & 0.91 & \textbf{0.76} & \textbf{0.83} & 0.96 & \textbf{0.93} & \textbf{0.95} \\
 & DBA-I & 0.93 & 0.69 & 0.79 & 0.94 & \textbf{0.93} & 0.93 & 0.91 & 0.74 & 0.81 & 0.94 & 0.91 & 0.92 \\
\midrule
Llama 3.3 70B & self-consistency & \textbf{1.00} & 0.02 & 0.03 & \textbf{1.00} & 0.07 & 0.13 & 0.83 & 0.04 & 0.08 & \textbf{1.00} & 0.02 & 0.05 \\
 & DBA-A & 0.94 & \textbf{0.90} & \textbf{0.92} & 0.88 & 0.89 & 0.89 & \textbf{0.83} & \textbf{0.72} & \textbf{0.77} & 0.88 & \textbf{0.85} & \textbf{0.87} \\
 & DBA-I & 0.89 & 0.81 & 0.84 & 0.87 & \textbf{0.91} & \textbf{0.89} & 0.82 & \textbf{0.77} & \textbf{0.79} & 0.86 & 0.82 & 0.84 \\
\midrule
Mistral 7B & self-consistency & 0.94 & 0.14 & 0.25 & \textbf{1.00} & 0.06 & 0.11 & \textbf{0.96} & 0.14 & 0.25 & \textbf{1.00} & 0.05 & 0.09 \\
 & DBA-A & \textbf{0.94} & \textbf{0.43} & \textbf{0.59} & 0.93 & 0.87 & 0.90 & 0.85 & \textbf{0.38} & \textbf{0.52} & 0.93 & \textbf{0.92} & \textbf{0.93} \\
 & DBA-I & 0.91 & 0.40 & 0.56 & 0.91 & \textbf{0.90} & \textbf{0.90} & 0.74 & 0.36 & 0.49 & 0.92 & \textbf{0.92} & \textbf{0.92} \\
\midrule
Qwen 2.5 72B & self-consistency & 0.88 & 0.18 & 0.30 & 0.90 & 0.23 & 0.37 & 0.68 & 0.14 & 0.23 & 0.91 & 0.13 & 0.23 \\
 & DBA-A & 0.95 & \textbf{0.82} & \textbf{0.88} & 0.91 & 0.85 & 0.88 & \textbf{0.87} & \textbf{0.77} & \textbf{0.81} & 0.90 & \textbf{0.83} & \textbf{0.86} \\
 & DBA-I & \textbf{0.97} & 0.73 & 0.83 & \textbf{0.93} & \textbf{0.89} & \textbf{0.91} & 0.87 & \textbf{0.78} & \textbf{0.82} & 0.87 & \textbf{0.86} & \textbf{0.86} \\
\midrule
Qwen3-32B & self-consistency & 0.90 & 0.25 & 0.39 & \textbf{0.97} & 0.45 & 0.61 & \textbf{1.00} & 0.12 & 0.22 & 0.93 & 0.38 & 0.54 \\
 & DBA-A & \textbf{0.95} & 0.89 & \textbf{0.92} & 0.93 & 0.89 & 0.91 & 0.88 & \textbf{0.83} & \textbf{0.86} & \textbf{0.93} & \textbf{0.90} & \textbf{0.92} \\
 & DBA-I & 0.94 & \textbf{0.89} & 0.91 & 0.92 & \textbf{0.94} & \textbf{0.93} & 0.87 & \textbf{0.83} & 0.85 & 0.91 & 0.86 & 0.88 \\
\midrule
Qwen3-8B & self-consistency & 0.92 & 0.10 & 0.18 & \textbf{0.96} & 0.38 & 0.54 & \textbf{0.96} & 0.10 & 0.17 & \textbf{0.97} & 0.46 & 0.63 \\
 & DBA-A & 0.96 & \textbf{0.90} & \textbf{0.93} & 0.94 & 0.92 & 0.93 & 0.92 & \textbf{0.82} & \textbf{0.87} & 0.92 & \textbf{0.91} & \textbf{0.91} \\
 & DBA-I & \textbf{0.99} & 0.67 & 0.80 & 0.94 & \textbf{0.95} & \textbf{0.94} & 0.90 & 0.74 & 0.81 & 0.92 & 0.87 & 0.89 \\
\bottomrule
\end{tabular}
\caption{Comparison of self-consistency vs.\@ cross-interface disagreement for error detection. self-consistency achieves high precision when it abstains but has extremely low recall, detecting only a small fraction of errors. \textsc{DBA-A} and \textsc{DBA-I} achieve similar precision while detecting 4--5$\times$ more errors. \textbf{Bold} indicates best value in each column within each model group.}
\label{tab:self_consistency_comparison}
\end{table*}

\subsection{Feasibility of Model Generated Decomposition}
\label{app:auto-decomp-feasibility}

Our main experiments use manually verified reference decompositions so
that comparisons across Direct, Assistive, and Incremental prompting are
not confounded by decomposition errors. A natural practical question,
however, is whether such decompositions must always be provided by
humans or a frontier teacher model at deployment time.

To assess this, we conduct an auxiliary feasibility analysis using
Qwen2.5-72B-Instruct as an automatic decomposition generator. We compare
its generated decompositions against the gold DSL references in two
ways. First, we perform an automatic comparison that measures how well
the generated decomposition matches the reference sequence of reasoning
steps. Second, because automatic matching may under-credit valid
paraphrases or minor structural variations, we also perform a manual
audit. This auxiliary analysis is also consistent with prior work
showing that high-quality decompositions can be obtained automatically,
either by using strong teacher models with manual verification or by
using compact-model pipelines to generate and rank synthetic
decompositions \citep{wolfson-etal-2020-break, han-gardent-2025-generating}.

For the manual audit, we inspect 120 generated decompositions (20 per
dataset) and judge whether each decomposition remains consistent with
the original question and preserves the intended multi-hop reasoning
structure. Under this criterion, 103 of 120 decompositions are valid,
for an overall validity rate of 85.8\%. This indicates that strong open
models can generate usable decompositions in most cases.

For completeness, we also report an automatic structural comparison
against the gold DSL templates. Averaged across the six datasets,
Qwen2.5-72B-Instruct achieves 0.841 precision, 0.869 recall, and 0.854
F1 under hop-level alignment, with an average hop ratio of 1.081. These
results are broadly consistent with the manual audit, although we view
the manual audit as the more interpretable measure of practical
usability. 

Overall, these results suggest that the decomposition requirement of
\textsc{DBA} is less restrictive for frontier and 70B-class models than
the main experimental setup alone might imply. At the same time, we do
not replace the manually verified decompositions in the main experiments,
since the goal of the main study is a controlled comparison under fixed,
high-quality plans, and automatic decomposition remains less dependable
for weaker models.

The automatic structural comparison reported above uses an LLM-as-judge
on top of \texttt{google/gemini-2.5-flash}; the exact prompt fed to the
judge is shown in Table~\ref{tab:prompt_decomp_equiv}. For each
question, the judge sees both the gold DSL and the model-generated DSL
and returns a strict JSON object with two fields: \texttt{equivalent\_final}
(0/1) for end-to-end equivalence, and \texttt{matches} (integer) for the
number of gold hops covered by some hop in the model decomposition.

\begin{table*}[!h]
\small
\begin{tabular}{p{0.97\textwidth}}
\toprule \textbf{System prompt:}\\
You are a careful evaluator of question decompositions. Focus on
semantic coverage, not exact wording.
\\\vspace{0.5mm}

Output must be strict JSON only (no markdown, no commentary, no
explanation).
\\\midrule
\textbf{User prompt:}\\
Task:
\\\vspace{0.5mm}

1) Decide if the two decompositions are semantically equivalent, that
is, if they would lead to the same final answer given the question.
Return \texttt{equivalent\_final} as 1 (yes) or 0 (no).
\\\vspace{0.5mm}

2) Count how many gold hops are covered by some hop in the model
decomposition. Coverage means the hop asks for the same intermediate
information (allow paraphrase and minor scope differences).
\\\vspace{0.5mm}

Return JSON with the following fields:
\\\vspace{0.5mm}

\hspace{1em}\texttt{- equivalent\_final: 0 or 1}\\
\hspace{1em}\texttt{- matches: integer (\# gold hops covered by some model hop)}
\\\vspace{0.5mm}

Dataset: \{dataset\}

Question: \{question\}
\\\vspace{0.5mm}

Gold DSL:\\
\{gold\_dsl\}
\\\vspace{0.5mm}

Gold DSL lines:\\
\{gold\_block\}
\\\vspace{0.5mm}

Model DSL:\\
\{model\_dsl\}
\\\vspace{0.5mm}

Model DSL lines:\\
\{model\_block\}
\\\bottomrule
\end{tabular}
\caption{LLM-as-judge prompt for evaluating equivalence between a
model-generated DSL decomposition and the gold reference. The judge
returns a strict JSON object with \texttt{equivalent\_final} (0/1) and
\texttt{matches} (integer hop coverage), which feeds the precision,
recall, F1, and hop-ratio statistics reported in
Section~\ref{app:auto-decomp-feasibility}.}
\label{tab:prompt_decomp_equiv}
\end{table*}

\section{Additional Qualitative Examples and Failure Modes}

Table~\ref{tab:inconsistency_breakdown_frontier} categorizes all inconsistent
examples into three primary failure modes identified during manual analysis of 100 examples across 6 datasets.
Table~\ref{tab:additional_examples} presents eight representative inconsistency
examples drawn from multiple datasets and frontier models, illustrating the
qualitative behaviors underlying these failure modes.

\begin{table*}
\setlength{\aboverulesep}{0.5pt}
\setlength{\belowrulesep}{0.5pt}
\setlength\tabcolsep{6pt}
\small
\centering
\begin{tabular}{l rrr rrr}
\toprule
& \multicolumn{3}{c}{\textbf{GPT 5.1}} & \multicolumn{3}{c}{\textbf{Gemini Pro}} \\
\cmidrule(lr){2-4} \cmidrule(lr){5-7}
\textbf{Dataset} & \textbf{Helped} & \textbf{Hurt} & \textbf{Both} & \textbf{Helped} & \textbf{Hurt} & \textbf{Both} \\
\midrule
Bamboogle & 38.1 & 19.0 & 42.9 & 25.0 & 10.0 & 65.0 \\
Mintaka & 17.0 & 14.9 & 68.1 & 17.6 & 17.6 & 64.7 \\
HotpotQA & 18.7 & 21.3 & 60.0 & 17.3 & 17.3 & 65.4 \\
CRAG & 9.7 & 11.3 & 79.0 & 10.4 & 16.4 & 73.1 \\
FRAMES & 13.9 & 18.8 & 67.3 & 17.3 & 20.2 & 62.5 \\
MuSiQue & 13.3 & 18.2 & 68.5 & 17.2 & 15.0 & 67.8 \\
\midrule
\textbf{Overall} & \textbf{15.3} & \textbf{17.6} & \textbf{67.1} & \textbf{16.7} & \textbf{16.7} & \textbf{66.5} \\
\bottomrule
\end{tabular}
\vspace{-1.0em}
\caption{Breakdown of inconsistent cases (\%) for frontier models. \emph{Helped}: Direct wrong, Assistive correct. \emph{Hurt}: Direct correct, Assistive wrong. \emph{Both}: Both regimes wrong (knowledge gap). Across both models, $\sim$67\% of disagreements reflect knowledge gaps where decomposed prompting cannot help.}
\label{tab:inconsistency_breakdown_frontier}
\vspace{-1.0em}
\end{table*}

\begin{table*}[h]
\setlength{\belowcaptionskip}{-5pt}
\footnotesize
\centering
\begin{tabular}{p{0.9cm}p{2.0cm}p{11.8cm}}
\textbf{Example} & \textbf{Case} & \textbf{Question and Reasoning Trace} \\
\toprule

\multirow{5}{=}{\textbf{Ex. 1}} & \multirow{5}{=}{\textbf{Decomposition Helped}} &
\textbf{Model:} \texttt{GPT 5.1} \quad \textbf{Dataset:} \textsc{Bamboogle} \\
& & \textbf{Q:} {When was the first location of the world's largest coffeehouse chain opened?} \\
& & \textbf{Chain:} \texttt{Q1:} ``What is the world's largest coffeehouse chain?'' $\to$ \textit{Starbucks} \\
& & \hspace{2.15em} \texttt{Q2:} ``When was the first location of \{answer\_1\} opened?'' $\to$ \textit{1971-03-30} \\
& & \textbf{Direct:} \textit{1971-03-31} \xmark \quad \textbf{Assistive:} \textit{1971-03-30} \cmark \quad \textbf{Gold:} March 30, 1971 \\
\midrule

\multirow{5}{=}{\textbf{Ex. 2}} & \multirow{5}{=}{\textbf{Decomposition Helped}} &
\textbf{Model:} \texttt{GPT 5.1} \quad \textbf{Dataset:} \textsc{Bamboogle} \\
& & \textbf{Q:} {In what year did work begin on the second longest road tunnel in the world?} \\
& & \textbf{Chain:} \texttt{Q1:} ``What is the second longest road tunnel in the world?'' $\to$ \textit{Yamate Tunnel} \\
& & \hspace{2.15em} \texttt{Q2:} ``In what year did work begin on the \{answer\_1\}?'' $\to$ \textit{1992} \\
& & \textbf{Direct:} \textit{2002} \xmark \quad \textbf{Assistive:} \textit{1992} \cmark \quad \textbf{Gold:} 1992 \\
\midrule

\multirow{6}{=}{\textbf{Ex. 3}} & \multirow{6}{=}{\textbf{Decomposition Helped}} &
\textbf{Model:} \texttt{gemini-2.5-pro} \quad \textbf{Dataset:} \textsc{FRAMES} \\
& & \textbf{Q:} {How many more career home runs did the MLB player who had the highest slugging percentage in 1954 have than the player who was the first African American to play in MLB?} \\
& & \textbf{Chain:} \texttt{Q1:} ``Which MLB player had the highest slugging percentage in 1954?'' $\to$ \textit{Willie Mays} \\
& & \hspace{2.15em} \texttt{Q2:} ``How many career home runs did \{answer\_1\} have?'' $\to$ \textit{660} \\
& & \hspace{2.15em} \texttt{Q3:} ``Who was the first African American to play in MLB?'' $\to$ \textit{Jackie Robinson} \\
& & \hspace{2.15em} \texttt{Q4:} ``How many career home runs did \{answer\_3\} have?'' $\to$ \textit{141} \quad \texttt{Q5:} $660 - 141 = 519$ \\
& & \textbf{Direct:} \textit{142} \xmark \quad \textbf{Assistive:} \textit{519} \cmark \quad \textbf{Gold:} 519 \\
\midrule


\multirow{5}{=}{\textbf{Ex. 4}} & \multirow{5}{=}{\textbf{Decomposition Hurt}} &
\textbf{Model:} \texttt{gemini-2.5-pro} \quad \textbf{Dataset:} \textsc{Mintaka} \\
& & \textbf{Q:} {What's the tallest building in the state where Yosemite National Park is located?} \\
& & \textbf{Chain:} \texttt{Q1:} ``In which state is Yosemite National Park located?'' $\to$ \textit{California} \\
& & \hspace{2.15em} \texttt{Q2:} ``What is the tallest building in \{answer\_1\}?'' $\to$ \textit{Salesforce Tower} \\
& & \textbf{Direct:} \textit{Wilshire Grand Center} \cmark \quad \textbf{Assistive:} \textit{Salesforce Tower} \xmark \quad \textbf{Gold:} Wilshire Grand Center \\
& & \textit{Chain retrieves outdated information; Wilshire Grand Center surpassed Salesforce Tower in 2017.} \\
\midrule

\multirow{5}{=}{\textbf{Ex. 5}} & \multirow{5}{=}{\textbf{Decomposition Hurt}} &
\textbf{Model:} \texttt{gemini-2.5-pro} \quad \textbf{Dataset:} \textsc{HotpotQA} \\
& & \textbf{Q:} {When was the artist who did the cover art of The Savage Frontier born?} \\
& & \textbf{Chain:} \texttt{Q1:} ``Who did the cover art for `The Savage Frontier'?'' $\to$ \textit{Keith Parkinson} \\
& & \hspace{2.15em} \texttt{Q2:} ``When was \{answer\_1\} born?'' $\to$ \textit{1958-10-22} \\
& & \textbf{Direct:} \textit{1948-08-05} \cmark \quad \textbf{Assistive:} \textit{1958-10-22} \xmark \quad \textbf{Gold:} August 5, 1948 \\
& & \textit{Wrong artist in the first hop (correct artist is Larry Elmore); the birth date retrieved is for the wrong person.} \\
\midrule

\multirow{5}{=}{\textbf{Ex. 6}} & \multirow{5}{=}{\textbf{Knowledge Gap}} &
\textbf{Model:} \texttt{GPT 5.1} \quad \textbf{Dataset:} \textsc{Bamboogle} \\
& & \textbf{Q:} {The most populous city in Punjab is how large (area wise)?} \\
& & \textbf{Chain:} \texttt{Q1:} ``What is the most populous city in Punjab?'' $\to$ \textit{Lahore} \\
& & \hspace{2.15em} \texttt{Q2:} ``What is the area of \{answer\_1\}?'' $\to$ \textit{1772 km\textsuperscript{2}} \\
& & \textbf{Direct:} \textit{159 km\textsuperscript{2}} \xmark \quad \textbf{Assistive:} \textit{1772 km\textsuperscript{2}} \xmark \quad \textbf{Gold:} 310 km\textsuperscript{2} \\
& & \textit{Both regimes retrieve incorrect area figures; the discrepancy reflects inconsistent world knowledge about city boundaries.} \\
\midrule


\multirow{5}{=}{\textbf{Ex. 7}} & \multirow{5}{=}{\textbf{Knowledge Gap}} &
\textbf{Model:} \texttt{gemini-2.5-pro} \quad \textbf{Dataset:} \textsc{MuSiQue} \\
& & \textbf{Q:} {Who is the father of the father of Anwer Ali?} \\
& & \textbf{Chain:} \texttt{Q1:} ``Who is the father of Anwer Ali?'' $\to$ \textit{Unknown} \\
& & \hspace{2.15em} \texttt{Q2:} ``Who is the father of \{answer\_1\}?'' $\to$ \textit{Unknown} \\
& & \textbf{Direct:} \textit{the grandfather of Anwer Ali} \xmark \quad \textbf{Assistive:} \textit{Unknown} \xmark \quad \textbf{Gold:} Ahmad Shah Bahadur \\
& & \textit{Obscure historical figure; both regimes fail to retrieve the correct Mughal lineage (Anwer Ali $\to$ Muhammad Shah $\to$ Ahmad Shah Bahadur).} \\
\bottomrule
\end{tabular}
\caption{Examples of inconsistencies from frontier models: \texttt{GPT 5.1} and \texttt{gemini-2.5-pro}.}
\label{tab:additional_examples}
\end{table*}

\end{document}